\documentclass[lettersize,journal]{IEEEtran}
\usepackage{amsmath,amsfonts}
\usepackage{algorithmic}
\usepackage{algorithm}
\usepackage{array}
\usepackage[caption=false,font=normalsize,labelfont=sf,textfont=sf]{subfig}
\usepackage{textcomp}
\usepackage{stfloats}
\usepackage{url}
\usepackage{verbatim}
\usepackage{graphicx}
\usepackage{bbm}
\usepackage{cite}
\usepackage{amsmath,amssymb,amsthm}
\usepackage{orcidlink}
\usepackage{multirow}
\usepackage{makecell}
\usepackage{booktabs} 
\hypersetup{hidelinks,
	colorlinks=true,
	allcolors=blue,
	pdfstartview=Fit,
	breaklinks=true}
\usepackage{tcolorbox}
\usepackage{xspace}
\usepackage{colortbl}
\newcommand{\method}{\texttt{\textbf{TAO-RL}}\xspace}

\definecolor{darkred}{rgb}{0.55, 0.0, 0.0}
\definecolor{darkgreen}{rgb}{0.0, 0.55, 0.0}

\usepackage{enumitem}

\usepackage{xcolor}
\usepackage{booktabs}
\usepackage{xcolor}
\usepackage{colortbl}
\usepackage{listings}
\definecolor{codegreen}{rgb}{0,0.6,0}
\definecolor{codeblue}{rgb}{0,0,0.6}
\definecolor{backcolour}{rgb}{0.97,0.97,0.97}
\definecolor{execresult}{rgb}{0.8, 0.3, 0.0} 
\usepackage{hyperref}

\newcommand{\codesite}{\href{https://github.com/WhyNot22222/TAO-RL}{CodeSite}}
\newenvironment{compactitemize}{\begin{itemize}[nosep,leftmargin=*]}{\end{itemize}}

\begin{document}

\title{Tool-Aware Optimization with Entropy Guidance for Efficient Agentic Reinforcement Learning}

\author{
Hongye Cao\orcidlink{0000-0002-2537-2295}, Nuo Yan\orcidlink{0009-0003-5030-7916}, Haoyuan Deng\orcidlink{0009-0002-5975-0394}, Ziwei Wang\orcidlink{0000-0001-9225-8495}, Tianpei Yang\orcidlink{0000-0002-5497-7146}, Jing Huo\orcidlink{0000-0002-8504-455X}, \\ Yuyao Zhang, and Yang Gao\orcidlink{0000-0002-2488-1813},~\IEEEmembership{Senior Member,~IEEE,}

\thanks{
Hongye Cao, Nuo Yan, Tianpei Yang, Jing Huo, and Yang Gao are with the National Key Laboratory for Novel Software Technology, Nanjing University, Nanjing 210093, China (e-mail: hongyecao528@gmail.com; ynpromise222@163.com; tianpei.yang@nju.edu.cn; huojing@nju.edu.cn;  gaoy@nju.edu.cn).
}
\thanks{
Haoyuan Deng, and Ziwei Wang are with Nanyang Technological University, Singapore (e-mail: haoyuan.deng@ntu.edu.sg; ziwei.wang@ntu.edu.sg).
}
\thanks{
Yuyao Zhang is with China Mobile NineVerse Artificial Intelligence Technology (Beijing) Co., Ltd., and Institute of Artificial Intelligence, NineVerse, Beijing 100032, China (e-mail: zhangyuyao@cmjt.chinamobile.com).
}
}

\markboth{Journal of \LaTeX\ Class Files,~Vol.~14, No.~8, August~2021}%
{Shell \MakeLowercase{\textit{et al.}}: A Sample Article Using IEEEtran.cls for IEEE Journals}


\maketitle

\begin{abstract}
Agentic reinforcement learning (RL) equips large language models (LLMs) with tool-use capabilities that substantially improve reasoning on complex tasks. However, integrating external tools often destabilizes training: over-reliance on tools can induce input distribution shift, while overly conservative tool use limits effective exploration. To address this issue, we propose a unified framework \method that couples tool-aware trajectory filtering with entropy-guided exploration for efficient policy optimization. Specifically, at the data level, \method filters rollout trajectories along two criteria: discarding those where all tool invocations fail to execute, and removing those where all rollouts are either correct or incorrect, as both cases yield degenerate advantage estimates that contribute no discriminative learning signal. This joint filtering retains data that are both tool-capable and informative, establishing a high-quality training distribution. At the algorithmic level, we introduce a tool-aware entropy-guided bonus that reshapes the advantage function at post-tool-call tokens, encouraging the policy to explore more diverse reasoning paths at critical decision points. These two components are mutually reinforcing: trajectory filtering establishes a clean and informative training foundation, while entropy-guided exploration drives stronger reasoning behaviors at critical tool-interaction junctures. Extensive experiments on $7$ challenging reasoning benchmarks across $3$ model scales demonstrate the superiority of \method over existing methods.
\end{abstract}

\begin{IEEEkeywords}
Agentic reinforcement learning, tool-aware, trajectory filtering, reasoning, entropy
\end{IEEEkeywords}

\section{Introduction}
\IEEEPARstart{A}{gentic} reinforcement learning (RL) has emerged as a powerful post-training paradigm for improving the reasoning ability of large language models (LLMs)~\cite{zhang2026the,singh2025agentic,jiang2025verltool,li2025torl,11123142,10938647,zheng2026advanced}. By enabling models to interact with external tools, such as code interpreters and search engines, LLMs have shown clear gains on complex multi-step reasoning tasks~\cite{dong2025agentic,xu2026incentivizing,dong2026agentic,feng2025retool,bai2025towards,wang2025otc}. However, these gains come with a central challenge: once tools are integrated into the training loop, policy optimization often becomes substantially less stable~\cite{mai2025agent,wang2025ragen,yu2025demystifying}.

\begin{figure}[t]
  \centering
    \includegraphics[width=0.49\linewidth]{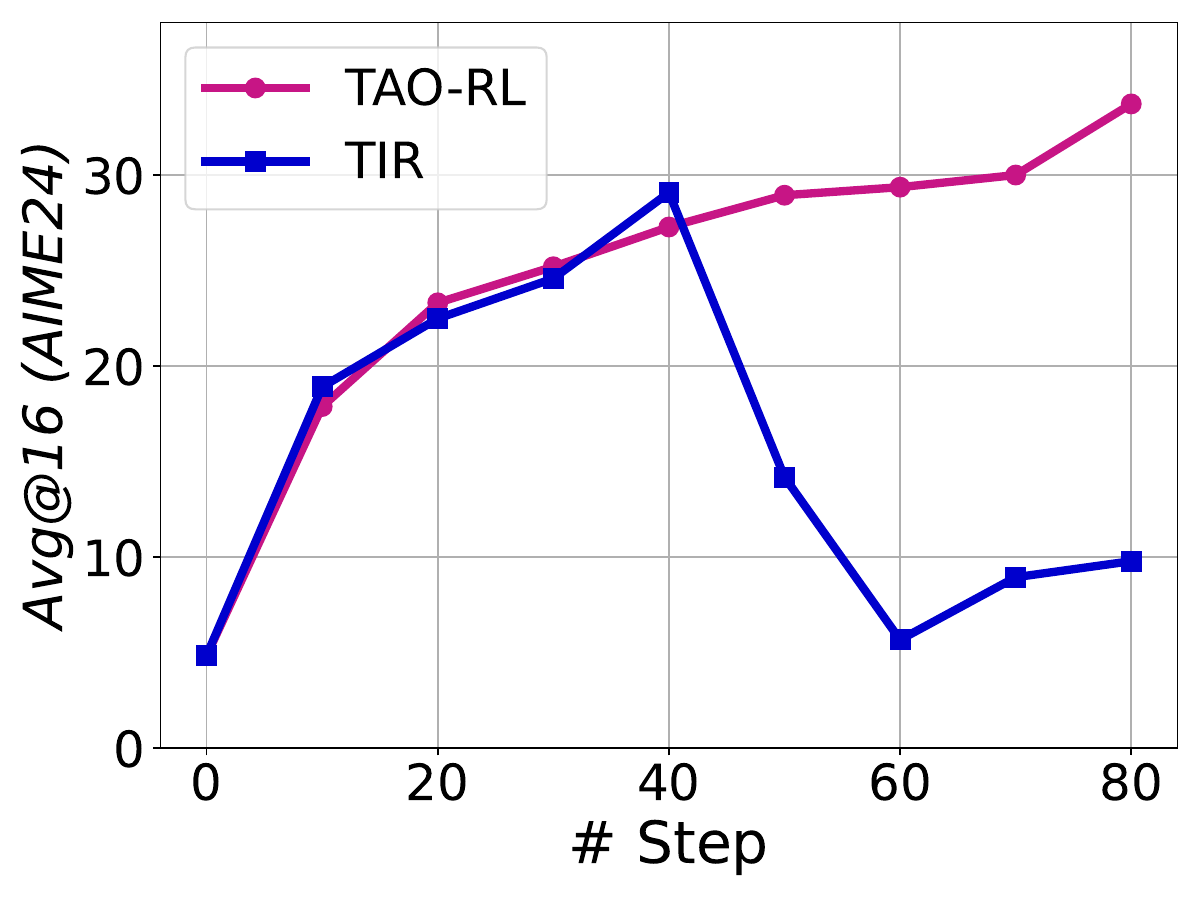} 
  \includegraphics[width=0.49\linewidth]{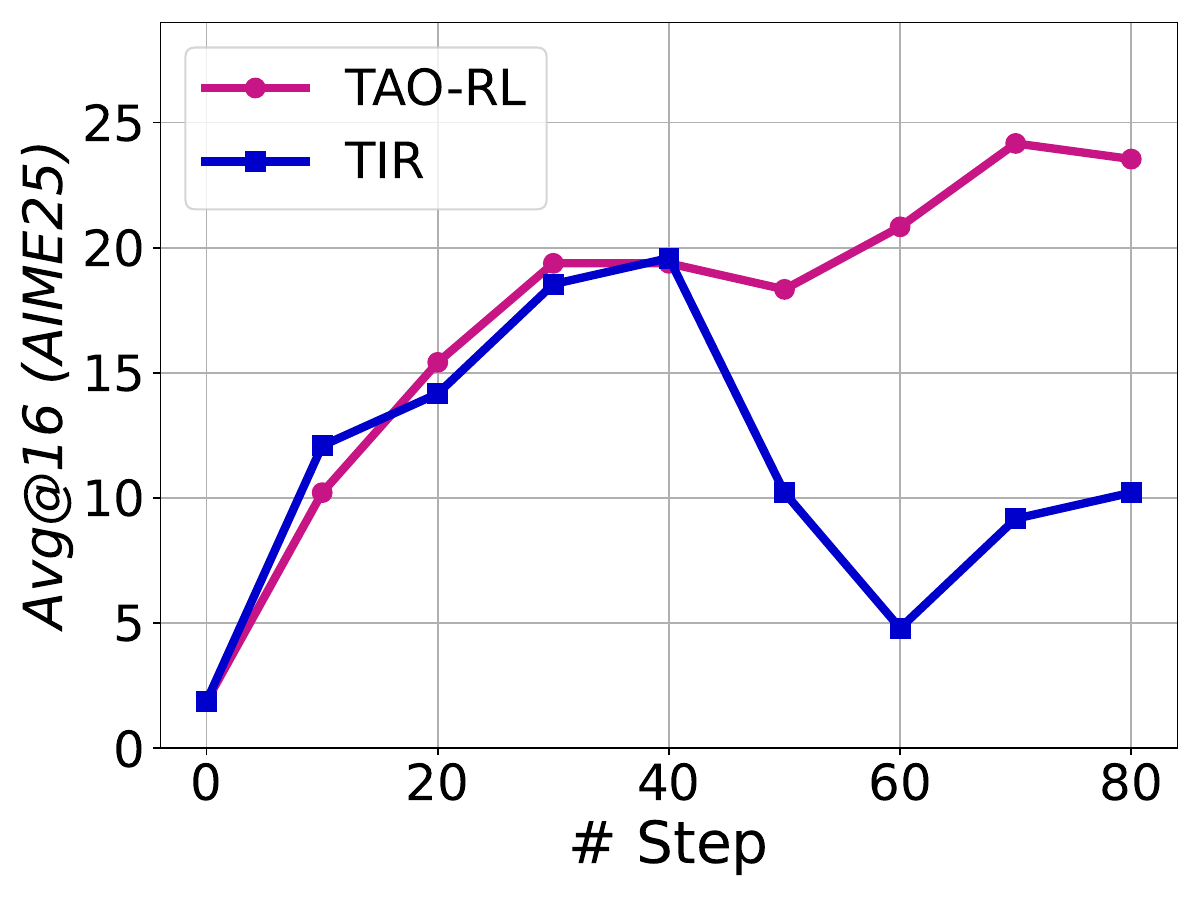} 
    \includegraphics[width=0.49\linewidth]{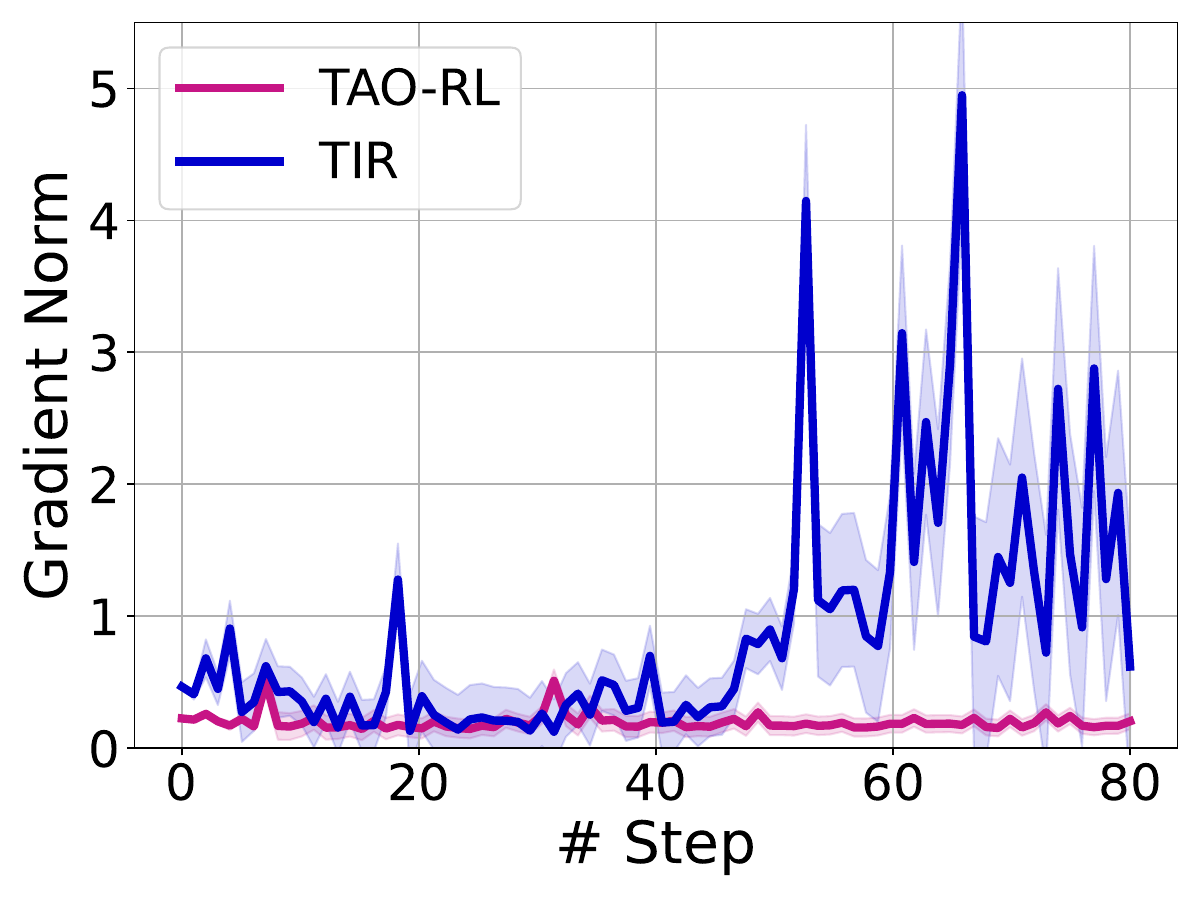} 
  \includegraphics[width=0.49\linewidth]{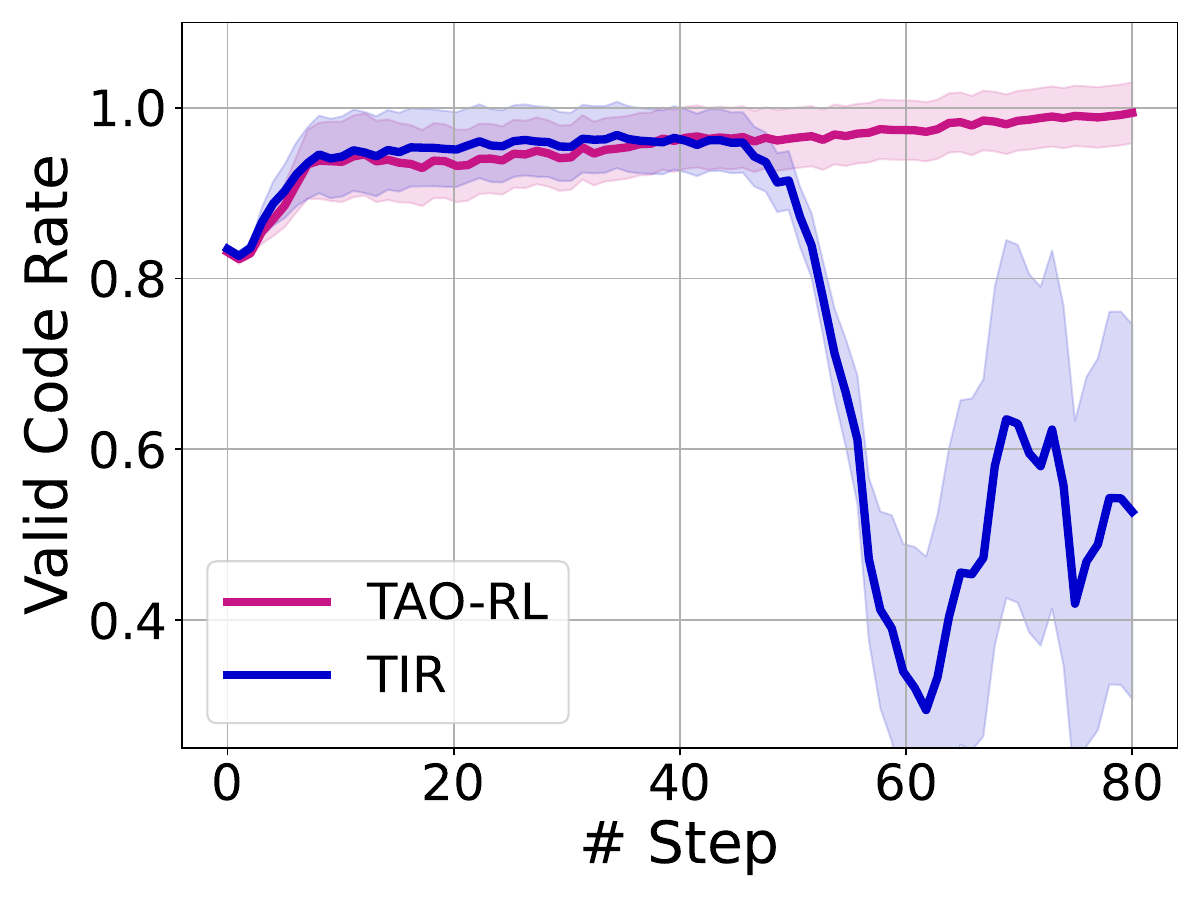}
  \caption{Comparison of tool-integrated reasoning (TIR) method and \method across \textit{Avg@16} in AIME 24\&25, gradient norm and valid code rate under Qwen2.5-7B base model during training.}
  \label{fig:example}
  \vspace{-2mm}
\end{figure}

This instability stems from how tool interaction reshapes model inputs. After each tool call, external execution results are injected back into the context, causing abrupt distributional shifts in token patterns and trajectory dynamics~\cite{xue2026simpletir,mai2025agent,qian2025toolrl}. These shifts increase stochasticity and make policy updates brittle~\cite{shang2025rstar2,bai2025towards}. As illustrated in Fig.~\ref{fig:example}, TIR exhibits large fluctuations in gradient norm, valid code rate, and average accuracy throughout training, empirically confirming that naively integrating tools destabilizes policy optimization. Furthermore, an overly conservative strategy that suppresses tool use is also undesirable, as it limits exploration and prevents the policy from discovering high-value agentic behaviors. Therefore, balancing training stability with effective exploration is a critical challenge in agentic RL.
\IEEEpubidadjcol

Existing works addressing this challenge focus primarily on the rollout process. One prominent line centers on rollout data curation, filtering trajectories with malformed formats or low-quality outputs~\cite{xue2026simpletir,yu2025demystifying}. But these approaches only discard trajectories based on format validity alone, ignoring whether the tool actually executed successfully and produced informative environmental feedback. Another line concentrates rollout resources on critical tokens through branching or targeted sampling~\cite{bai2025towards,dong2025agentic,dong2026agentic}, these methods identify high-value decision points based on model-internal signals, without considering how successful tool outputs reshape the informational value of post-tool-call reasoning steps. As a result, neither line of work comprehensively accounts for the effects introduced by tool interaction. \emph{How to simultaneously stabilize training and enhance exploration for agentic policy optimization thus remains a critical challenge.}

To tackle this challenge, we propose \method, built on a key insight that \emph{stabilizing tool-integrated training noise followed by targeted exploration drives efficient optimization.}
\method couples tool-aware trajectory filtering with entropy-guided exploration, operating jointly at the data and algorithmic levels to drive effective policy optimization. \textit{At the data level}, \method filters rollout trajectories according to two complementary criteria. First, trajectories in which tool invocations fail to execute are discarded, removing trajectories where tool interaction provides no informative environmental feedback and only injects noise into training. Second, trajectories where all rollouts are correct or all are incorrect are removed: the former indicates tasks already trivially solved by the current policy, yielding zero advantage estimates, while the latter provides no positive reward signal from which the policy can learn. This joint filtering retains trajectories that are both tool-capable and exhibit genuine answer diversity, constructing a high-quality training distribution that avoids the degenerate cases making uninformative or unstable policy updates.

\textit{At the algorithmic level}, we introduce a tool-aware entropy-guided exploration bonus based on the filtered training distribution. Rather than applying exploration incentives uniformly across all tokens, we identify tokens that follow tool executions and exhibit high entropy as critical decision points. By applying targeted entropy bonuses at these positions, \method encourages the policy to explore more diverse reasoning paths precisely where exploration is critical, progressively raising the reasoning ceiling without destabilizing the training foundation established by trajectory filtering.
The two components are mutually reinforcing: trajectory filtering establishes a clean and informative training foundation by removing failed tool interactions and uninformative trajectories, while entropy-guided exploration drives stronger agentic reasoning behaviors at the critical tokens. Our contributions of this work can be summarized as follows:

\begin{compactitemize}
\item We propose \method, a unified framework that innovatively couples data-level trajectory filtering with algorithm-level entropy-guided exploration, achieving stable training through principled data selection and effective exploration at critical tool-aware decision points.
\item \method integrates a two-criterion trajectory filtering strategy that jointly considers tool execution validity and answer discriminability to construct a high-quality training distribution, paired with an entropy-guided advantage reshaping mechanism that concentrates exploration incentives at post-tool-call tokens exhibiting high entropy.
\item Extensive experiments on seven challenging mathematical reasoning benchmarks across three base model scales demonstrate that \method consistently outperforms existing methods, achieving superior reasoning performance and training stability across diverse tasks.
\end{compactitemize}

The remainder of this paper is organized as follows. Section~\ref{sec:related work} provides an overview of related works, including agentic RL, and entropy-based exploration in RL. Section~\ref{sec:preliminaries} details the necessary preliminaries. Section~\ref{sec:approach} presents the proposed approach in detail. In section \ref{sec:experiments}, we introduce the experiments conducted to demonstrate the superiority of \method. Finally, section~\ref{sec:conclusion} draws conclusions and discusses future works.

\section{Related Work}
\label{sec:related work}
\subsection{Agentic RL}
Agentic RL extends LLMs from passive text generation to active environmental interaction, requiring multi-turn decision-making and precise tool-use strategies \cite{shinn2023reflexion,li2025search,mai2025agent,luo2025agentmath,chen2026agent,dong2025tool}. Recent advancements have transitioned from heuristic-based prompt engineering to end-to-end policy optimization via environmental feedback. On the architectural side, frameworks such as VERLTOOL \cite{jiang2025verltool} and ARTIST \cite{singh2025agentic} focus on the co-design of distributed RL systems and agentic reasoning traces. Similarly, TORL \cite{li2025torl} and ReTool \cite{feng2025retool} investigate the scalability of integrating code execution into the training loop to improve sample efficiency. On the algorithmic side, the focus shifts to stabilize the training of tool-integrated reasoning. For instance, SimpleTIR \cite{xue2026simpletir} mitigates gradient instability by filtering void turns, while GRPO-RoC \cite{shang2025rstar2} addresses environment noise through a resample-on-correct strategy.

However, existing methods do not comprehensively account for the effects introduced by tool interaction. Data curation approaches filter trajectories based on format or output quality, but ignore whether tool executions actually succeeded or produced valid environmental feedback that is critical in agentic settings where failed tool calls inject pure noise into the training distribution. Token-level sampling identifies high-value decision points based on model-internal signals alone, without considering how successful tool outputs reshape the informational value of subsequent reasoning steps. Our work addresses both gaps within a unified framework: at the data level, we apply tool-aware trajectory filtering that jointly considers tool execution validity and answer discriminability; at the algorithmic level, we introduce an entropy-guided advantage reshaping mechanism that concentrates exploration incentives specifically at post-tool-call tokens.

\subsection{Entropy-Based Exploration in RL}
Entropy-based exploration prevents entropy collapse, the rapid degradation of generation diversity frequently observed in DAPO \cite{yu2025dapo}, standard PPO ~\cite{schulman2017proximal} and GRPO \cite{shao2024deepseekmath}. To counter this, methods like integrating entropy into advantage functions encourage exploratory reasoning \cite{jin2025revisiting,cheng2025reasoning,10783450,10476495,cui2025entropy} and selectively penalize high-covariance tokens to halt entropy decline~\cite{cao2025efficient}, and others restrict policy updates to pivotal high-entropy minority tokens based on the "80/20 rule" \cite{wang2025beyond}, or dynamically scale learning signals via entropy-based advantage reweighting \cite{deng2025token}. In agentic RL, ARPO \cite{dong2026agentic} leverages characteristic entropy spikes following tool invocations to balance rollout sampling, while AEPO \cite{dong2025agentic} extends this by penalizing consecutive high-entropy steps to prevent over-branching.
Despite these advances, existing entropy-based methods apply exploration incentives without accounting for the structural role of tool interaction. \method addresses both gaps by coupling tool-aware trajectory filtering with entropy-guided advantage reshaping explicitly localized to high-entropy post-tool-call tokens, achieving precise and stable exploration at the most consequential agentic decision points.

\section{Preliminary}
\label{sec:preliminaries}

\paragraph{GRPO~\cite{shao2024deepseekmath}}

\method is built upon GRPO as its basic optimization algorithm. GRPO stabilizes policy updates by normalizing advantage estimates over groups of sampled responses, eliminating the need for a separate value network. For each query $q$, the rollout policy $\pi_{\theta_{\mathrm{old}}}$ generates a group of responses $\{o^i\}^{G}_{i=1}$ with corresponding rewards $\{R^i\}^{G}_{i=1}$, and the advantage is estimated as:
\begin{equation}    
\hat{A}^i_t=\frac{r^i-\mathrm{mean}(\{R^i\}^{G}_{i=1}) }{\mathrm{std}(\{R^i\}^{G}_{i=1}) }.
\label{eq:adv}
\end{equation}

The learning objective of GRPO is:
\begin{equation*}
\label{eq:grpo}
    \begin{aligned}
\mathcal{J}_{\mathrm{GRPO}}(\theta) 
&= \mathbb{E}_{(q,a)\sim \mathcal{D},\,\{o^i\}_{i=1}^G \sim \pi_{\theta_{\mathrm{old}}}(\cdot | q)} 
\\
& \Bigg[ \frac{1}{G} \sum_{i=1}^{G} \frac{1}{|o^i|} 
\sum_{t=1}^{|o^i|}
\Big( \min\!\big(r^i_t(\theta)\hat{A}^i_t,\, \mathrm{clip}(r^i_t(\theta),
\\
& 1-\epsilon,\, 1+\epsilon)\hat{A}^i_t\big)
- \beta D_{\mathrm{KL}}(\pi_{\theta} \,\|\, \pi_{\mathrm{ref}}) \Big)
\Bigg],
    \end{aligned}
\end{equation*}
where $\epsilon$ is the clipping hyperparameter and $\beta$ controls the strength of the KL penalty.

\begin{figure*}[t]
\centering
\includegraphics[width=1\linewidth]{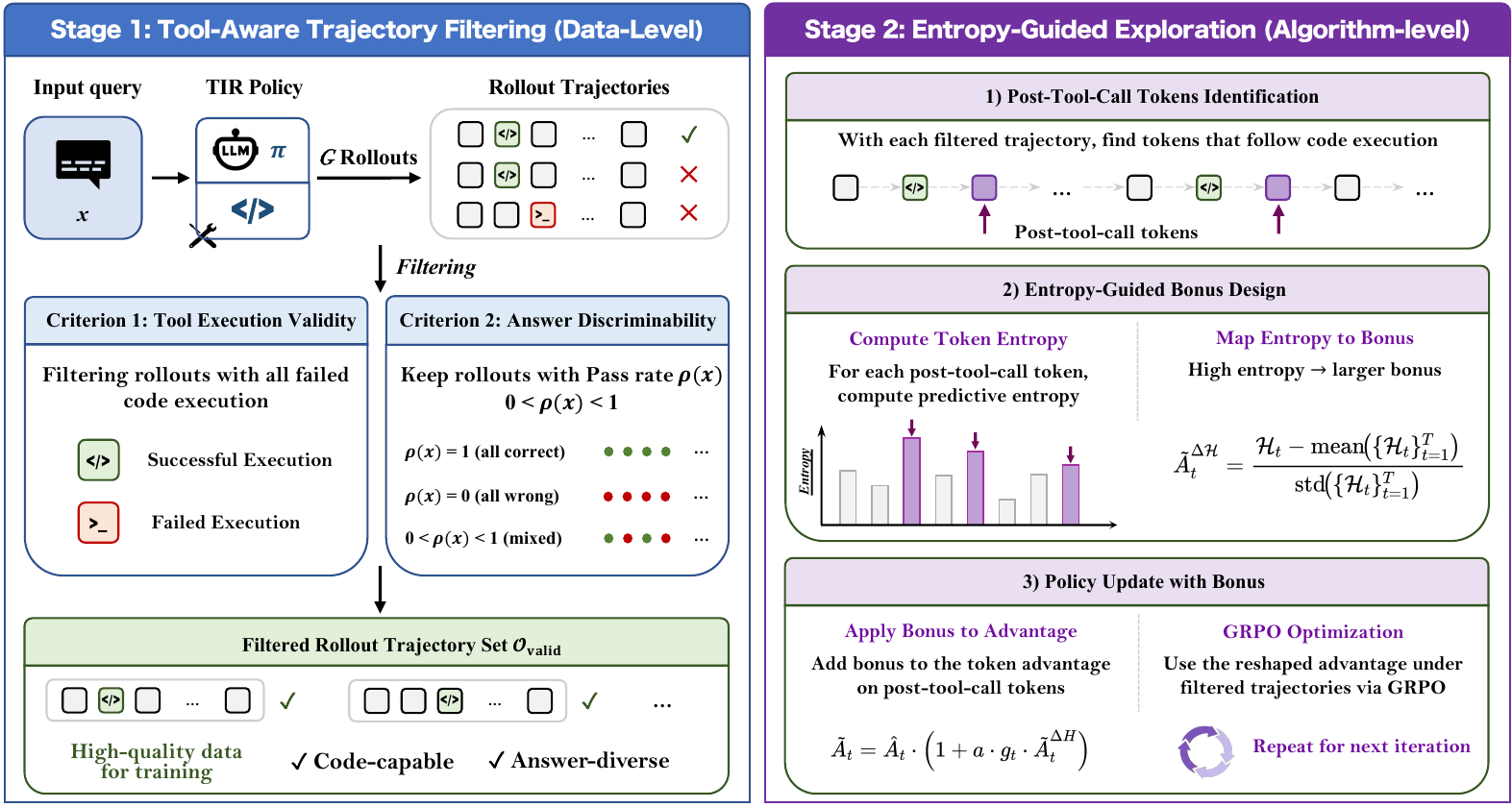}
\caption{Overall framework of \method with two components for policy optimization: (a) Tool-aware trajectory filtering based on two criteria in the data-level. (b) Entropy-guided exploration at post-tool-call tokens in the algorithm-level.}
\label{fig:framework}
\end{figure*}

\paragraph{Agentic RL}
We formalize the agentic RL setting. Unlike standard LLM policy optimization over pure text generation, agentic RL interleaves language generation with tool interaction. In this work, we focus on empowering the model to invoke the \emph{code interpreter} during the reasoning process to execute programs and obtain runtime feedback. Let $x \sim \mathcal{D}$ denote an input query and $\pi_\theta$ as the policy model. In this setting, the overall response $o$ consists of two coupled components: an agentic reasoning trajectory $\mathcal{R} = \{s_t\}_{t=1}^{|\mathcal{R}|}$, which includes generated reasoning steps, code generation blocks, and returned execution results, and a final answer $y = \{y_t\}_{t=1}^{|y|}$ conditioned on the trajectory. The joint generation process of $o = (\mathcal{R}, y)$ is defined as:
$$
P_\theta(\mathcal{R}, y \mid x) =
\underbrace{ \prod_{t=1}^{t_\mathcal{R}} P_\theta\!\left(\mathcal{R}_t \mid \mathcal{R}_{<t}, x\right)
}_{\text{Agentic Reasoning} }
\cdot
\underbrace{ \prod_{t=1}^{t_y} P_\theta\!\left(y_t \mid y_{<t}, \mathcal{R}, x\right)}_{\text{Answer Generation}},
$$
where $\mathcal{R}$ is the reasoning trajectory of length $t_\mathcal{R}$, and $y$ is the final answer with length $t_y$.
Following GRPO-based optimization, the policy is trained to maximize the expected reward of the overall trajectory and final answer:
$$
\max_{\theta}\;\mathbb{E}_{x \sim \mathcal{D},\,(\mathcal{R},y) \sim \pi_\theta(\cdot \mid x)}\left[r(x, \mathcal{R}, y)\right].
$$

\section{Approach}
\label{sec:approach}

As illustrated in Fig.~\ref{fig:framework}, \method consists of two core components: tool-aware trajectory filtering and an entropy-guided exploration mechanism. First, we generate model rollouts across the training dataset and apply two complementary filtering criteria to curate a high-quality training distribution. Following this, we execute token-level optimization by identifying high-entropy tokens following tool executions and applying an entropy-guided bonus to encourage exploration at these critical decision points.

\subsection{Tool-Aware Trajectory Filtering}
\label{sec:filtering}

\emph{Not all rollout trajectories provide valid or informative learning signals for training .} Trajectories where tool interactions consistently fail inject only noise into training, while those where the model either answers correctly or incorrectly on every rollout yield degenerate advantage estimates that contribute no discriminative gradient information. \method filters rollout trajectories along two complementary criteria, retaining only those that are both tool-capable and informative contrastive signals.

For each query $x$, we sample $G$ rollout trajectories $\{o^i\}_{i=1}^G \sim \pi_{\theta_{\text{old}}}(\cdot \mid x)$ with corresponding binary rewards $\{r^i\}_{i=1}^G \in \{0,1\}^G$. We then apply the following two criteria.

\textit{\textbf{Criterion 1: Tool Execution Validity.}} In agentic RL, the value of a trajectory depends not only on whether the final answer is correct, but also on whether the intermediate tool interactions provide meaningful environmental feedback. A trajectory that produces an incorrect answer while all of its tool invocations fail with execution errors is doubly uninformative: it neither guides the policy toward correct reasoning nor provides valid grounding from external tool outputs. The entire reasoning chain in such a trajectory is constructed without reliable environmental signal, making it indistinguishable from pure hallucination and harmful to stable policy learning. Including such trajectories in training risks reinforcing erroneous reasoning patterns that happen to invoke tools in syntactically valid but semantically meaningless ways. We therefore discard trajectory $o^i$ if:
\begin{equation*}
    r^i = 0 \;\wedge\; \mathcal{S}_{\text{tool}}(o^i) = 0,
\end{equation*}
where $\mathcal{S}_{\text{tool}}(o^i)$ denotes the number of successful tool invocations in trajectory $o^i$. This criterion specifically targets the intersection of answer incorrectness and tool failure, preserving incorrect trajectories that nonetheless contain at least one successful tool execution, as these still carry informative signals about how tool feedback shapes reasoning.

\textit{\textbf{Criterion 2: Answer Discriminability.}} Beyond individual trajectory quality, the collective behavior of rollouts for a given query determines whether meaningful advantage estimates can be derived. We define the empirical pass rate as:
\begin{equation}
    \rho(q) = \frac{1}{G}\sum_{i=1}^{G} r^i,
\label{eq:pass_rate}
\end{equation}
and analyze the two degenerate extremes. When $\rho(q) = 1$, all rollouts produce correct answers, indicating that the current policy has already mastered the task. In this case, all trajectories receive identical rewards, the group-normalized advantage estimates collapse to zero, and no gradient signal is propagated that training on such queries is entirely wasteful. When $\rho(q) = 0$, all rollouts produce incorrect answers, meaning the policy receives uniformly zero reward across the group. Without any positive reward signal to contrast against, the policy has no direction from which to improve, and the resulting advantage estimates again carry no useful information for policy improvement. Both extremes thus represent a failure of the contrastive learning mechanism that underpins GRPO-style optimization. We retain only queries satisfying:
\begin{equation*}
    0 < \rho(q) < 1,
\end{equation*}
ensuring that each retained query contains at least one correct and one incorrect rollout, from which informative contrastive advantage estimates can be reliably derived.

Let $\mathcal{O}_{\text{valid}}$ denote the set of trajectories surviving both criteria. Unlike SimpleTIR~\cite{xue2026simpletir}, which filters solely on format validity, \method additionally removes tool-execution-failed incorrect trajectories and discriminability-degenerate queries, yielding a training distribution that is both tool-capable and contrastively informative for stable policy optimization.

\begin{figure}[t]
  \centering
    \includegraphics[width=0.49\linewidth]{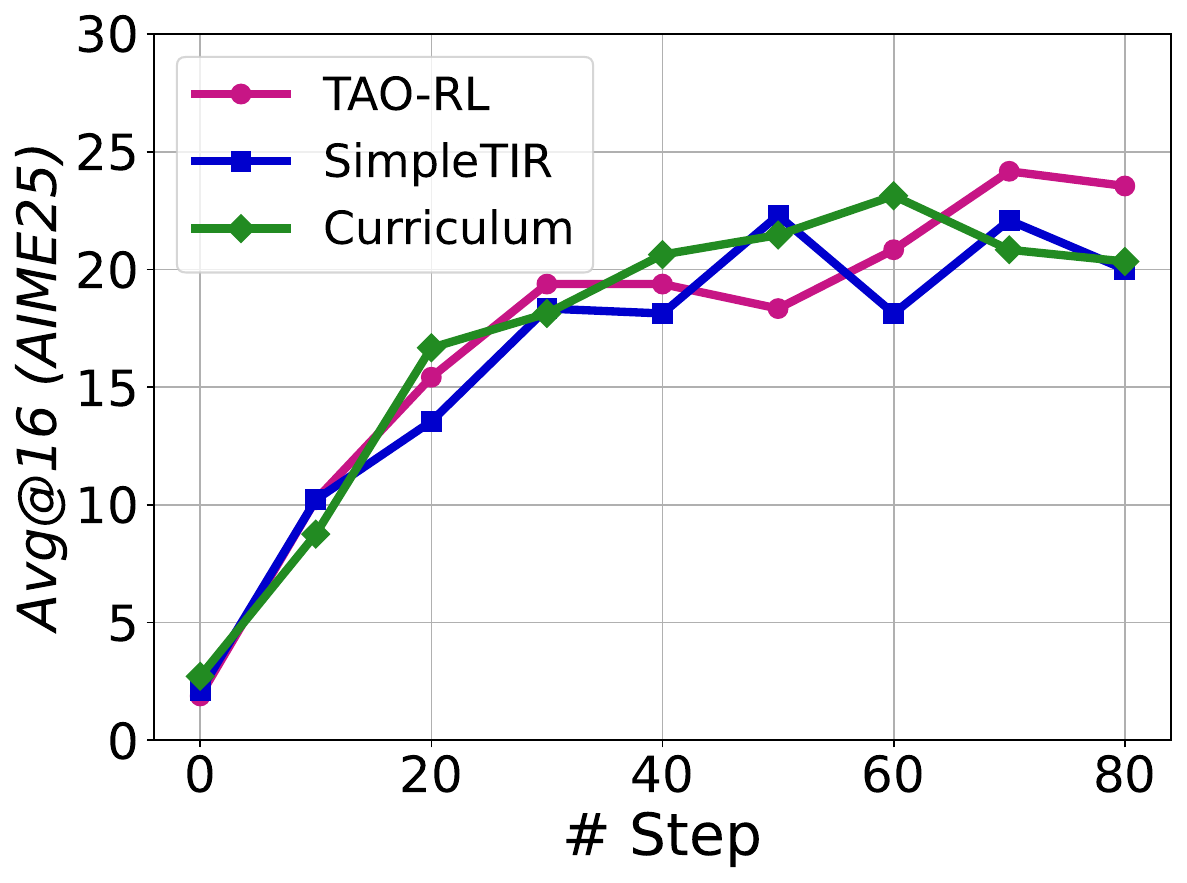} 
  \includegraphics[width=0.49\linewidth]{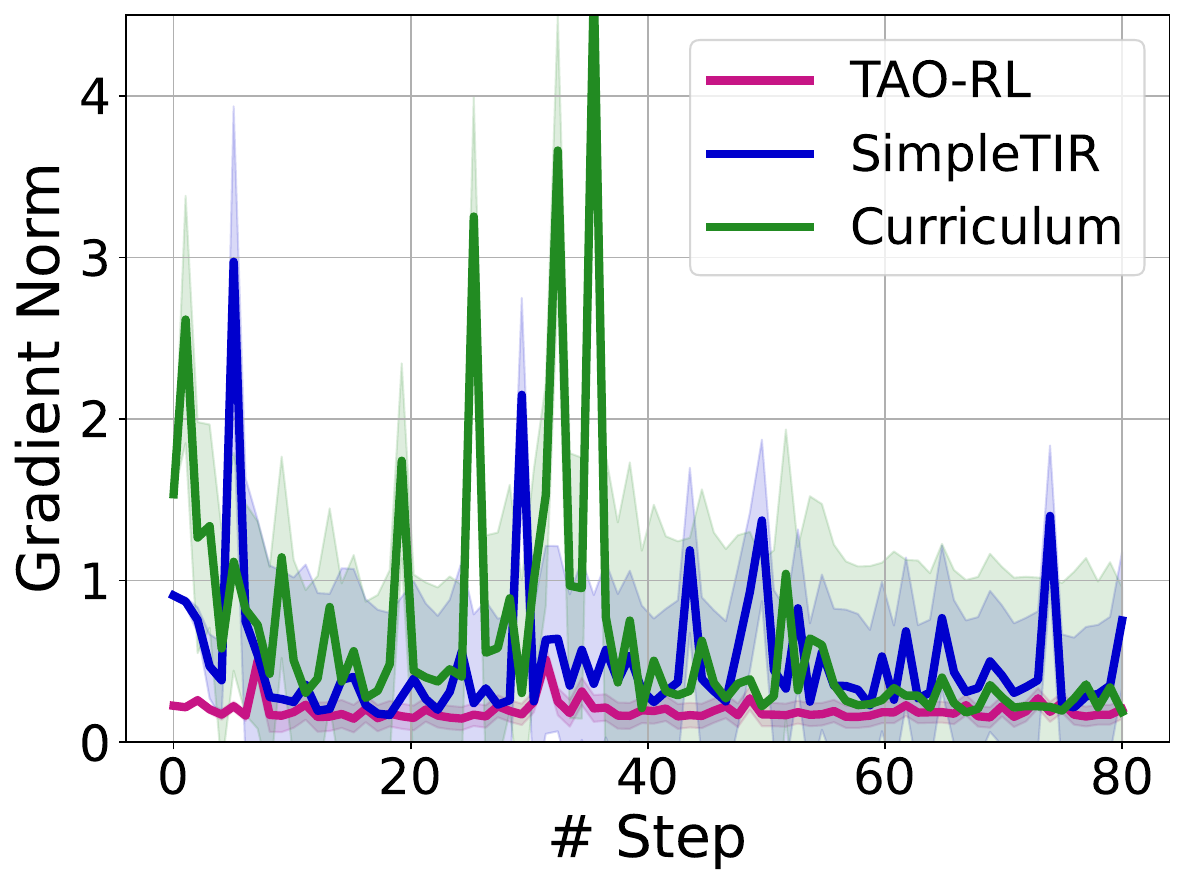} 
  \caption{Comparison of SimpleTIR, curriculum and \method across \textit{Avg@16} in AIME 25, and gradient norm under Qwen2.5-7B base model during training. }
  \label{fig:data_analyze}
  \vspace{-1mm}
\end{figure}

\begin{tcolorbox}[
    colback=gray!8,
    colframe=gray!50,
    boxrule=0.8pt,
    arc=4pt,
    left=6pt, right=6pt, top=6pt, bottom=6pt
]
\textbf{Empirical Analysis.} To investigate the impact of different data processing strategies on training performance, we compare \method against two baselines. The first is SimpleTIR, which only filters trajectories based on format validity. The second is a curriculum-based variant, where instead of discarding degenerate queries via pass rate filtering, we organize all training data in a progressive easy-to-hard order based on task difficulty, allowing us to directly assess whether difficulty-aware data organization is a viable alternative to our answer discriminability filtering. As shown in Fig.~\ref{fig:data_analyze}, \method achieves the best average accuracy on AIME25, outperforming both baselines. In terms of training stability, \method exhibits the smallest gradient norm fluctuations during training, while both baselines suffer from larger oscillations. These results demonstrate that neither format-based filtering alone nor progressive difficulty ordering can adequately handle the degenerate training signals introduced by tool interaction, validating the effectiveness of our joint tool execution validity and answer discriminability filtering strategy. We conduct further experimental analysis of data organization in section~\ref{sec:property}, and Appendix D. 

\end{tcolorbox}

\vspace{-0.5mm}
\subsection{Entropy-Guided Exploration}
\label{sec:token_exploration}

While tool-aware trajectory filtering establishes a clean and informative training foundation, it does not explicitly encourage exploration within the retained trajectories. Existing methods address exploration through uniform entropy regularization across all token positions, which is ill-suited to agentic settings: it introduces unnecessary noise at inconsequential positions while failing to concentrate sufficient incentives at the critical junctures where tool execution results are interpreted. Our key insight is that \emph{exploration should be concentrated at tokens following successful tool executions, where predictive uncertainty is highest and the downstream consequence of the policy's choice is critical.} To this end, we introduce an entropy-guided bonus that reshapes the advantage function at these critical positions.

\paragraph{Post-Tool-Call Token Identification}

Let $\mathcal{M}_{\text{post}}$ denote the set of all token positions that follow tool execution within the filtered trajectories $\mathcal{O}_{\text{valid}}$ across current rollout group. For each token $o_t \in \mathcal{M}_{\text{post}}$, we compute its predictive entropy:
\begin{equation*}
    \mathcal{H}_t = -\sum_{v} \pi_\theta(v \mid x, o_{<t}) \cdot \log \pi_\theta(v \mid x, o_{<t}).
\end{equation*}

We retain the top-$q_{\mathcal{H}}$ proportion of post-tool-call tokens ranked by predictive entropy for encouraging exploration, and define a binary gate accordingly:
\begin{equation*}
    g_t = \mathbf{1}\left[ t \in \mathcal{M}_{\text{post}} \;\wedge\; \mathcal{H}_t \geq \tau_{\mathcal{H}} \right],
\end{equation*}
where $\tau_{\mathcal{H}}$ is the entropy value at the $(1-q_{\mathcal{H}})$-th percentile of $\{\mathcal{H}_t\}_{t \in \mathcal{M}_{\text{post}}}$. This selective gating concentrates the entropy bonus exclusively on the most uncertain post-tool-call positions, avoiding unnecessary perturbation at positions where the policy is already sufficiently confident.

\paragraph{Entropy-Aware Advantage Reshaping}

For the gated tokens, we introduce a normalized entropy-based advantage term to quantify how much each token's uncertainty deviates from the trajectory mean:
\begin{equation}
    \tilde{A}_{t}^{\Delta \mathcal{H}} = \frac{\mathcal{H}_t - \mathrm{mean}\!\left(\{\mathcal{H}_t\}_{t=1}^{T}\right)}{\mathrm{std}\!\left(\{\mathcal{H}_t\}_{t=1}^{T}\right)},
\label{eq:entropy_adv}
\end{equation}
where $T$ is the total number of tokens across all trajectories in the group. This term is estimated per token based on the average token entropy within the same micro-batch. We treat it as a regularization bonus and incorporate it into the advantage defined in Eq.~\ref{eq:adv}, yielding the reshaped advantage:
\begin{equation}
    \tilde{A}_{t} = \hat{A}_{t} \cdot \left(1 + a \cdot g_t \cdot \tilde{A}_{t}^{\Delta \mathcal{H}}\right),
\label{eq:reshaped_adv}
\end{equation}
where $a$ is a scaling coefficient controlling the strength of the entropy bonus. The gate $g_t$ ensures the reshaping is applied exclusively to post-tool-call tokens, preserving the original advantage estimates at all other positions. The reshaped $\tilde{A}_{t}$ is computed before the policy update and substituted into the standard clipped surrogate objective of GRPO.
Integrating the two components of \method, the final training objective is: 
\begin{equation}
\label{eq:final_objective}
\hspace{-2mm}
\begin{aligned}
& \mathcal{J}_{\method}(\theta)
= \mathbb{E}_{(q,a)\sim \mathcal{D},\,\{o^i\}_{i=1}^G \sim \pi_{\theta_{\mathrm{old}}}(\cdot | q)} \\
& \quad \Bigg[ \frac{1}{|\mathcal{O}_{\text{valid}}|} \sum_{i \in \mathcal{O}_{\text{valid}}}  \frac{1}{|o^i|}
\sum_{t=1}^{|o^i|}
\Big( \min\!\big(r^i_t(\theta)\tilde{A}^i_{t},\, \mathrm{clip}(r^i_t(\theta), \\
& \quad 1-\epsilon,\, 1+\epsilon)\tilde{A}^i_{t}\big)
- \beta D_{\mathrm{KL}}(\pi_{\theta} \,\|\, \pi_{\mathrm{ref}}) \Big)
\Bigg],
\end{aligned}
\end{equation}
where $\mathcal{O}_{\text{valid}} \subseteq \{o^i\}_{i=1}^G$ denotes the filtered trajectories passing both criteria. Trajectories outside $\mathcal{O}_{\text{valid}}$ are excluded from the policy update, ensuring that gradient updates are computed exclusively over tool-capable rollouts with discriminative learning signals. For tokens outside $\mathcal{M}_{\text{post}}$, $\tilde{A}^i_t$ reduces to the standard advantage $\hat{A}^i_t$, so \method introduces no additional computational overhead beyond the entropy computation at post-tool-call positions.

The two components are mutually dependent for achieving stable exploration. Without trajectory filtering, the entropy bonus would amplify gradients indiscriminately over noisy degenerate rollouts, causing unstable policy updates. Without the entropy bonus, the filtered gradients would treat all token positions equally, failing to concentrate exploration at the most consequential decision points. \method achieves targeted exploration without sacrificing training stability.

\begin{tcolorbox}[
    colback=gray!8,
    colframe=gray!50,
    boxrule=0.8pt,
    arc=4pt,
    left=6pt, right=6pt, top=6pt, bottom=6pt
]
\textbf{Theoretical Analysis.}
We formally analyze how the two components of \method jointly improve the optimization landscape via gradient signal-to-noise ratio (SNR) and targeted step-size modulation. The policy gradient over the filtered trajectory set $\mathcal{O}_{\text{valid}}$ with reshaped advantage:
\begin{equation*}
\nabla\mathcal{J}_{\method}(\theta) = \mathbb{E}_{o^i\in\mathcal{O}_{\text{valid}}}\left[\sum_{t=1}^{|o^i|}\nabla_\theta\log\pi_\theta(o^i_t|s_t)\cdot\tilde{A}_t\right].
\label{eq:grad_method}
\end{equation*}

\quad \emph{SNR Maximization via Trajectory Filtering:} Standard GRPO computes gradients over all sampled trajectories. Degenerate queries with $\rho(q)\in\{0,1\}$ yield zero advantage estimates ($\hat{A}_t = 0$), which waste compute and dilute the expected gradient signal. Furthermore, trajectories composed entirely of failed tool calls lack valid environmental grounding, meaning their token updates are driven by pseudo-random advantage fluctuations, injecting zero-mean, high-variance noise into the batch. By restricting the expectation to $\mathcal{O}_{\text{valid}}$, \method removes both signal diluters and noise injectors, fundamentally improving the SNR of the policy gradient:
\begin{equation*}
\frac{\|\mathbb{E}[\nabla\mathcal{J}_{\method}]\|^2}{\mathrm{Var}[\nabla\mathcal{J}_{\method}]} > \frac{\|\mathbb{E}[\nabla\mathcal{J}_{\text{GRPO}}]\|^2}{\mathrm{Var}[\nabla\mathcal{J}_{\text{GRPO}}]}.
\end{equation*}

\quad \emph{Targeted Exploration via Entropy Bonus:} within the stabilized gradient, the reshaped advantage further modulates the logit update~\cite{NIPS2001_4b86abe4,cui2025entropy} at post-tool-call positions:

\begin{equation*}
\begin{aligned}
\theta^{k+1}_{s_t,o_t} - \theta^k_{s_t,o_t}
&= \underbrace{\eta\cdot\pi^k_\theta(o_t|s_t)\cdot\hat{A}_t}_{\text{basic update}} \\
&\quad + \underbrace{\eta\cdot a\cdot g_t\cdot\pi^k_\theta(o_t|s_t)\cdot\hat{A}_t\cdot\tilde{A}_t^{\Delta\mathcal{H}}}_{\text{exploration bonus}},
\end{aligned}
\end{equation*}

where $\eta$ is the learning rate. The gate $g_t$ ensures the exploration bonus is applied exclusively to the top-$q_{\mathcal{H}}$ highest-entropy post-tool-call tokens. For these gated tokens, the bonus amplifies the logit update in proportion to their entropy deviation, concentrating targeted exploration precisely at the most uncertain positions where the downstream consequence of each action is greatest. For all other positions, $g_t=0$ and the update reduces to the standard basic update, preserving stable optimization behavior throughout the remainder of the trajectory. We provide detailed theoretical derivation in Appendix B.

\end{tcolorbox}

\subsection{Practical Implementation}

\begin{algorithm}[t]
\caption{\method}
\label{alg:method}
\begin{algorithmic}[1]
\STATE \textbf{Input:} Training dataset $\mathcal{D}$, initial policy $\pi_\theta$, sampling number $G$, entropy bonus coefficient $a$
\STATE \textbf{Output:} Optimized policy $\pi_\theta$
\FOR{each training epoch}
    \FOR{each batch $\mathcal{B} \subseteq \mathcal{D}$}
        \item[] {\color{darkred} \textbf{// Phase 1: Tool-Aware Trajectory Filtering}}
        \STATE Sample $G$ rollouts $\{o^i\}_{i=1}^G \sim \pi_{\theta_{\text{old}}}(\cdot \mid q)$ for each $q \in \mathcal{B}$
        \STATE Compute pass rate $\rho(q) = \frac{1}{G}\sum_{i=1}^{G} r^i$ via Eq.~\ref{eq:pass_rate}
        \STATE Apply Criterion 1: $\mathcal{O}' \leftarrow \{ o^i \mid \neg(r^i = 0 \wedge \mathcal{S}_{\text{tool}}(o^i) = 0) \}$
        \STATE Apply Criterion 2: $\mathcal{O}_{\text{valid}} \leftarrow \mathcal{O}'$ if $0 < \rho(q) < 1$, else $\mathcal{O}_{\text{valid}} \leftarrow \emptyset$
        \item[] {\color{darkred} \textbf{// Phase 2: Entropy-Guided Exploration}}
        \STATE Compute advantages $\{\hat{A}^i_t\}$ for each $o^i \in \mathcal{O}_{\text{valid}}$
        \STATE Identify post-tool-call token positions $\mathcal{M}_{\text{post}}$
        \STATE Compute token entropy $\mathcal{H}_t$ for each $t \in \mathcal{M}_{\text{post}}$
        \STATE Compute normalized entropy advantage bonus $\tilde{A}^{\Delta\mathcal{H}}_t$ via Eq.~\ref{eq:entropy_adv}
        \STATE Reshape advantage $\tilde{A}^i_t \leftarrow \hat{A}^i_t \cdot (1 + a \cdot g_t \cdot \tilde{A}^{\Delta\mathcal{H}}_t)$ via Eq.~\ref{eq:reshaped_adv}
        \STATE Update policy $\pi_{\theta}$ by maximizing $\mathcal{J}_{\method}(\theta)$ over $\mathcal{O}_{\text{valid}}$ via Eq.~\ref{eq:final_objective}
    \ENDFOR
\ENDFOR
\end{algorithmic}
\end{algorithm}

Algorithm~\ref{alg:method} presents the complete training procedure of \method. For each training batch, the procedure executes two sequential phases before each policy update.

In \textbf{Phase 1} (lines 5-8), for each query $q \in \mathcal{B}$, the policy samples $G$ rollout trajectories and applies two filtering criteria to construct $\mathcal{O}_{\text{valid}}$. First, trajectories that are simultaneously incorrect and have all tool invocations failing are discarded, as they provide neither a correct reasoning signal nor valid environmental feedback. Second, the empirical pass rate $\rho(q)$ is computed, and queries at the degenerate extremes where all rollouts are correct ($\rho(q)=1$) or all are incorrect ($\rho(q)=0$) are removed, as both cases yield uninformative advantage estimates. Only trajectories in $\mathcal{O}_{\text{valid}}$ proceed to optimization, ensuring policy gradients are computed exclusively over tool-capable rollouts with contrastive learning signals.

In \textbf{Phase 2} (lines 9-14), advantages $\{\hat{A}^i_t\}$ are computed over the valid trajectories in $\mathcal{O}_{\text{valid}}$. Post-tool-call token positions $\mathcal{M}_{\text{post}}$ are then identified, and for each such position the token entropy $\mathcal{H}_t$ and normalized entropy advantage $\tilde{A}^{\Delta\mathcal{H}}_t$ are computed. The advantage is reshaped via Eq.~\ref{eq:reshaped_adv}, amplifying the learning signal at post-tool-call tokens in proportion to their entropy deviation and naturally concentrating exploration incentives at the most uncertain decision points, while leaving all other token positions unaffected. The policy is finally updated by maximizing $\mathcal{J}_{\method}(\theta)$ over $\mathcal{O}_{\text{valid}}$ via Eq.~\ref{eq:final_objective}.
This two-phase per-batch design ensures that every policy update is grounded in high-quality, tool-capable trajectories carrying discriminative contrastive signals.

\section{Experiment}
\label{sec:experiments}

Our experiments are designed to answer the following questions: (i) How does \method compare against existing agentic RL approaches across diverse reasoning tasks? (ii) Can tool-aware trajectory filtering and entropy-guided exploration jointly improve LLM reasoning performance? (iii) How do individual components and hyperparameters contribute to the overall effectiveness of \method? (iv) How well does \method generalize to out-of-distribution tasks? (v) What insights can be drawn from case studies of specific reasoning trajectories?

\subsection{Experimental Setup}
\subsubsection{Benchmarks and Implementation Details} To comprehensively validate the effectiveness of \method across different model scales and data regimes, we conduct training under two settings: we train Qwen2.5-1.5B on the Math3-5 dataset from SimpleRL~\cite{zeng2025simplerl}, and scale up to Qwen3-4B-Base and Qwen2.5-7B trained on the larger DAPO-MATH-17K dataset~\cite{yu2025dapo}, ensuring thorough evaluation across both model capacity and training data diversity. The rollout size $M$ is $16$, temperature factor is $1.0$, and learning rate $\eta$ is $1e-6$. For fair comparisons, we reproduce all baselines and conduct training under the same hyperparameters in the VeRL platform~\cite{sheng2025hybridflow}. We select GRPO as our basic post-training algorithm. We conduct extensive validation on seven benchmarks, including AIME 2025\&2024~\cite{codeforcesamerican}, AMC 2023~\cite{amc23}, MATH500~\cite{hendrycks2measuring}, OlympiadBench~\cite{he2024olympiadbench}, Hmmt Feb-25~\cite{balunovic_srimatharena_2025}, and Minerva~\cite{lewkowycz2022solving}. 
We compare against TIR, SimpleTIR~\cite{xue2026simpletir}, and AEPO~\cite{dong2025agentic} as our baselines. More details of the experimental setup and benchmarks are provided in Appendix C.

\subsubsection{Metrics} We assess the reasoning ability boundaries using the \underline{\textit{Pass@K}}: represents at least one of $K$ sampled model outputs passes verification, and assess the average performance using the \underline{\textit{Avg@K}}: denotes the average accuracy over $K$ evaluations, \underline{\textit{Len@K}}: average response length over $K$ evaluations per benchmark.
We vary $K$ across experiments to ensure the statistical reliability of our evaluation results. For agentic behavior, we introduce two complementary metrics: \underline{\textit{Tool\_Call}}, which quantifies the average number of tool invocations per response, capturing the degree to which the model leverages external computation; and \underline{\textit{Valid\_Code}}, which measures the proportion of tool calls that produce syntactically correct and executable code, reflecting the quality of the model's code generation and tool-use proficiency.
\underline{\textit{Box\_Rate}} measures the proportion of responses that correctly follow the required answer format. \underline{\textit{Code\_Line}} measures the number of lines of code generated per rollout.

\begin{table*}[t]
\caption{Performance comparison on seven datasets under Qwen2.5-1.5B, and Qwen3-4B-Base base models. "Avg@k" denotes the average accuracy over $K$ evaluations per benchmark. "Pass@k" represents at least one of $K$ sampled model outputs passes verification. We bold the best scores and underline the sub-optimal results. $\Delta$ means the difference between \method and sub-optimal results.}
\renewcommand{\arraystretch}{1.4}
\setlength{\tabcolsep}{2.2pt}
\centering
\begin{tabular}{ccccccccccccccc}
\toprule
\multicolumn{1}{c}{\textbf{Method}} & \multicolumn{2}{c}{\textbf{AIME24}}         & \multicolumn{2}{c}{\textbf{AIME25}}         & \multicolumn{2}{c}{\textbf{MATH500}}        & \multicolumn{2}{c}{\textbf{Olympiad}}       & \multicolumn{2}{c}{\textbf{AMC23}}          & \multicolumn{2}{c}{\textbf{Hmmt25}}         &
 \multicolumn{2}{c}{\textbf{Minerva}} \\ 
\cmidrule(lr){2-3} \cmidrule(lr){4-5} \cmidrule(lr){6-7} \cmidrule(lr){8-9} \cmidrule(lr){10-11} \cmidrule(lr){12-13}
\cmidrule(lr){10-11} \cmidrule(lr){14-15}
                           & \textit{Avg@16} & \textit{Pass@16} & \textit{Avg@16} & \textit{Pass@16} & \textit{Avg@16} & \textit{Pass@16} & \textit{Avg@16} & \textit{Pass@16} & \textit{Avg@16} & \textit{Pass@16} & \textit{Avg@16} & \textit{Pass@16} & \textit{Avg@16} & \textit{Pass@16} \\ 
                           \bottomrule
\rowcolor{black!10} \multicolumn{15}{c}{\textit{Qwen2.5-1.5B}}           \\ \hline
\textbf{Base model}
&      0.21       &      3.33            
&      0.21       &      3.33         
&      3.11       &      35.80                  
&      1.48       &      15.40               
&      1.63       &      18.00            
&      0.21       &      \underline{3.33 }
&      0.76       &      8.09       \\
\textbf{SimpleTIR}
&      \underline{8.75}      &        \underline{33.33}
&      \underline{11.46}     &        \textbf{33.33} 
&      \underline{63.24}     &        \underline{87.60}
&      \textbf{30.48}        &        \underline{57.00}
&      \underline{40.00}     &        \textbf{90.00} 
&      \underline{10.83}     &        \textbf{23.33}          
&      \underline{10.04 }    &        \underline{28.31 }        \\
\rowcolor{blue!10}  \method     
&      \textbf{11.88}      &       \textbf{50.00} 
&      \textbf{13.33}      &       \underline{30.00}
&      \textbf{63.51}      &       \textbf{90.60}
&      \underline{30.21}   &       \textbf{58.00} 
&      \textbf{40.63}      &       \underline{87.50}
&      \textbf{13.54}      &       \textbf{23.33}
&      \textbf{10.55}      &       \textbf{29.04}        \\ 
$\Delta$ 
& \color{darkgreen}{+3.13}
& \color{darkgreen}{+16.67}
& \color{darkgreen}{+1.87}
& \color{darkred}{-3.33}
& \color{darkgreen}{+0.27}
& \color{darkgreen}{+3.00}
& \color{darkred}{-0.27}
& \color{darkgreen}{+1.00}
& \color{darkgreen}{+0.63}
& \color{darkred}{-2.50}
& \color{darkgreen}{+2.71}
& \color{darkgreen}{+20.00}
& \color{darkgreen}{+0.52}
& \color{darkgreen}{+0.73}
\\
\bottomrule
\rowcolor{black!10}
\multicolumn{15}{c}{\textit{Qwen3-4B-Base}}                                                       \\ \hline
\textbf{Base model}
&      6.25       &       40.00      
&      6.67       &       \underline{33.33}    
&      40.78      &       90.60 
&      22.54      &       65.40   
&      27.66      &       90.00     
&      \underline{7.08}       &       \underline{26.67}    
&      8.87       &       33.09          \\
\textbf{SimpleTIR}
&      \underline{27.50}      &       \underline{56.67}
&      \underline{23.54}      &       \textbf{53.33}   
&      \underline{81.13}      &       \underline{95.00}
&      \underline{46.71}      &       \underline{72.80}
&      \underline{64.53}      &       \textbf{97.50}
&      \textbf{19.79}         &       \textbf{43.33}
&      \underline{23.81}      &       \underline{40.44}   \\
\rowcolor{blue!10} \method
&      \textbf{28.13}      &       \textbf{63.33}
&      \textbf{24.71}      &       \textbf{53.33}
&      \textbf{83.74}      &       \textbf{96.20}
&      \textbf{47.64}      &       \textbf{73.20}
&      \textbf{65.47}      &       \underline{95.00}
&      \textbf{19.79}      &       \textbf{43.33}
&      \textbf{24.10}      &       \textbf{42.65}          \\
$\Delta$
& \color{darkgreen}{+0.63}
& \color{darkgreen}{+6.66}
& \color{darkgreen}{+1.17}
& \color{darkgreen}{+20.00}
& \color{darkgreen}{+2.61}
& \color{darkgreen}{+1.20}
& \color{darkgreen}{+0.93}
& \color{darkgreen}{+0.40}
& \color{darkgreen}{+0.94}
& \color{darkred}{-2.50}
& \color{darkgreen}{+12.71}
& \color{darkgreen}{+16.66}
& \color{darkgreen}{+0.29}
& \color{darkgreen}{+2.21}
\\
\bottomrule          
\end{tabular}
\label{tab:exp_1.5b}
\end{table*}

\begin{table*}[t]
\caption{Performance comparison on seven datasets under Qwen2.5-7B base model. "Avg@k" denotes the average accuracy over $K$ evaluations per benchmark. "Pass@k" represents at least one of $K$ sampled model outputs passes verification. We bold the best scores and underline the sub-optimal results. $\Delta$ means the difference between \method and sub-optimal results.}
\renewcommand{\arraystretch}{1.4}
\setlength{\tabcolsep}{2.2pt}
\centering
\begin{tabular}{ccccccccccccccc}
\toprule
\multicolumn{1}{c}{\textbf{Method}} & \multicolumn{2}{c}{\textbf{AIME24}}         & \multicolumn{2}{c}{\textbf{AIME25}}         & \multicolumn{2}{c}{\textbf{MATH500}}        & \multicolumn{2}{c}{\textbf{Olympiad}}       & \multicolumn{2}{c}{\textbf{AMC23}}          & \multicolumn{2}{c}{\textbf{Hmmt25}}         &
 \multicolumn{2}{c}{\textbf{Minerva}} \\ 
\cmidrule(lr){2-3} \cmidrule(lr){4-5} \cmidrule(lr){6-7} \cmidrule(lr){8-9} \cmidrule(lr){10-11} \cmidrule(lr){12-13}
\cmidrule(lr){10-11} \cmidrule(lr){14-15}
                           & \textit{Avg@16} & \textit{Pass@16} & \textit{Avg@16} & \textit{Pass@16} & \textit{Avg@16} & \textit{Pass@16} & \textit{Avg@16} & \textit{Pass@16} & \textit{Avg@16} & \textit{Pass@16} & \textit{Avg@16} & \textit{Pass@16} & \textit{Avg@16} & \textit{Pass@16} \\ 
                           \bottomrule
\rowcolor{black!10}
\multicolumn{15}{c}{\textit{Qwen2.5-7B}}      \\ \hline
\textbf{Base model}
&       5.00         &        36.67
&       3.96         &    \underline{26.67}
&       31.99        &        88.00
&       13.80        &        55.00
&       21.56        &        80.00           
&       6.67         &        26.67 
&       5.00         &        36.67           \\
\textbf{TIR}
&       10.63        &        40.00
&       10.42        &        20.00
&       42.11        &        79.80
&       20.88        &        49.80
&       28.91        &        70.00           
&       11.04        &        23.33
&       7.95         &        30.51          \\
\textbf{AEPO}
&       16.67        &        50.00
&       10.21        &        33.33
&       57.20        &        92.40
&       29.35        &        63.60
&       41.25        &        85.00    
&       14.17        &        \underline{40.00}
&       13.79        &        38.24          \\
\textbf{SimpleTIR}
&   \underline{33.33}  &    \underline{66.67}
&   \underline{21.04}  &    \textbf{46.67}
&   \underline{77.51}  &    \underline{93.40}
&   \underline{42.95}  &    \underline{66.60}
&   \underline{68.75}  &    \underline{92.50}     
&   \underline{16.66}  &        36.67 
&   \underline{21.23}  &    \underline{39.34}           \\
\rowcolor{blue!10} \method
&   \textbf{35.42}   &    \textbf{73.33}
&   \textbf{23.96}   &    \textbf{46.67}
&   \textbf{78.80}   &    \textbf{95.60}
&   \textbf{46.99}   &    \textbf{70.00}
&   \textbf{69.06}   &    \textbf{97.50}         
&   \textbf{18.33}   &    \textbf{46.67} 
&   \textbf{23.12}   &    \textbf{40.81}           \\
$\Delta$             
& \color{darkgreen}{+2.09}
& \color{darkgreen}{+6.66}
& \color{darkgreen}{+2.92}
& \color{darkgreen}{+20.00}
& \color{darkgreen}{+1.29}
& \color{darkgreen}{+2.20}
& \color{darkgreen}{+4.04}
& \color{darkgreen}{+3.40}
& \color{darkgreen}{+0.31}
& \color{darkgreen}{+5.00}
& \color{darkgreen}{+1.67}   
& \color{darkgreen}{+6.67}
& \color{darkgreen}{+1.89}
& \color{darkgreen}{+1.47}               \\
\bottomrule          
\end{tabular}
\label{tab:exp_7b}
\end{table*}

\begin{figure*}[t]
\centering
\includegraphics[width=0.24\linewidth]{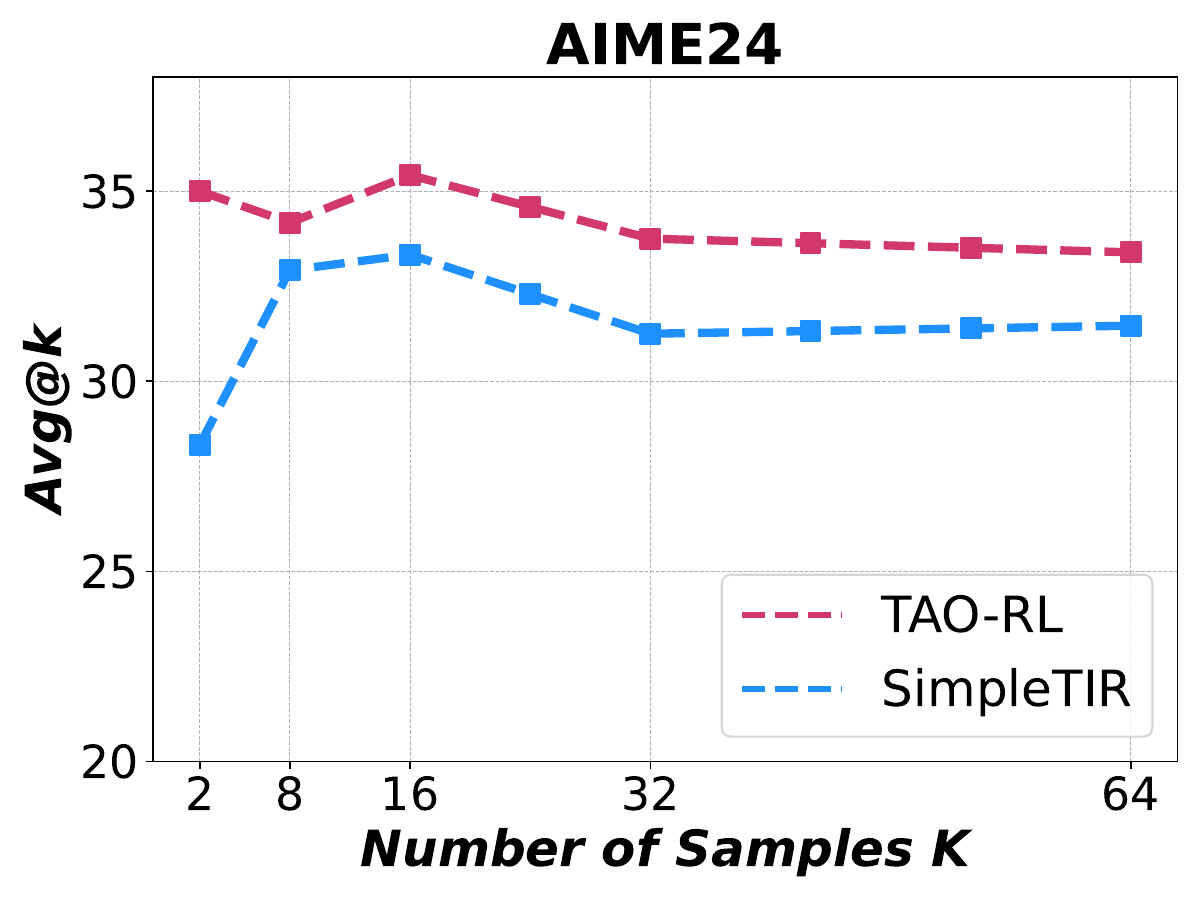}
\includegraphics[width=0.24\linewidth]{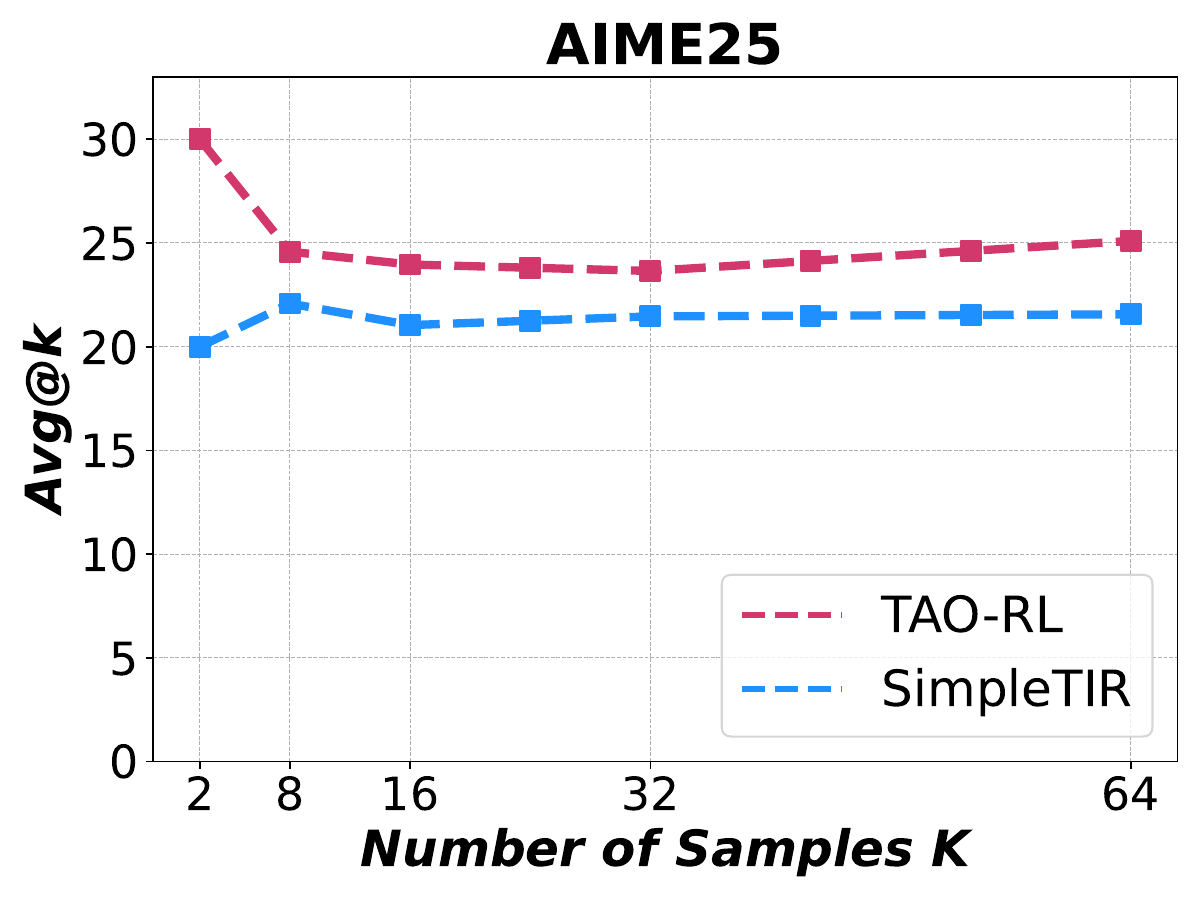}
\includegraphics[width=0.24\linewidth]{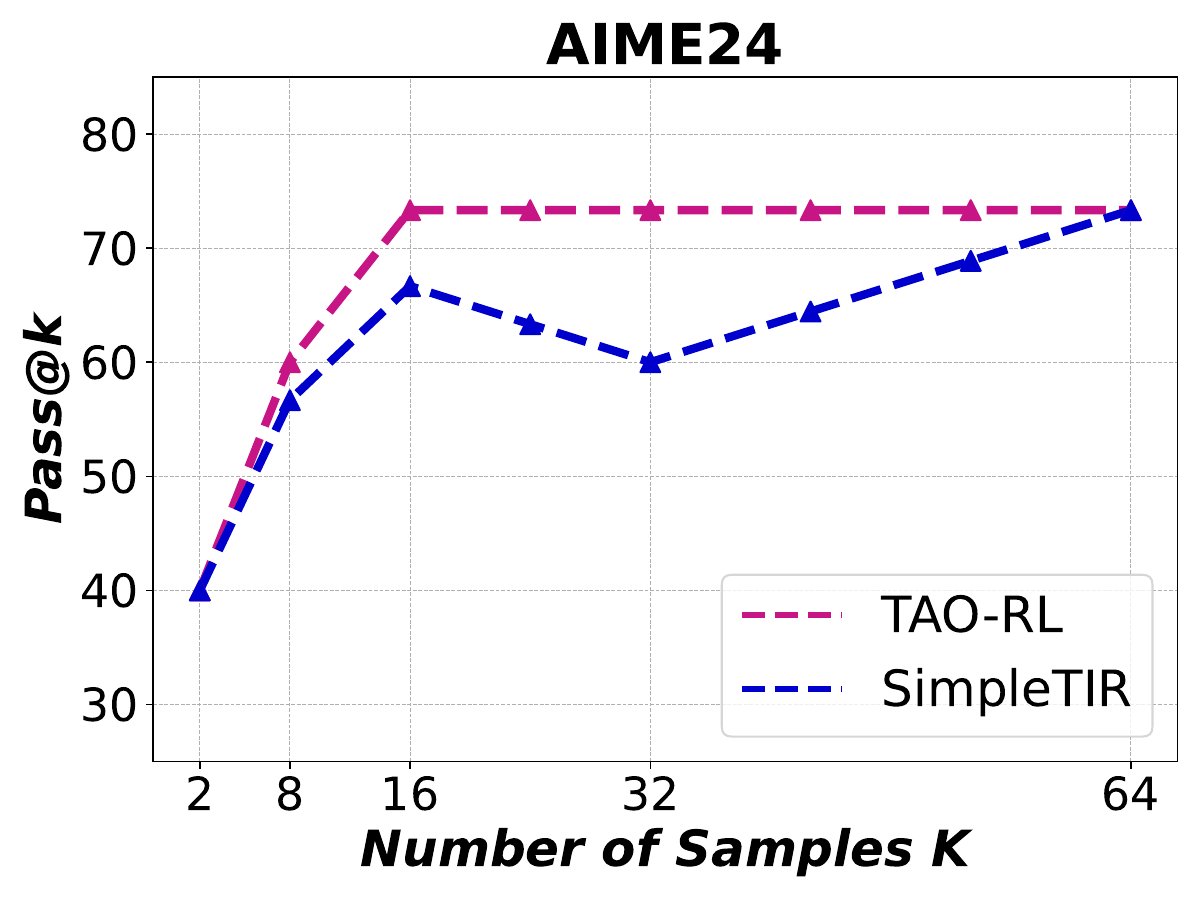}
\includegraphics[width=0.24\linewidth]{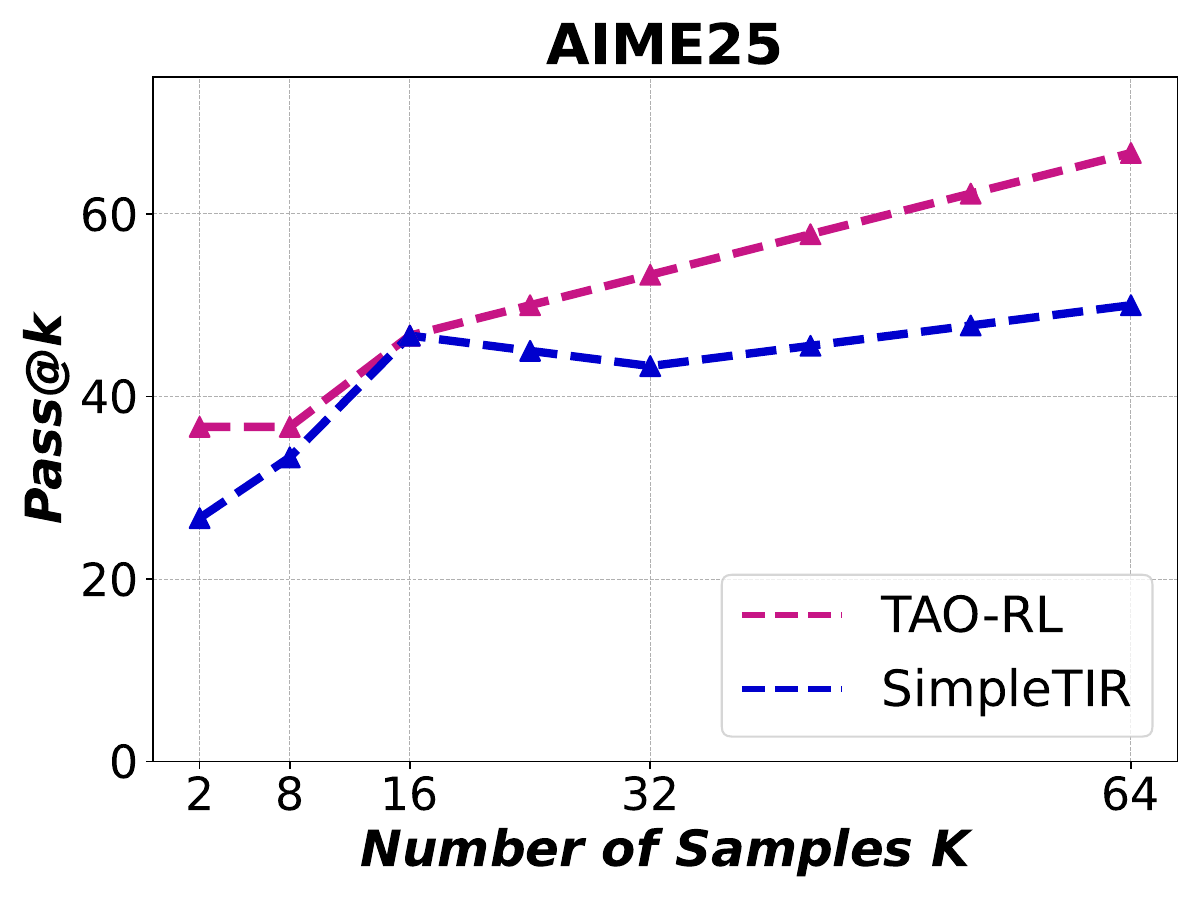}
\includegraphics[width=0.24\linewidth]{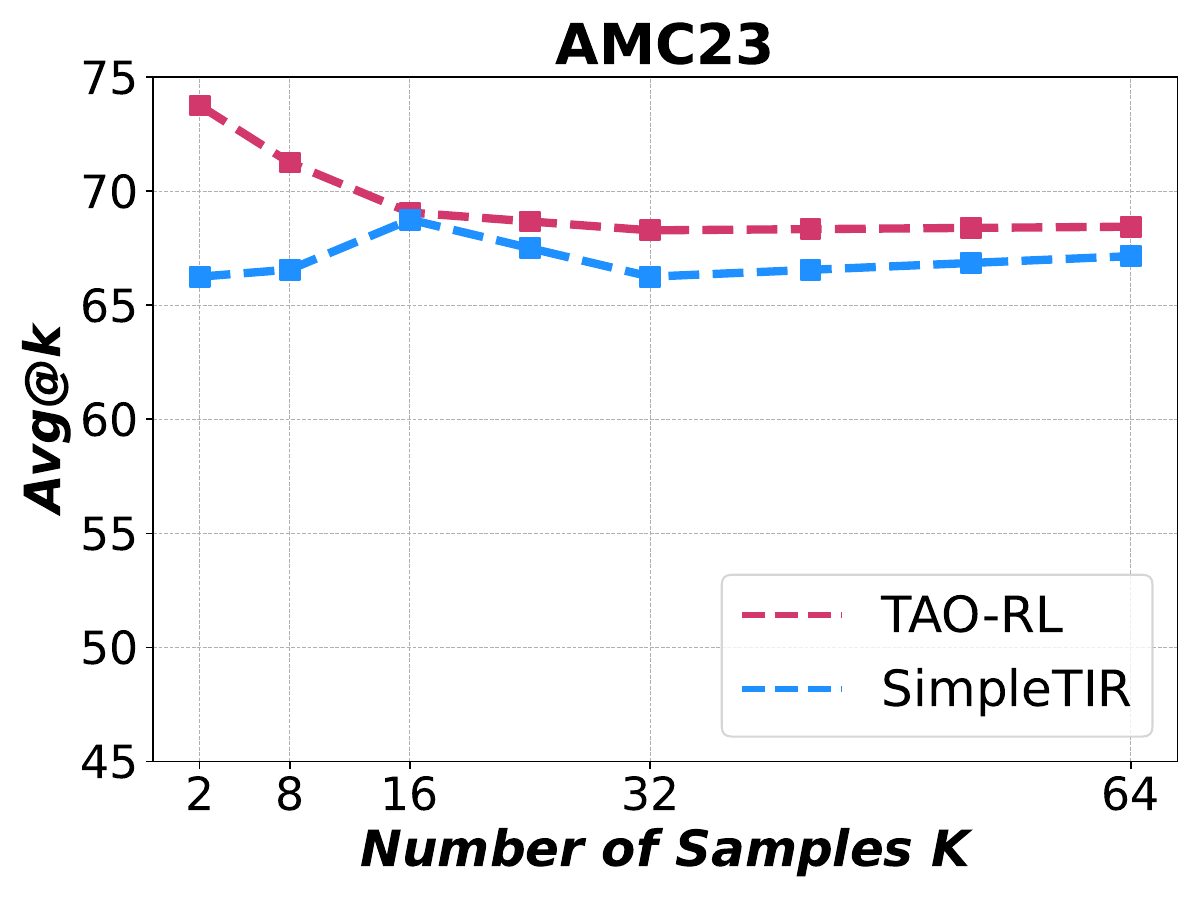}
\includegraphics[width=0.24\linewidth]{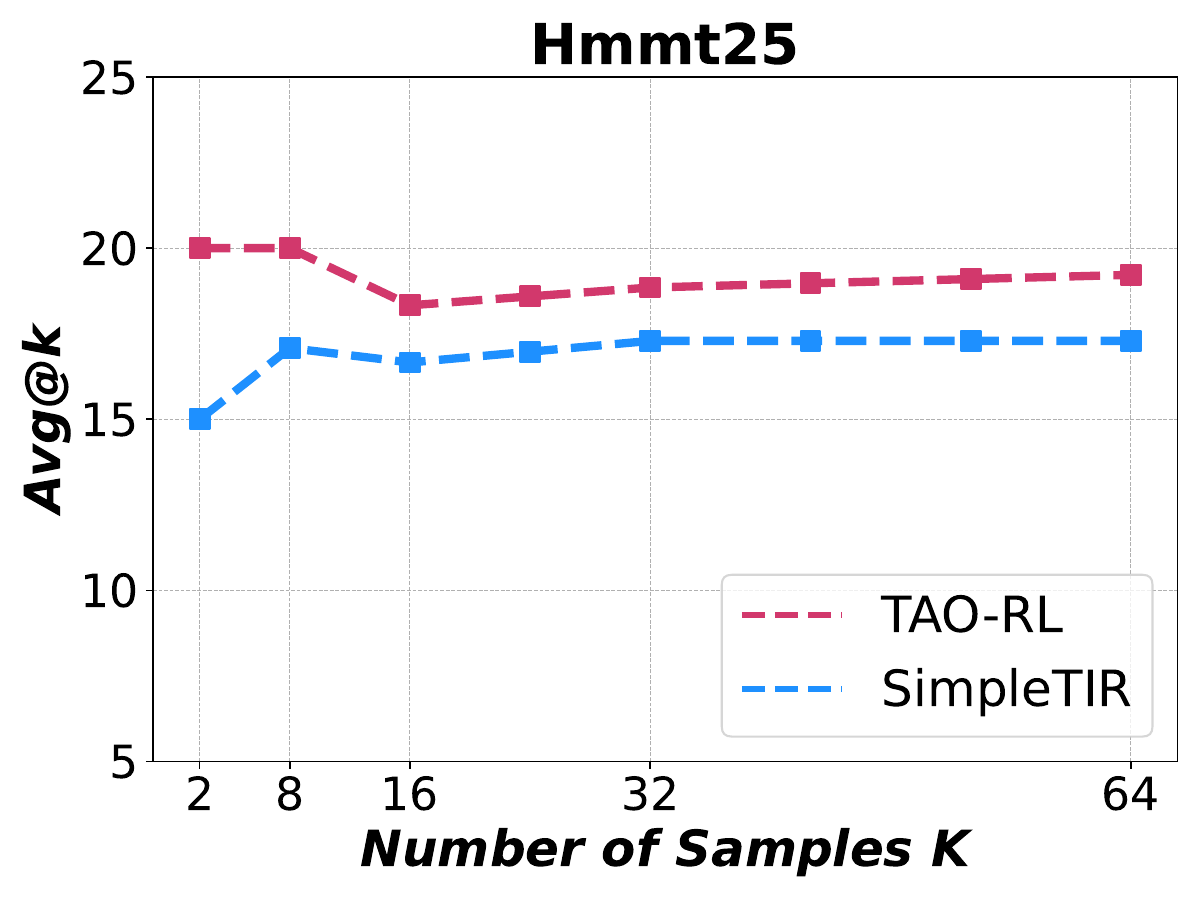}
\includegraphics[width=0.24\linewidth]{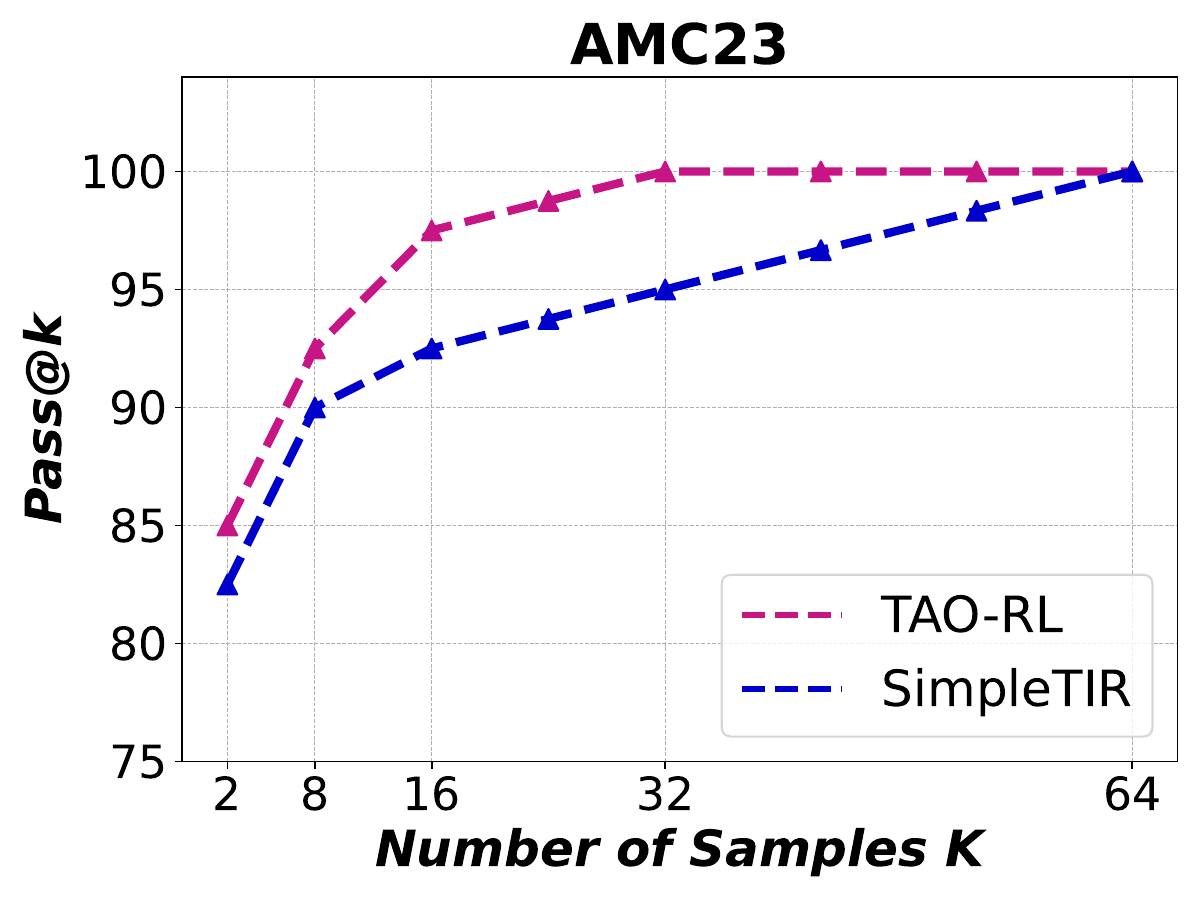}
\includegraphics[width=0.24\linewidth]{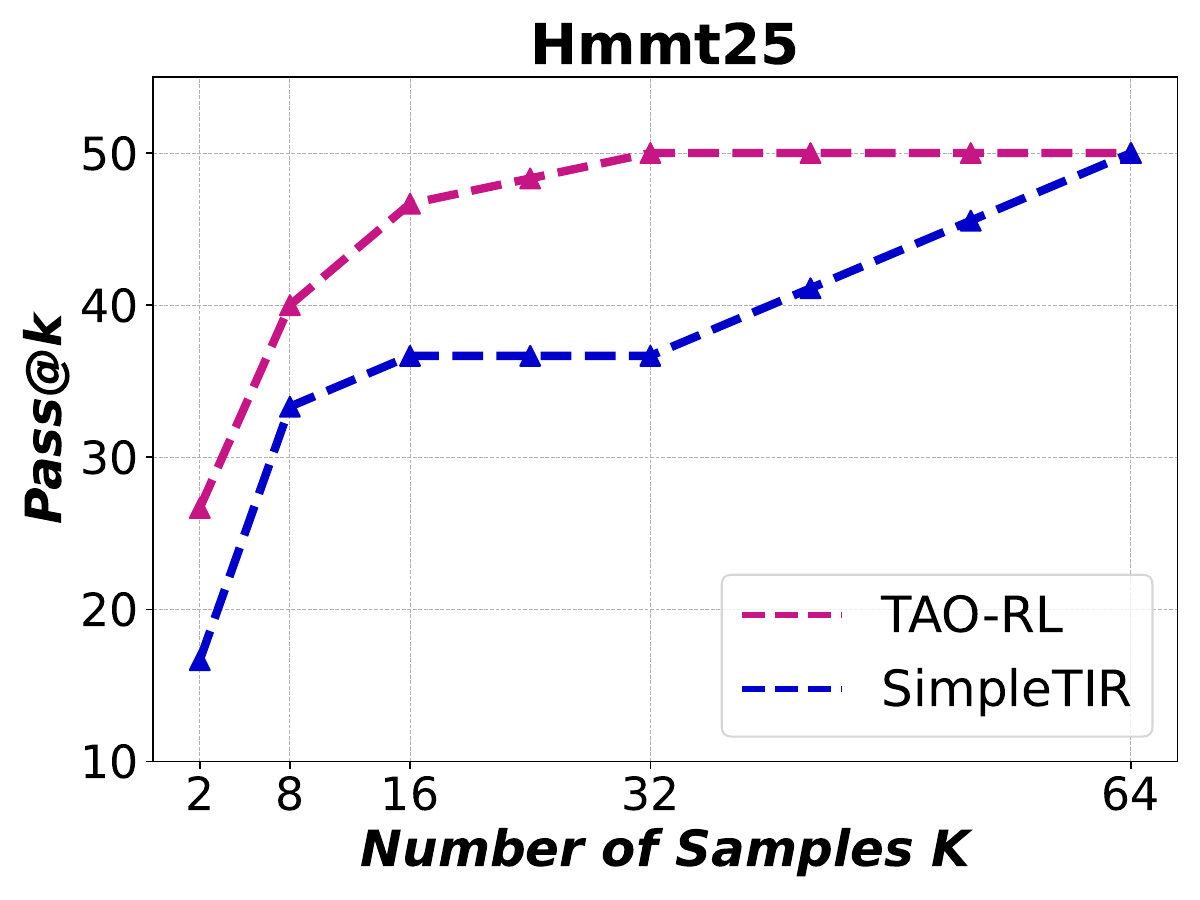}
\includegraphics[width=0.24\linewidth]{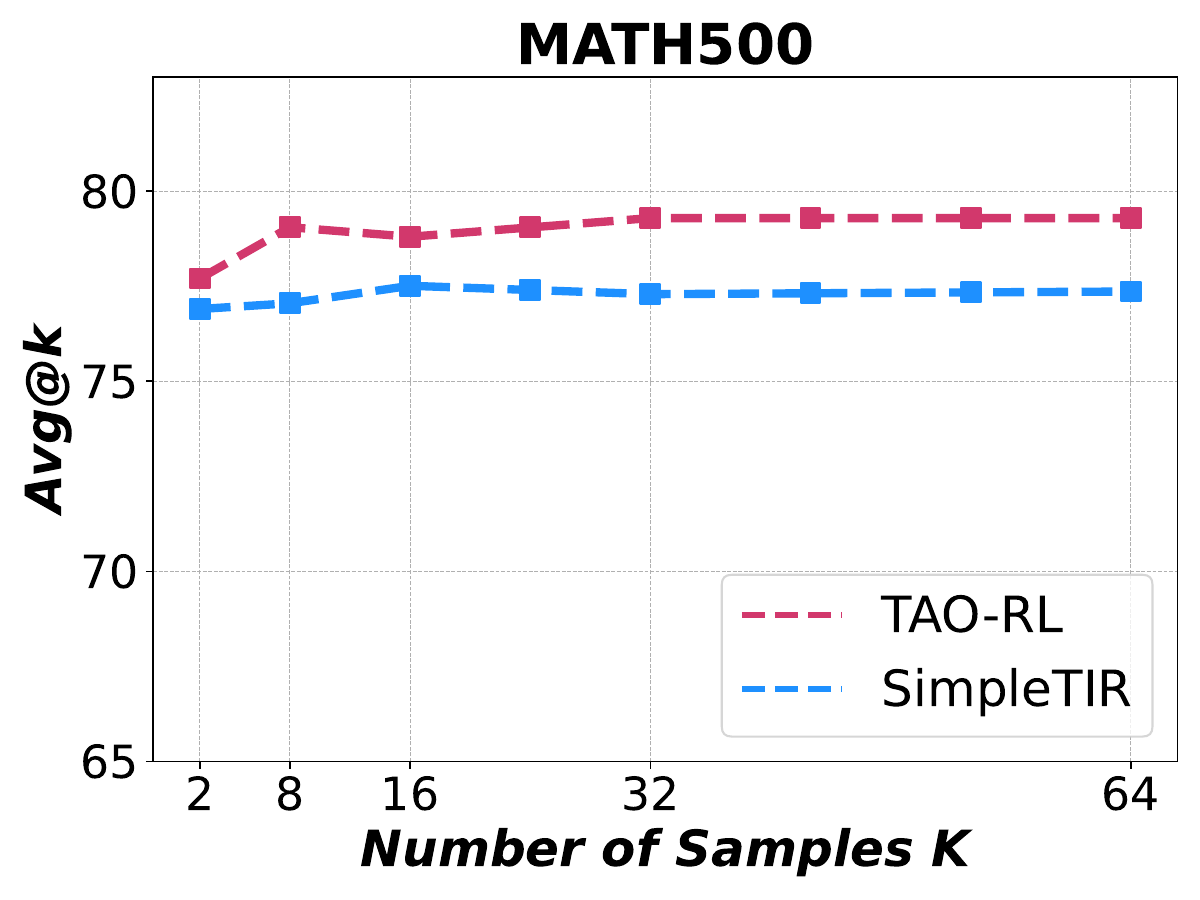}
\includegraphics[width=0.24\linewidth]{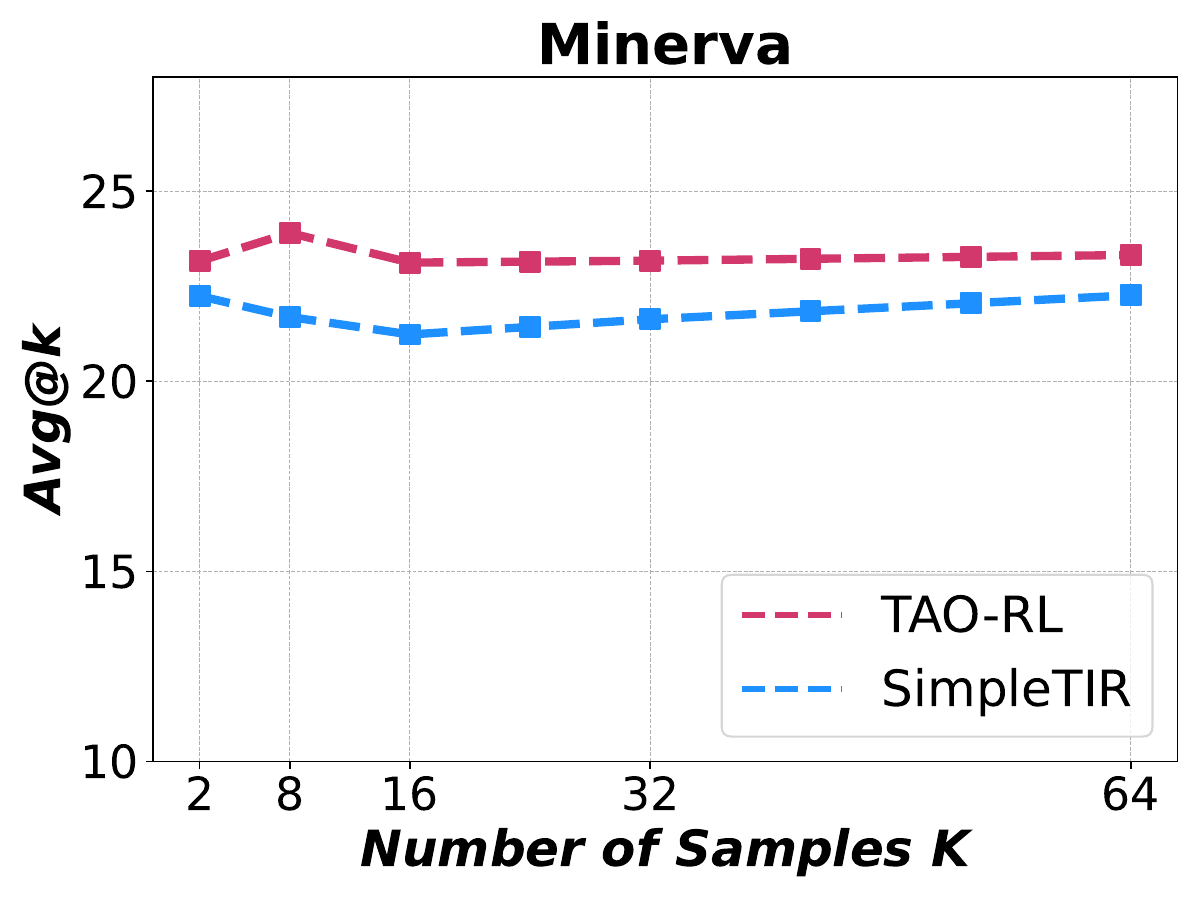}
\includegraphics[width=0.24\linewidth]{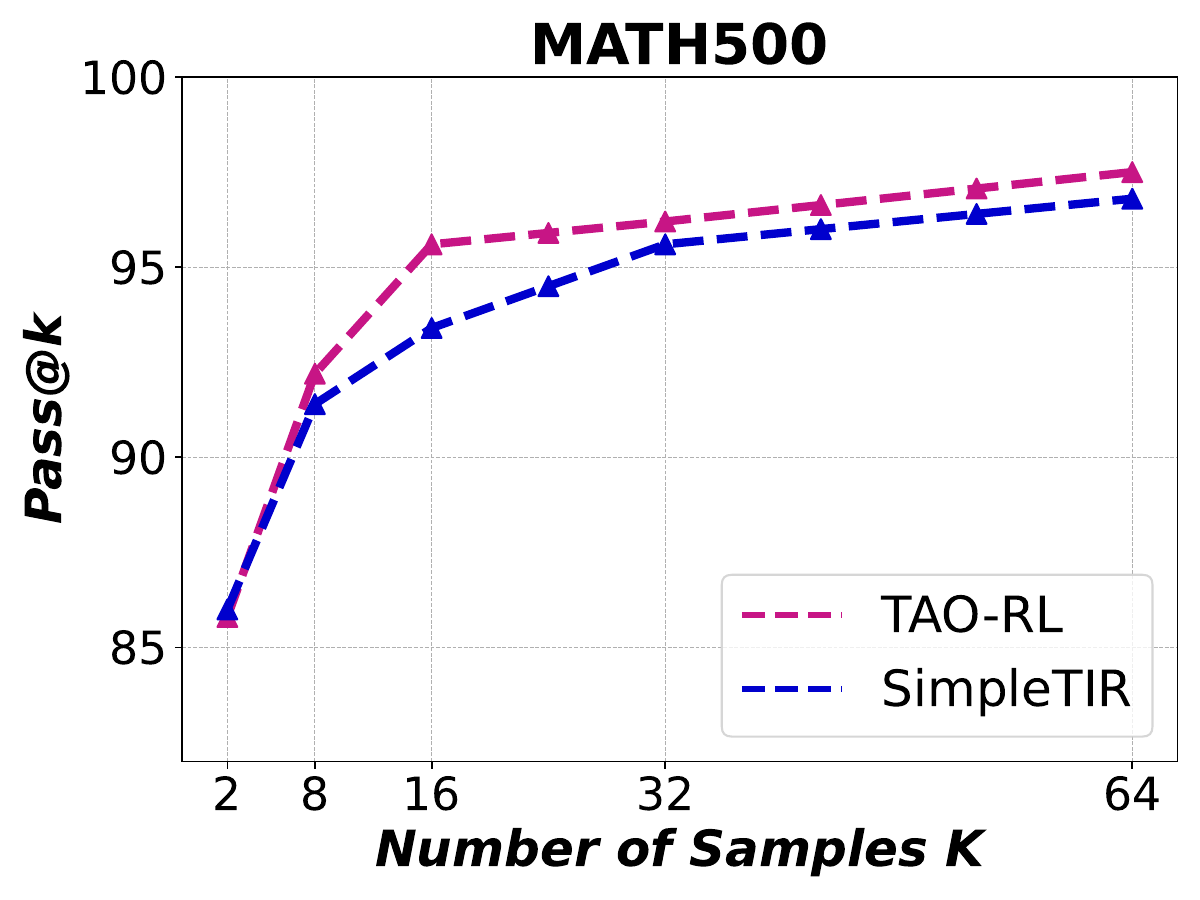}
\includegraphics[width=0.24\linewidth]{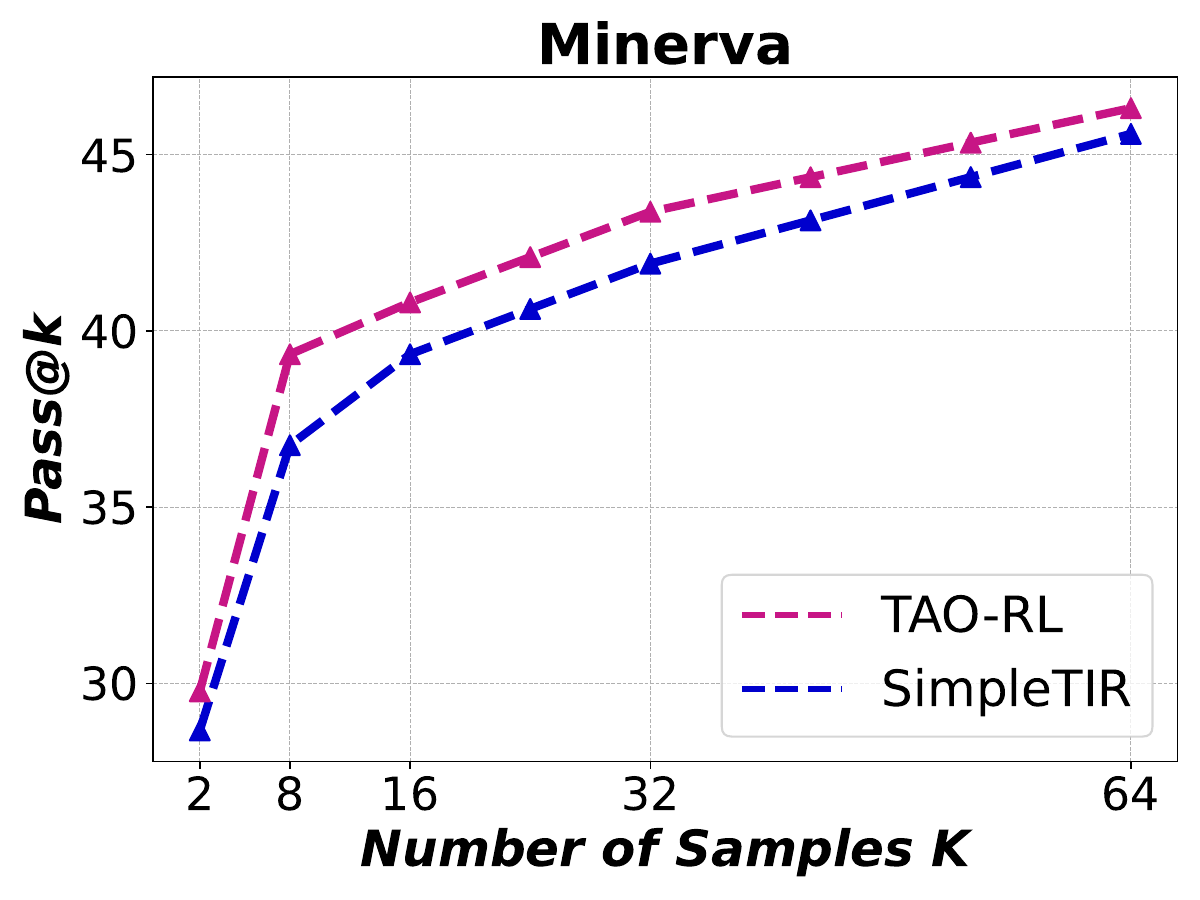}
\caption{Asymptotic performance of \textit{Pass@K} and \textit{Avg@K} curves of \method compared with SimpleTIR across $6$ benchmarks with the increase of number of samples $K$ under Qwen2.5-$7$B base models.}
\label{fig:pass@k_avg@k_curves}
\end{figure*}

\begin{table*}[t]
 \caption{Performance comparison on seven benchmarks on \emph{Len@16}. We bold the best results. $\Delta$ means the difference between the results of \method and sub-optimal results.} 
\renewcommand{\arraystretch}{1.4}
\setlength{\tabcolsep}{7pt} 
    \centering 
    \label{tab:exp_length}
\begin{tabular}{llcccccccc}
\toprule & \textbf{Method} & \textbf{AIME24} & \textbf{AIME25} & \textbf{AMC23}  & \textbf{MATH500} & \textbf{OlympiadBench} & \textbf{Hmmt25}                       & \textbf{Minerva}  & \textbf{Avg.} \\ \bottomrule 

\multirow{3}{*}{\textbf{Qwen2.5-1.5B}} &
 \textbf{SimpleTIR}  
 & 6040.80 & 5892.62 & 4275.34 & 2859.51  & 4318.35  & 5789.22  & 3519.68 & 4670.79  \\   
 &  
 \cellcolor{blue!10}{\textbf{\method}} & \cellcolor{blue!10}\textbf{7027.20} & \cellcolor{blue!10}\textbf{6782.57} & \cellcolor{blue!10}\textbf{4695.61} & 
 \cellcolor{blue!10}\textbf{3102.81} & \cellcolor{blue!10}\textbf{4907.56} & \cellcolor{blue!10}\textbf{6420.75} &    \cellcolor{blue!10}\textbf{3777.11} &
 \cellcolor{blue!10}\textbf{5244.80}  \\ &
\quad $\Delta$ &
\color{darkgreen}{+986.40} & 
\color{darkgreen}{+889.95} & 
\color{darkgreen}{+420.27} & 
\color{darkgreen}{+243.30} & 
\color{darkgreen}{+589.21} & 
\color{darkgreen}{+631.53} & 
\color{darkgreen}{+257.43} &
\color{darkgreen}{+574.01} 
\\ \hline 

\multirow{3}{*}{\textbf{Qwen3-4B-Base}} &
 \textbf{SimpleTIR}   & 16548.96  &  14316.50  & 9669.31   &  5894.48   &  10106.93   & 15361.80  &  6996.83 & 11270.69  \\   &  
 \cellcolor{blue!10}{\textbf{\method}} & \cellcolor{blue!10}\textbf{24056.69} & \cellcolor{blue!10}\textbf{23212.94} & \cellcolor{blue!10}\textbf{13057.05} & \cellcolor{blue!10}\textbf{9350.45} & \cellcolor{blue!10}\textbf{15164.22} & \cellcolor{blue!10}\textbf{22056.64} & \cellcolor{blue!10}\textbf{9312.66} &    \cellcolor{blue!10}\textbf{16601.52} \\ &
\quad $\Delta$ &
\color{darkgreen}{+7507.73} & 
\color{darkgreen}{+8896.44} & 
\color{darkgreen}{+3387.74} & 
\color{darkgreen}{+3455.97} & 
\color{darkgreen}{+5057.29} & 
\color{darkgreen}{+6694.84} & 
\color{darkgreen}{+2315.83} &
\color{darkgreen}{+5330.83}
\\ \hline 

\multirow{3}{*}{\textbf{Qwen2.5-7B}} & 
 \textbf{SimpleTIR}   & 5605.12 & 5527.23 & 3663.48 & 2548.87 & 4184.63 & 5972.69 & 3390.33 & 4413.19 \\   &  
 \cellcolor{blue!10}{\textbf{\method}} & \cellcolor{blue!10}\textbf{6655.03} & \cellcolor{blue!10}\textbf{7217.36} & \cellcolor{blue!10}\textbf{4145.81} & \cellcolor{blue!10}\textbf{2828.20} & \cellcolor{blue!10}\textbf{4746.74} & \cellcolor{blue!10}\textbf{7161.11} & \cellcolor{blue!10}\textbf{3705.75} &    \cellcolor{blue!10}\textbf{5208.57} \\ &
\quad $\Delta$ & 
\color{darkgreen}{+1049.91} & 
\color{darkgreen}{+1690.13} & 
\color{darkgreen}{+482.33} & 
\color{darkgreen}{+279.33} & 
\color{darkgreen}{+562.11} & 
\color{darkgreen}{+1188.42} & 
\color{darkgreen}{+315.42} &
\color{darkgreen}{+795.38} \\
\bottomrule    
\end{tabular}
\label{tab:res_len}
\end{table*}

\subsection{Main Results}

\subsubsection{Performance in Qwen2.5-1.5B}

Table~\ref{tab:exp_1.5b} presents the results under the Qwen2.5-1.5B base model. \method consistently outperforms SimpleTIR across the majority of benchmarks in both Avg@16 and Pass@16 metrics. On the challenging AIME24 benchmark, \method achieves an Avg@16 of $11.88$ and a Pass@16 of $50.00$, surpassing SimpleTIR by $3.13$ and $16.67$ percentage points respectively, demonstrating substantially stronger reasoning capability and exploration diversity at the harder end of the difficulty spectrum. On AIME25, \method improves Avg@16 by $1.87$, though Pass@16 shows a marginal decline, suggesting that at this model scale, the exploration gain on extremely hard problems is better reflected in average accuracy than in peak solve rate. On MATH500 and AMC23, \method delivers consistent improvements in both metrics, confirming its effectiveness across benchmarks of varying difficulty. On Hmmt25, \method achieves a notable Avg@16 gain of $2.71$ while maintaining parity on Pass@16. These results collectively demonstrate that at small model scale, \method effectively leverages tool-aware filtering and entropy-guided exploration to improve reasoning performance.

\subsubsection{Performance in Qwen3-4B-Base}

As shown in Table~\ref{tab:exp_1.5b}, scaling to the Qwen3-4B-Base model further amplifies the advantages of \method. On AIME24, \method achieves an Avg@16 of $28.13$ and a Pass@16 of $63.33$, outperforming SimpleTIR by $0.63$ and $6.66$ respectively. On AIME25, \method matches SimpleTIR on Pass@16 while improving Avg@16 by $1.17$, indicating more consistent reasoning across rollouts. Gains are observed across all remaining benchmarks: \method improves MATH500 Avg@16 by $2.61$ and Pass@16 by $1.20$, Olympiad by 
$0.93$ and $0.40$, and Minerva by $0.29$ and $2.21$. On AMC23, \method improves Avg@16 by $0.94$. On Hmmt25, both methods achieve identical scores, suggesting the benchmark approaches the current capability ceiling for this model scale. \method demonstrates consistent and broad improvements at the 4B scale, confirming its scalability.

\subsubsection{Performance in Qwen2.5-7B}

Table~\ref{tab:exp_7b} reports results under the Qwen2.5-7B model trained on the larger dataset, where \method is compared against more baselines of TIR, AEPO, and SimpleTIR. \method achieves the best performance on all seven benchmarks in both Avg@16 and Pass@16. Compared to the strongest baseline SimpleTIR, \method delivers consistent improvements: $2.09/6.66$ on AIME24, $2.92$ on AIME25, $1.29/2.20$ on MATH500, and notably $4.04/3.40$ on OlympiadBench, which represents the largest absolute Avg@16 gain across all settings. On AMC23 and Hmmt25 benchmarks, \method improves Pass@16 by $5.00$ and $6.67$ respectively, indicating substantially stronger peak reasoning capability. Compared to AEPO and TIR, \method outperforms by a wider margin across nearly all metrics, confirming the advantage of jointly addressing data quality and exploration at the algorithmic level. \textbf{These results demonstrate that \method scales effectively to larger models and benefits further from richer training data.}

Furthermore, Fig.~\ref{fig:pass@k_avg@k_curves} presents the asymptotic performance curves of Avg@K and Pass@K as the number of samples $K$ increases across six benchmarks under Qwen2.5-7B. On the Avg@K metric, \method maintains a clear and persistent advantage over SimpleTIR at every value of $K$, demonstrating that the overall reasoning accuracy of \method is consistently superior regardless of the sampling budget. This sustained gap confirms that \method improves the average quality of individual reasoning trajectories rather than merely increasing the diversity of outputs, reflecting a genuine uplift in the policy's core reasoning capability.

On the Pass@K metric, \method demonstrates two complementary advantages. First, at small $K$, \method achieves substantially higher Pass@K than SimpleTIR, indicating that the policy can reach correct solutions with far fewer samples. Second, \method reaches its performance ceiling rapidly as $K$ increases. In contrast, SimpleTIR requires $K$ up to $64$ to approach \method's performance on only three benchmarks, and remains inferior on the remaining ones even at large $K$. These results demonstrate that \method raises the average and ceiling performance, establishing a more stable and effective optimization framework.

\begin{table*}[t]
 \caption{Performance comparison on seven benchmarks on \emph{Valid\_Code (\%)}. We bold the best results and underline the sub-optimal results. $\Delta$ means the difference between the results of \method and sub-optimal results.} 
\renewcommand{\arraystretch}{1.4}
\setlength{\tabcolsep}{7pt} 
    \centering 
\begin{tabular}{llcccccccc}
\toprule & \textbf{Method} & \textbf{AIME24} & \textbf{AIME25} & \textbf{AMC23}  & \textbf{MATH500} & \textbf{OlympiadBench} & \textbf{Hmmt25}                       & \textbf{Minerva} & \textbf{Avg.} \\ \bottomrule 
\multirow{4}{*}{\textbf{Qwen2.5-7B}} &
 \textbf{TIR}   & 48.96  & 45.63  &  56.41  & 63.60  &  57.40  &  44.58  & 62.64  & 54.17 \\
  &  \textbf{SimpleTIR}   & \underline{97.71}  & \underline{98.54}  &  \underline{99.22}  & \underline{99.60}  & \textbf{99.40}  & \underline{99.22}  &  \textbf{99.58} & \underline{99.03}  \\   &  
 \cellcolor{blue!10}{\textbf{\method}} & \cellcolor{blue!10}\textbf{97.92} & \cellcolor{blue!10}\textbf{99.17} & \cellcolor{blue!10}\textbf{99.69} & \cellcolor{blue!10}\textbf{99.64} & \cellcolor{blue!10}\underline{99.38} & \cellcolor{blue!10}\textbf{99.79} &    \cellcolor{blue!10}\underline{99.36} &
 \cellcolor{blue!10}\textbf{99.28}  \\ &
\quad $\Delta$ & \color{darkgreen}{+0.21} & \color{darkgreen}{+0.63} & \color{darkgreen}{+ 0.47} & \color{darkgreen}{+0.04} & \color{darkred}{-0.02} & \color{darkgreen}{+0.57} & \color{darkred}{-0.22} & \color{darkgreen}{+0.24} \\
\bottomrule    
\end{tabular}
\label{tab:res_valid_code}
\end{table*}

\begin{table*}[t]
 \caption{Performance comparison on seven benchmarks on \emph{Code\_Line}. We bold the best results and underline the sub-optimal results. $\Delta$ means the difference between the results of \method and sub-optimal results.} 
\renewcommand{\arraystretch}{1.4}
\setlength{\tabcolsep}{7pt} 
    \centering 
\begin{tabular}{llcccccccc}
\toprule & \textbf{Method} & \textbf{AIME24} & \textbf{AIME25} & \textbf{AMC23}  & \textbf{MATH500} & \textbf{OlympiadBench} & \textbf{Hmmt25}     &     \textbf{Minerva}             & \textbf{Avg.} \\ \bottomrule 
\multirow{4}{*}{\textbf{Qwen2.5-7B}} & 
\textbf{TIR}   &  10.34  &  9.51  &  9.93  &  9.07  &  10.50  &  9.02  &  10.91  &  9.90 \\ 
& \textbf{SimpleTIR}   & \underline{64.24}  & \underline{65.15}  & \underline{41.68}  & \underline{28.00}  & \underline{46.90}  & \underline{71.77}  & \underline{32.47} & \underline{50.03}   \\   &  
 \cellcolor{blue!10}{\textbf{\method}} & \cellcolor{blue!10}\textbf{ 75.35 } & \cellcolor{blue!10}\textbf{ 87.26 } & \cellcolor{blue!10}\textbf{ 46.62 } & \cellcolor{blue!10}\textbf{ 30.22 }  & \cellcolor{blue!10}\textbf{ 52.18 }  & \cellcolor{blue!10}\textbf{ 86.76 }    &    \cellcolor{blue!10}\textbf{ 37.88 }  &
 \cellcolor{blue!10}\textbf{59.48} \\ &
\quad $\Delta$ & \color{darkgreen}{+11.11} & \color{darkgreen}{+22.11} & \color{darkgreen}{+4.94} & \color{darkgreen}{+2.22} & \color{darkgreen}{+5.28} & \color{darkgreen}{+14.99} & \color{darkgreen}{+5.41} & \color{darkgreen}{+9.45} \\
\bottomrule    
\end{tabular}
\label{tab:res_len}
\end{table*}

\begin{table*}[t]
\caption{Performance comparison on seven datasets under Qwen2.5-7B base model. \textit{Box\_Rate} measures the proportion of responses that correctly follow the required answer format, and \textit{Tool\_Call} quantifies the average number of tool invocations per response. We bold the best scores and underline the sub-optimal results. $\Delta$ means the difference between \method and sub-optimal results.}
\renewcommand{\arraystretch}{1.4}
\setlength{\tabcolsep}{1.5pt}
\centering
\begin{tabular}{ccccccccccccccc}
\toprule
\multicolumn{1}{c}{\textbf{Method}} & \multicolumn{2}{c}{\textbf{AIME24}}         & \multicolumn{2}{c}{\textbf{AIME25}}         & \multicolumn{2}{c}{\textbf{MATH500}}        & \multicolumn{2}{c}{\textbf{Olympiad}}       & \multicolumn{2}{c}{\textbf{AMC23}}          & \multicolumn{2}{c}{\textbf{Hmmt25}}         &
 \multicolumn{2}{c}{\textbf{Minerva}} \\ 
\cmidrule(lr){2-3} \cmidrule(lr){4-5} \cmidrule(lr){6-7} \cmidrule(lr){8-9} \cmidrule(lr){10-11} \cmidrule(lr){12-13}
\cmidrule(lr){10-11} \cmidrule(lr){14-15}
                           & \textit{Box\_Rate} & \textit{Tool\_Call} & \textit{Box\_Rate} & \textit{Tool\_Call} & \textit{Box\_Rate} & \textit{Tool\_Call} & \textit{Box\_Rate} & \textit{Tool\_Call} & \textit{Box\_Rate} & \textit{Tool\_Call} & \textit{Box\_Rate} & \textit{Tool\_Call} & \textit{Box\_Rate} & \textit{Tool\_Call} \\ 
                           \bottomrule
\rowcolor{black!10}
\multicolumn{15}{c}{\textit{Qwen2.5-7B}}      \\ \hline
\textbf{TIR}
&      31.04      &       0.49 
&      29.58      &       0.46 
&      52.30      &       0.64   
&      41.11      &       0.57 
&      40.63      &       0.56 
&      29.79      &       0.45 
&      34.58      &       0.63          \\

\textbf{SimpleTIR} 
&      \underline{51.67}      &       \underline{2.04}
&      \underline{44.58}      &       \underline{1.89}
&    \underline{88.08}    &    \underline{1.39}
&    \underline{69.14}    &    \underline{1.72}
&    \underline{77.97}    &    \underline{1.66}
&      \underline{43.13}    &       \underline{1.95}
&     \underline{78.68}     &     \underline{1.58}        \\
\rowcolor{blue!10} \method
&      \textbf{58.75}      &       \textbf{2.26}
&      \textbf{51.46}      &       \textbf{2.35}
&      \textbf{91.43}      &       \textbf{1.45}
&      \textbf{77.78}      &       \textbf{1.87}
&      \textbf{81.88}      &       \textbf{1.79}
&      \textbf{54.79}     &       \textbf{2.33}
&      \textbf{85.41}      &       \textbf{1.75}      \\ 
$\Delta$
& \color{darkgreen}{+7.08} & \color{darkgreen}{+0.22}
& \color{darkgreen}{+6.88} & \color{darkgreen}{+0.46}
& \color{darkgreen}{+3.35} & \color{darkgreen}{+0.06}
& \color{darkgreen}{+8.64} & \color{darkgreen}{+0.15}
& \color{darkgreen}{+3.91} & \color{darkgreen}{+0.13}
& \color{darkgreen}{+11.66} & \color{darkgreen}{+0.38}
& \color{darkgreen}{+6.73} & \color{darkgreen}{+0.17}
\\
\bottomrule          
\end{tabular}
\label{tab:exp_too_call}
\end{table*}

\begin{figure*}[t]
\centering
\includegraphics[width=0.245\linewidth]{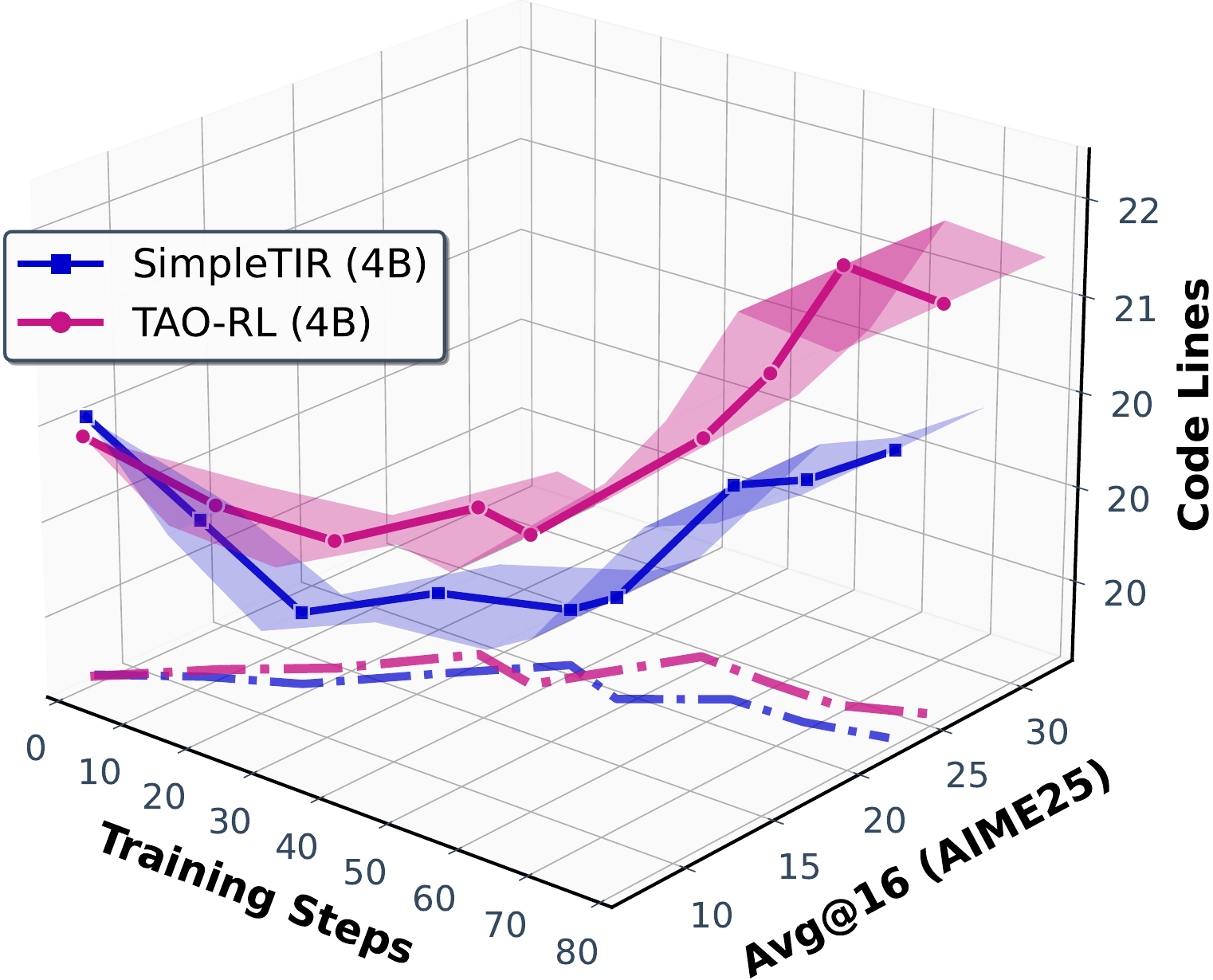}
\includegraphics[width=0.245\linewidth]{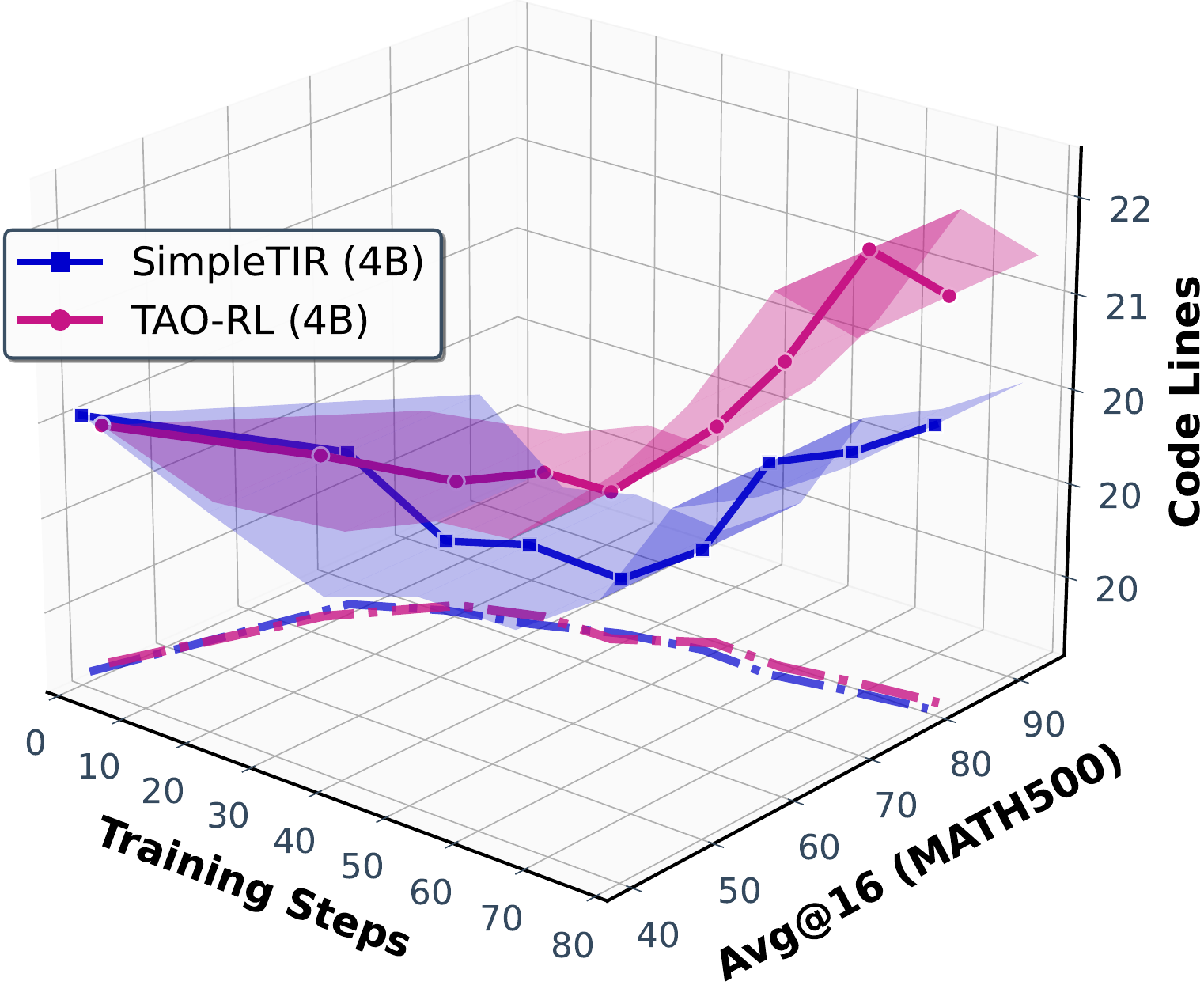}
\includegraphics[width=0.245\linewidth]{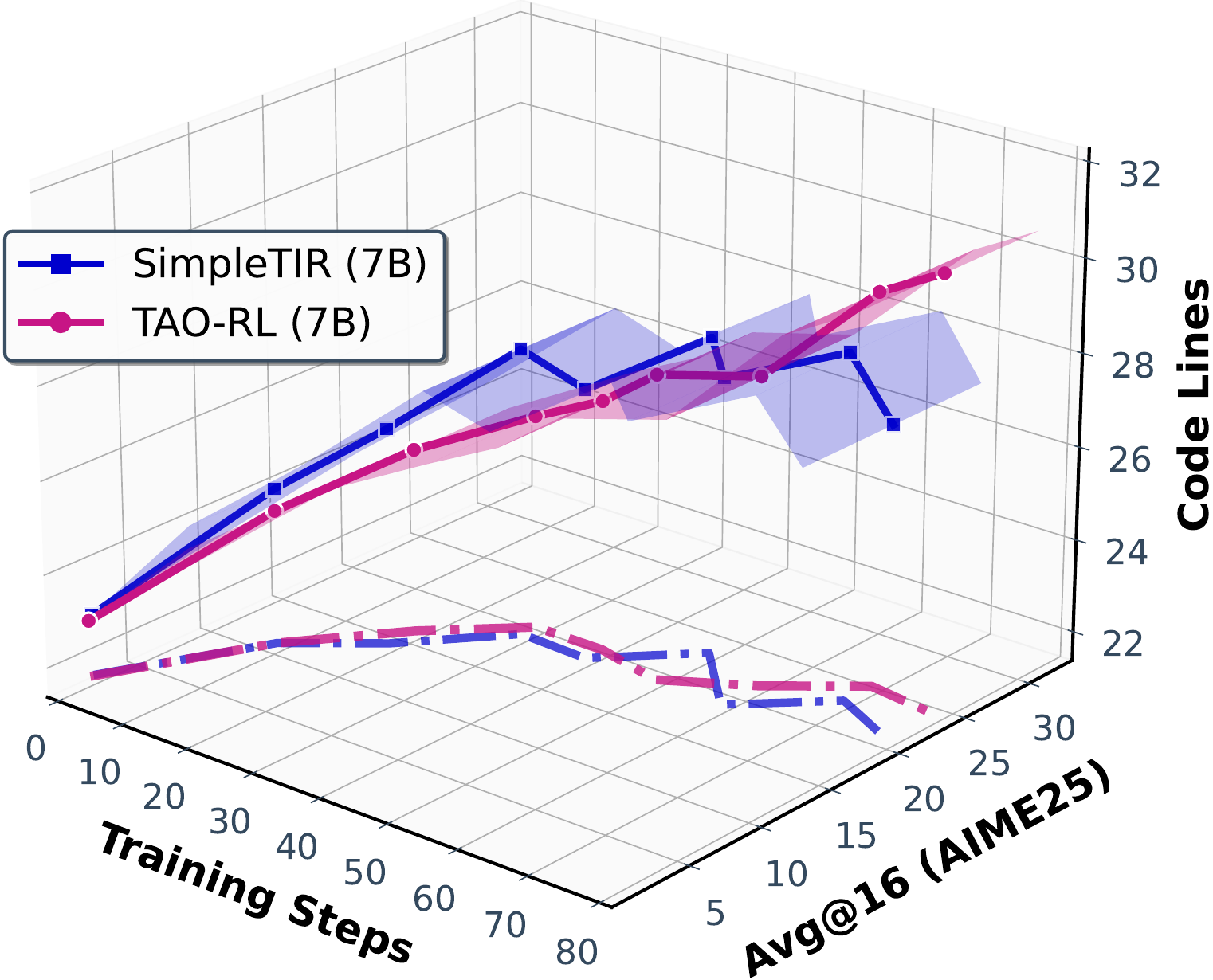}
\includegraphics[width=0.245\linewidth]{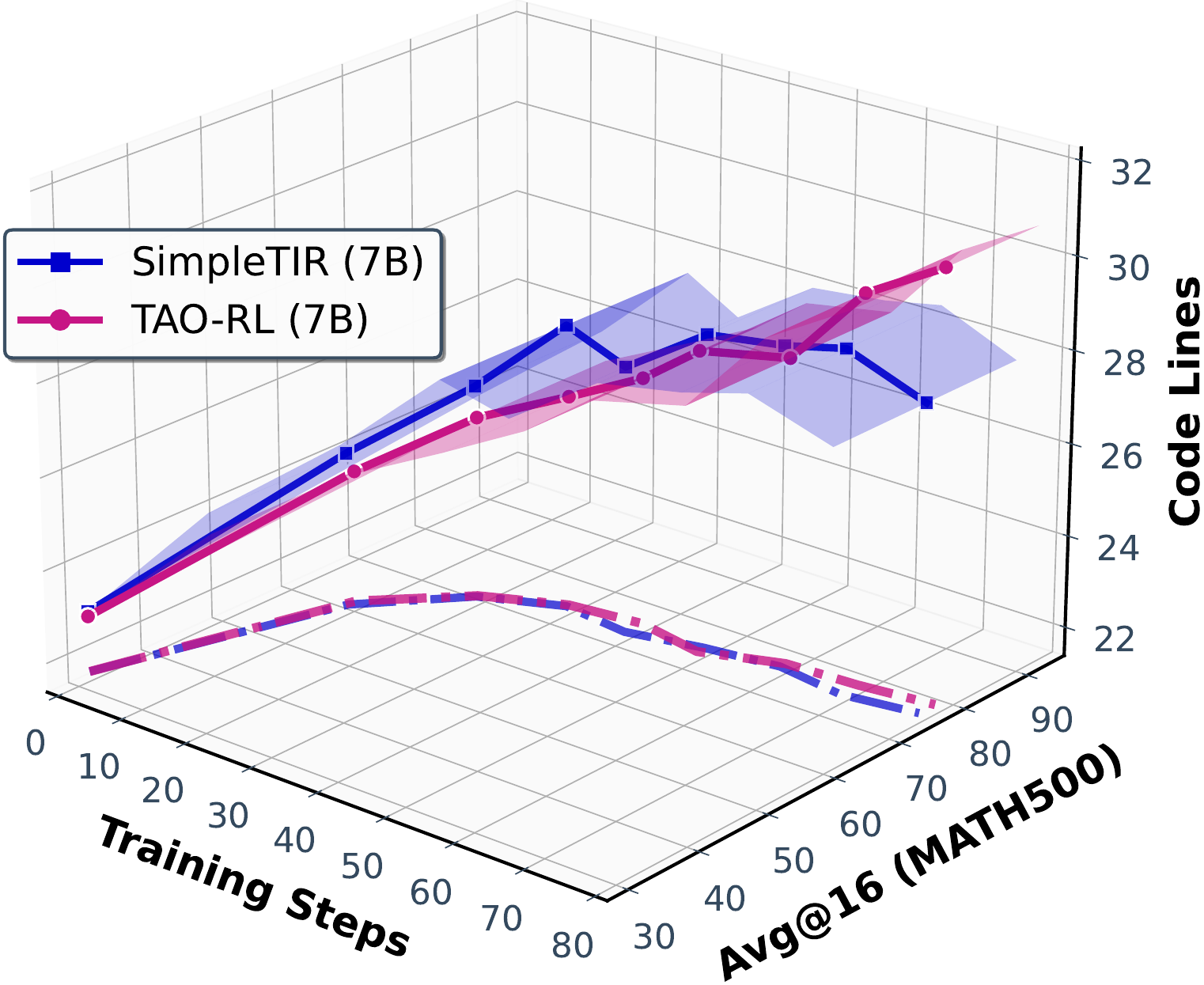}
\caption{Learning curves of average accuracy (Avg@16) and Code\_Line throughout training for \method and SimpleTIR under Qwen2.5-7B and Qwen3-4B-Base. The x-axis denotes training steps, the y-axis denotes average accuracy, and the z-axis denotes Code\_Lines.}
\label{fig:3d}
\end{figure*}

\subsubsection{Reasoning Ability Analysis}

Table~\ref{tab:exp_length} reports Len@16, the average response length across benchmarks, which serves as a metric for reasoning depth. Across all three model scales, \method consistently generates substantially longer responses than SimpleTIR. For Qwen2.5-1.5B, \method achieves an average Len@16 of $5244.80$, surpassing SimpleTIR by $574.01$ characters. The gap widens considerably at larger scales: for Qwen3-4B-Base, \method generates an average of $16601.52$ characters compared to SimpleTIR's $11270.69$, a difference of $5330.83$, indicating that the entropy-guided exploration encourages the model to engage in substantially more elaborate reasoning and tool interaction chains. For Qwen2.5-7B, the improvement of $795.38$ on average further confirms this trend. Notably, \method achieves larger reasoning improvement than SimpleTIR in hard benchmarks of AIME24\&25.
The consistent increase in response length across all benchmarks and model scales suggests that \textbf{\method not only improves accuracy but also cultivates deeper reasoning behaviors, particularly at post-tool-call decision points where the entropy bonus concentrates exploration incentives.}

\subsection{Property Analysis}
\label{sec:property}

\begin{figure}[t]
  \centering
   \includegraphics[width=0.49\linewidth]{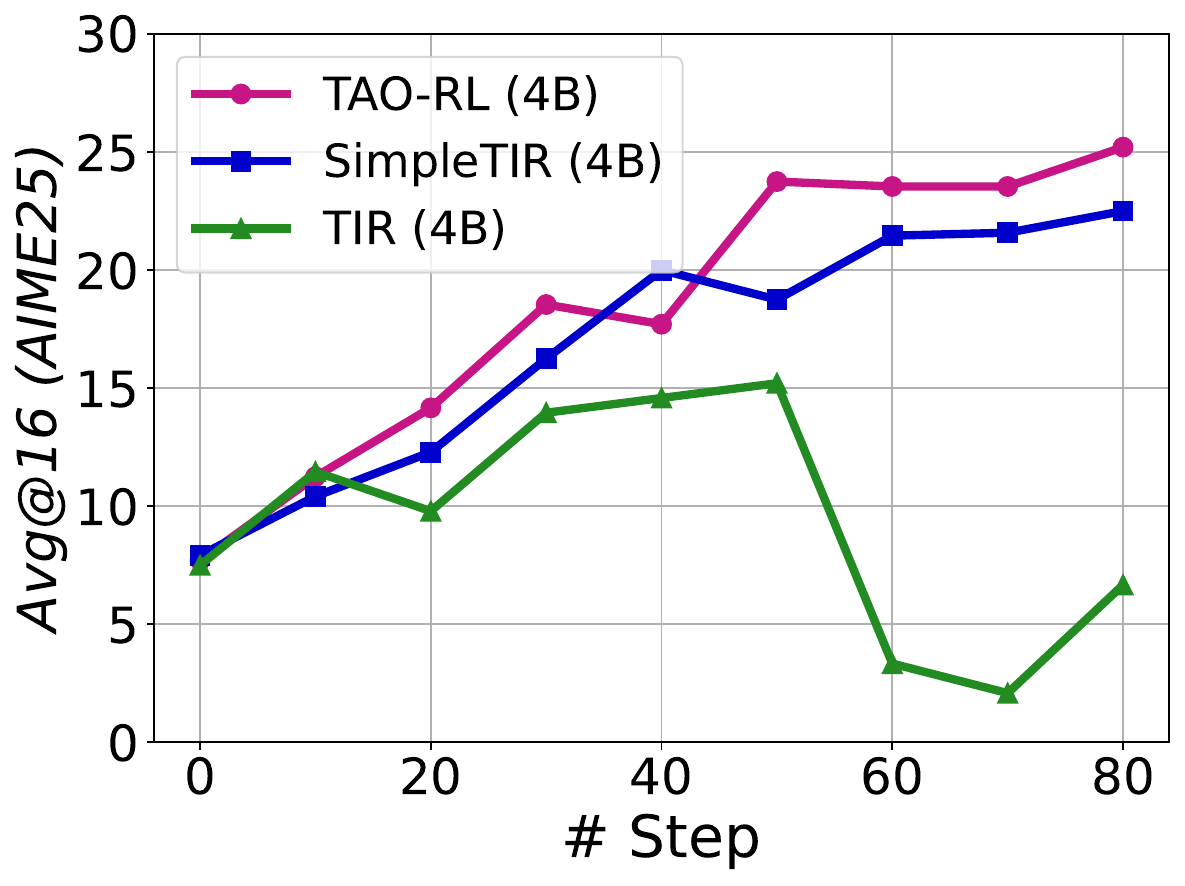} 
    \includegraphics[width=0.49\linewidth]{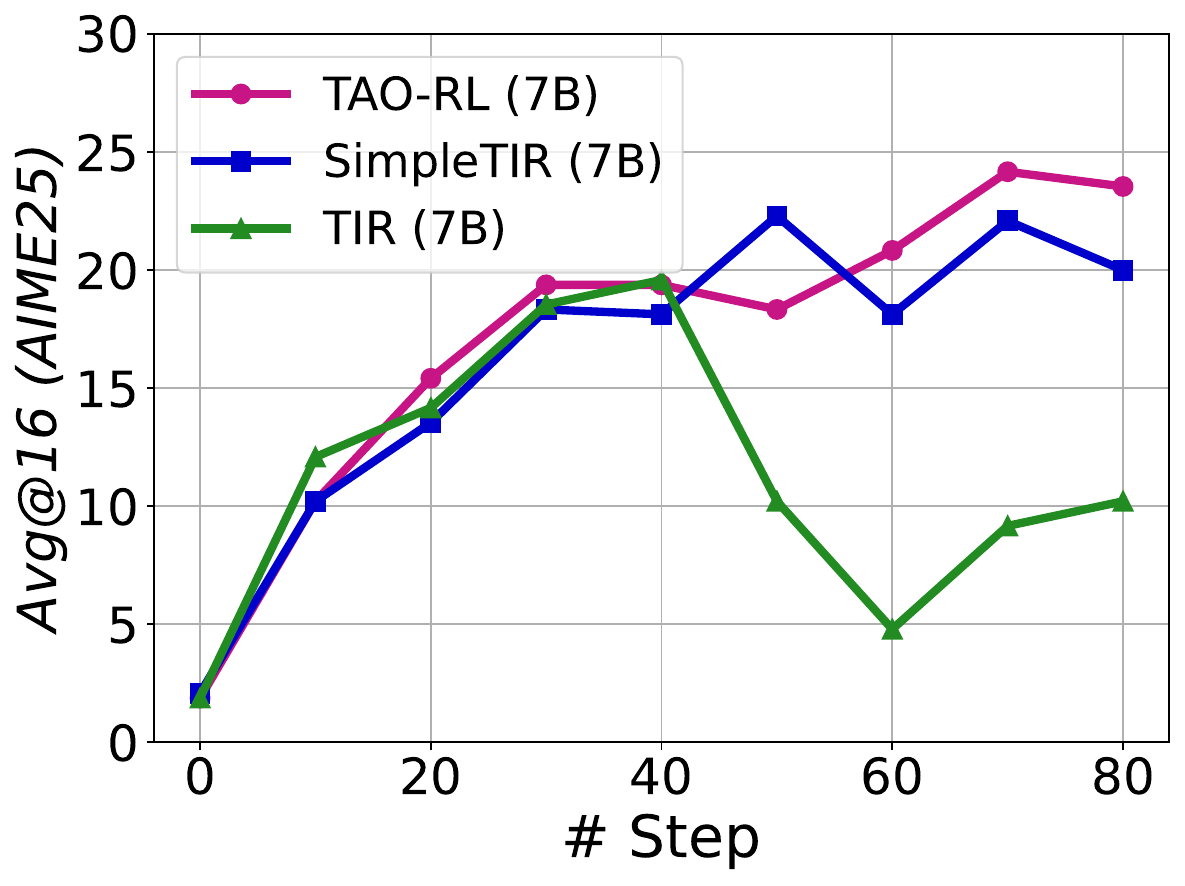} 
    \includegraphics[width=0.49\linewidth]{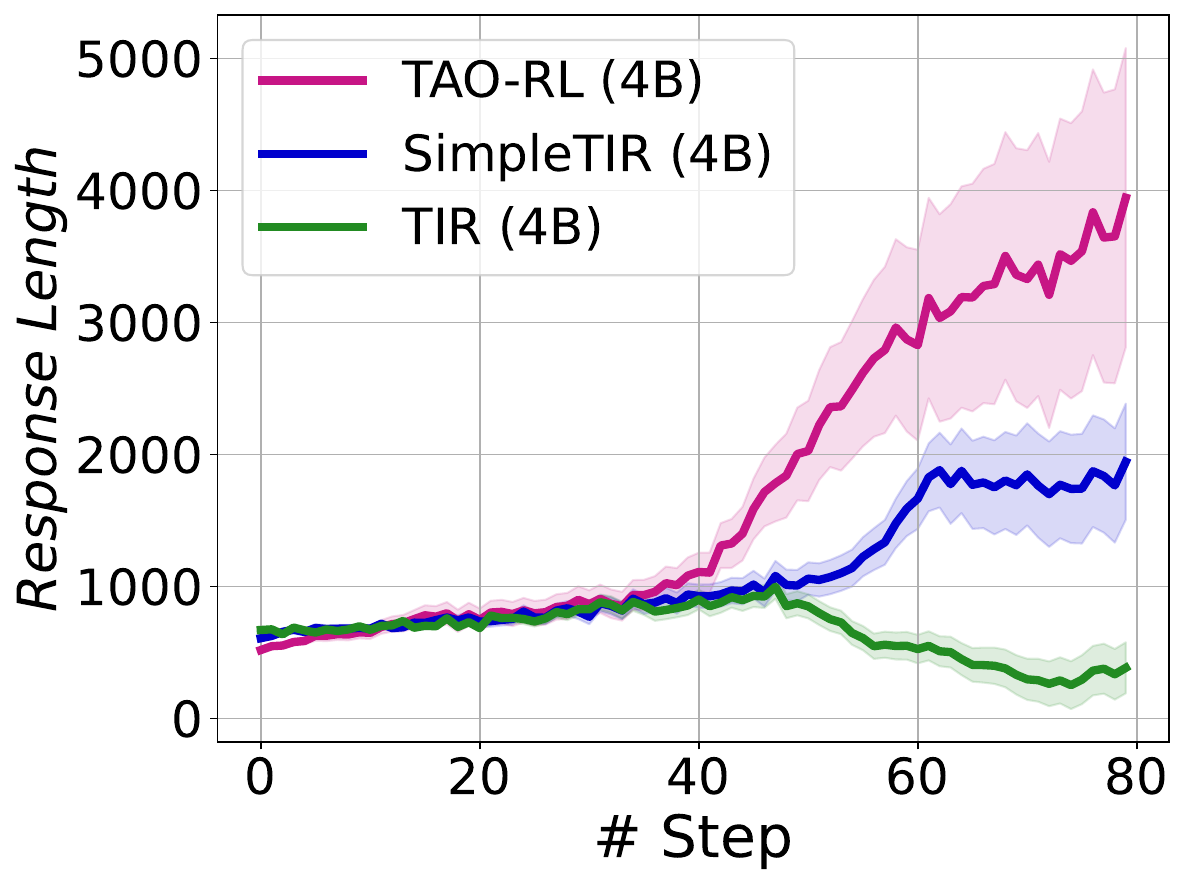} 
  \includegraphics[width=0.49\linewidth]{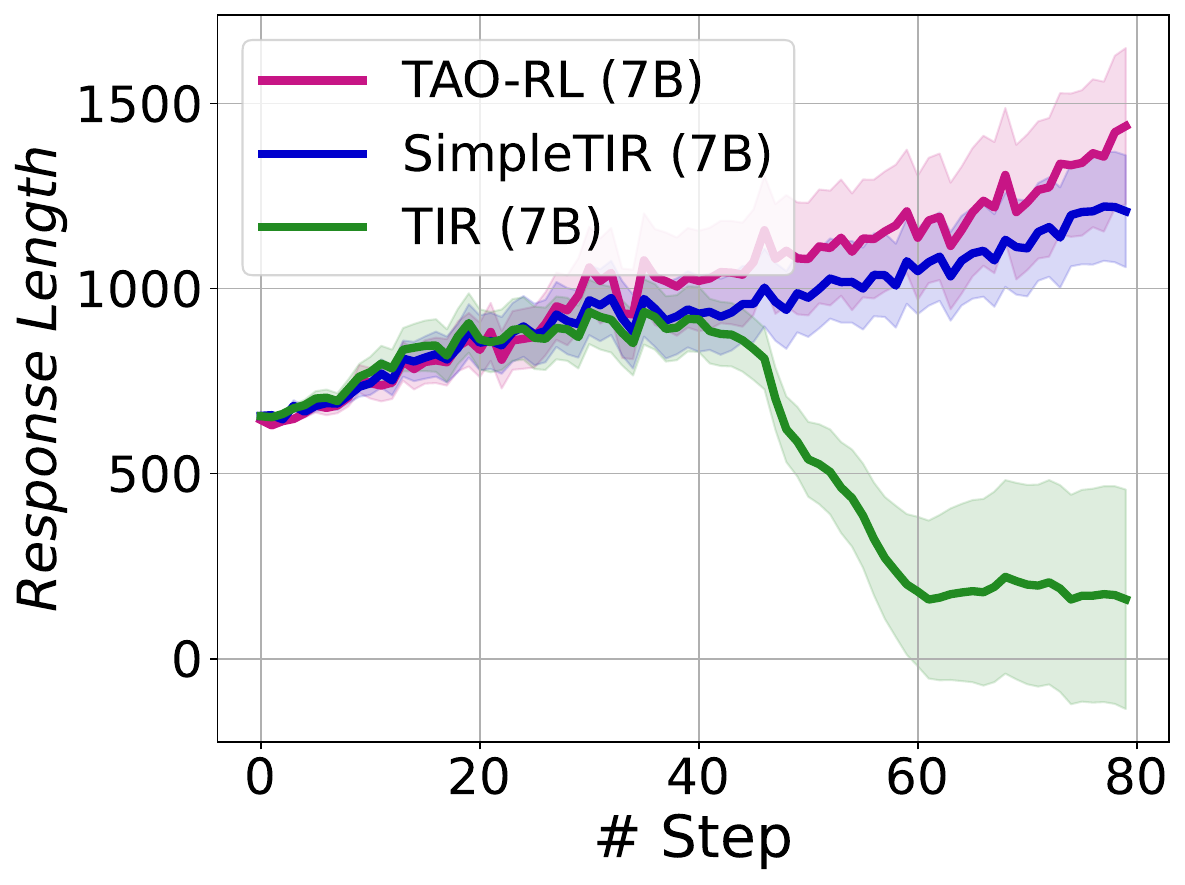} 
  \caption{Comparison of SimpleTIR, TIR and \method across \textit{Avg@16} in AIME 24\&25, and response length under Qwen3-4B and Qwen2.5-7B base models during training. The shadow of line is the standard error.}
  \label{fig:avg_length}
\end{figure}

\subsubsection{Tool-Integrated Efficiency} 

To further evaluate the efficiency of tool use, we analyze four complementary agentic metrics: Valid\_Code, Code\_Line, Box\_Rate, and Tool\_Call, all reported under Qwen2.5-7B base model.

As shown in Table~\ref{tab:res_valid_code}, \method achieves a Valid\_Code average of $99.28\%$, marginally surpassing SimpleTIR's $99.03\%$ and substantially outperforming TIR's $54.17\%$. The near-ceiling Valid\_Code scores of both \method and SimpleTIR confirm that tool-aware trajectory filtering effectively eliminates malformed tool interactions, ensuring that the vast majority of generated code is executable. The large gap over TIR further highlights the importance of filtering tool-execution-failed trajectories from training.

Furthermore, Table~\ref{tab:res_len} reveals a substantial advantage for \method on Code\_Line, with an average of $59.48$ lines per rollout compared to SimpleTIR's $50.03$ $(+9.45)$ and TIR's $9.90$. The gains are particularly pronounced on AIME25 ($+22.11$) and Hmmt25 ($+14.99$), which are among the most challenging benchmarks, suggesting that \method cultivates a tendency to write more elaborate and substantive code when tackling hard reasoning problems, reflecting deeper engagement with external tool-aware computation.

Also, the Box\_Rate and Tool\_Call results in Table~\ref{tab:exp_too_call} further reinforce these findings. \method achieves the highest Box\_Rate across all benchmarks, with an average improvement of nearly $7$ percentage points compared to SimpleTIR, indicating that \method more reliably produces well-formatted final answers following tool interactions. Meanwhile, \method consistently invokes tools more frequently, with Tool\_Calls improvements of up to $0.46$ on AIME25, $0.38$ on Hmmt25 and $0.17$ on Minerva over SimpleTIR. 

To analyze the asymptotic performance of these gains, Fig.~\ref{fig:3d} visualizes the joint evolution of Avg@16 accuracy and Code\_Lines throughout training for both \method and SimpleTIR on 4B and 7B models. The trajectories reveal that \method's average accuracy on AIME25 and MATH500 improves steadily and surpasses SimpleTIR after $60$ training steps, while Code\_Lines remains consistently higher than SimpleTIR across both model scales throughout training. This concurrent growth in accuracy and code complexity demonstrates that \method progressively strengthens the model's tool-use capability during training, and that this deeper tool engagement directly translates into sustained reasoning performance gains.
\textbf{Overall, these results demonstrate that \method not only maintains high tool execution quality, but significantly enhances the model's propensity to engage in deeper, more structured tool-use behavior.}

\subsubsection{Stability Analysis}
\label{sec:Stability}

We analyze the training stability of \method compared with SimpleTIR and TIR across training steps, as shown in Fig.~\ref{fig:avg_length}. In terms of average accuracy on AIME24\&25, \method exhibits a steady and consistent upward trend throughout training, in sharp contrast to TIR, whose performance degrades substantially due to distributional shifts induced by tool integration, and SimpleTIR, which displays fluctuations. These observations confirm that our tool-aware trajectory filtering effectively suppresses the noise introduced by degenerate tool interactions, providing a stable optimization foundation that enables reliable capability growth.
The response length curves further corroborate this conclusion. Across both the 4B and 7B model scales, \method generates substantially longer responses than SimpleTIR and TIR throughout training, reflecting consistently deeper reasoning and more elaborate tool-use behavior. The simultaneous achievement of higher accuracy and better response ability demonstrates that \method drives progressive and stable improvements in both reasoning depth and final performance.

\begin{figure}[t]
  \centering
    \includegraphics[width=0.49\linewidth]{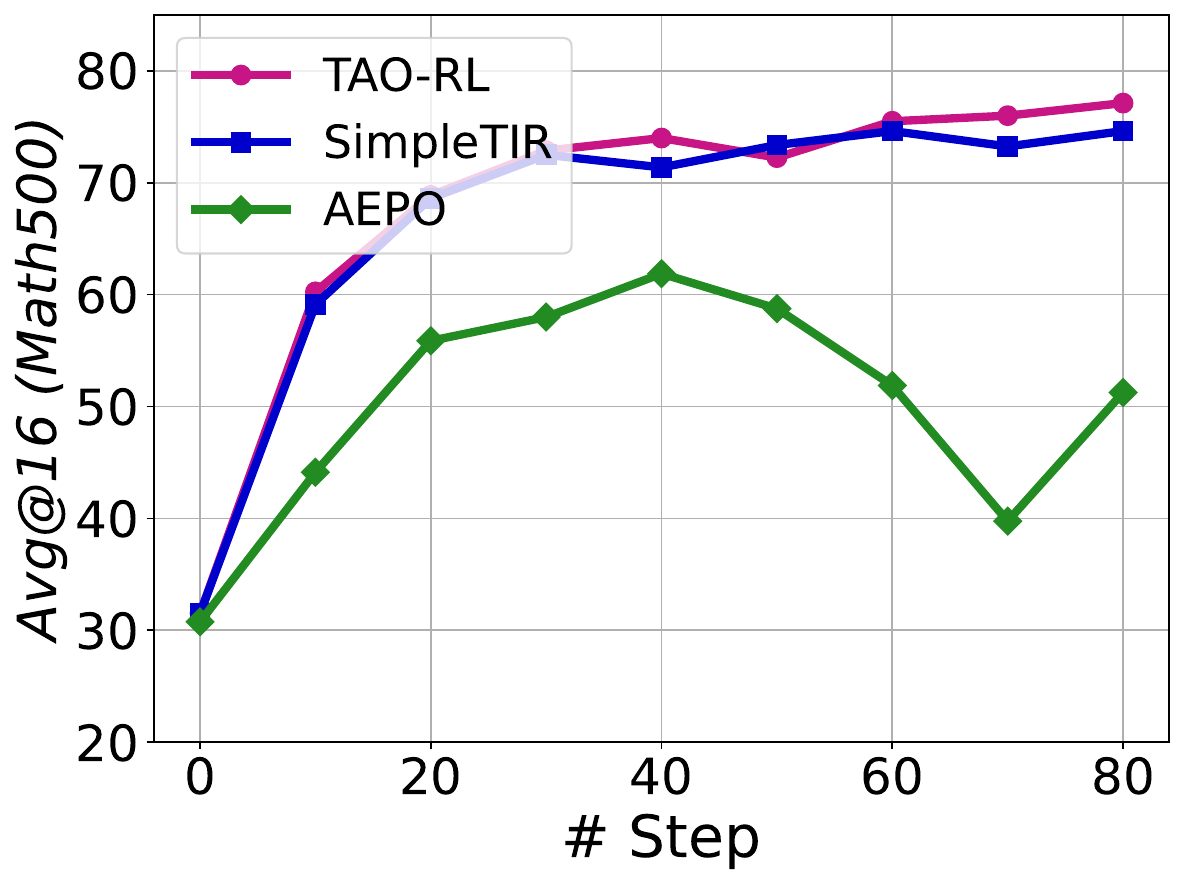} 
  \includegraphics[width=0.49\linewidth]{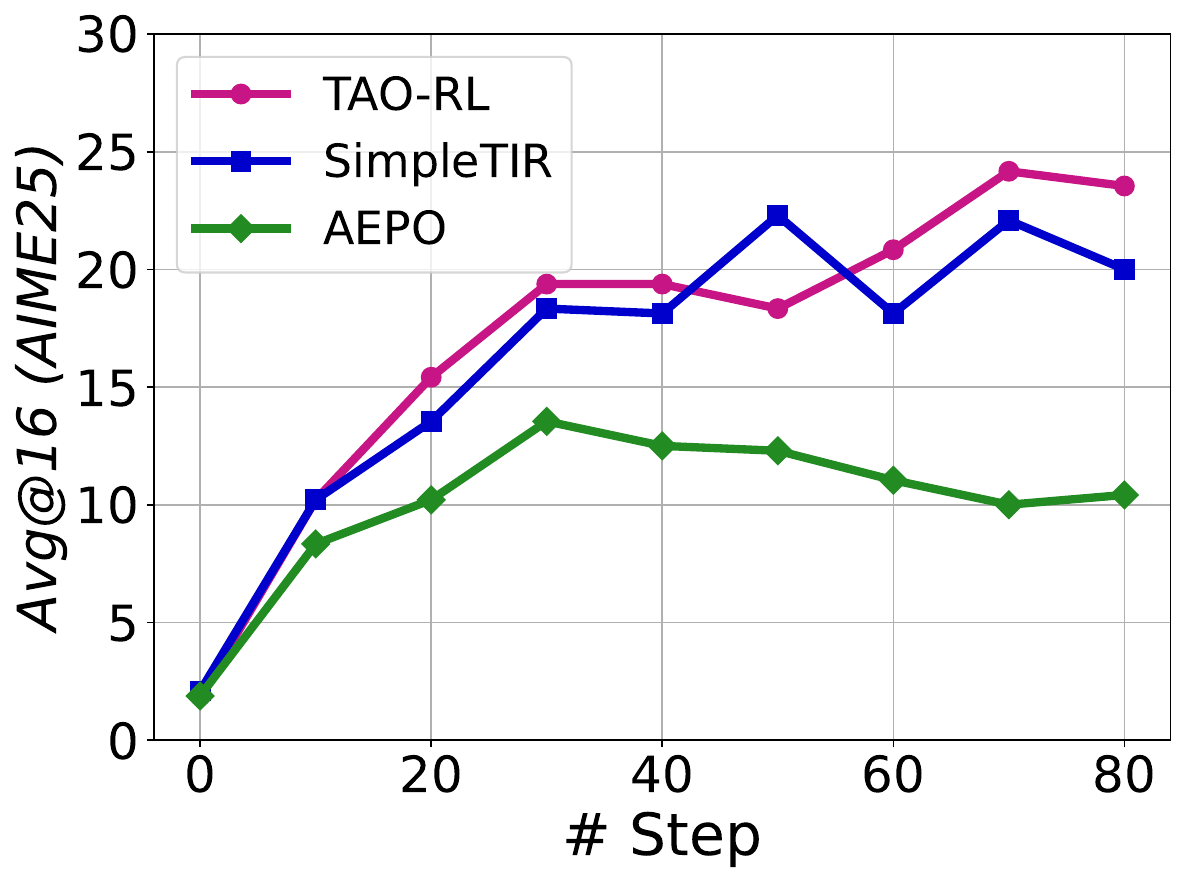} 
  \includegraphics[width=0.49\linewidth]{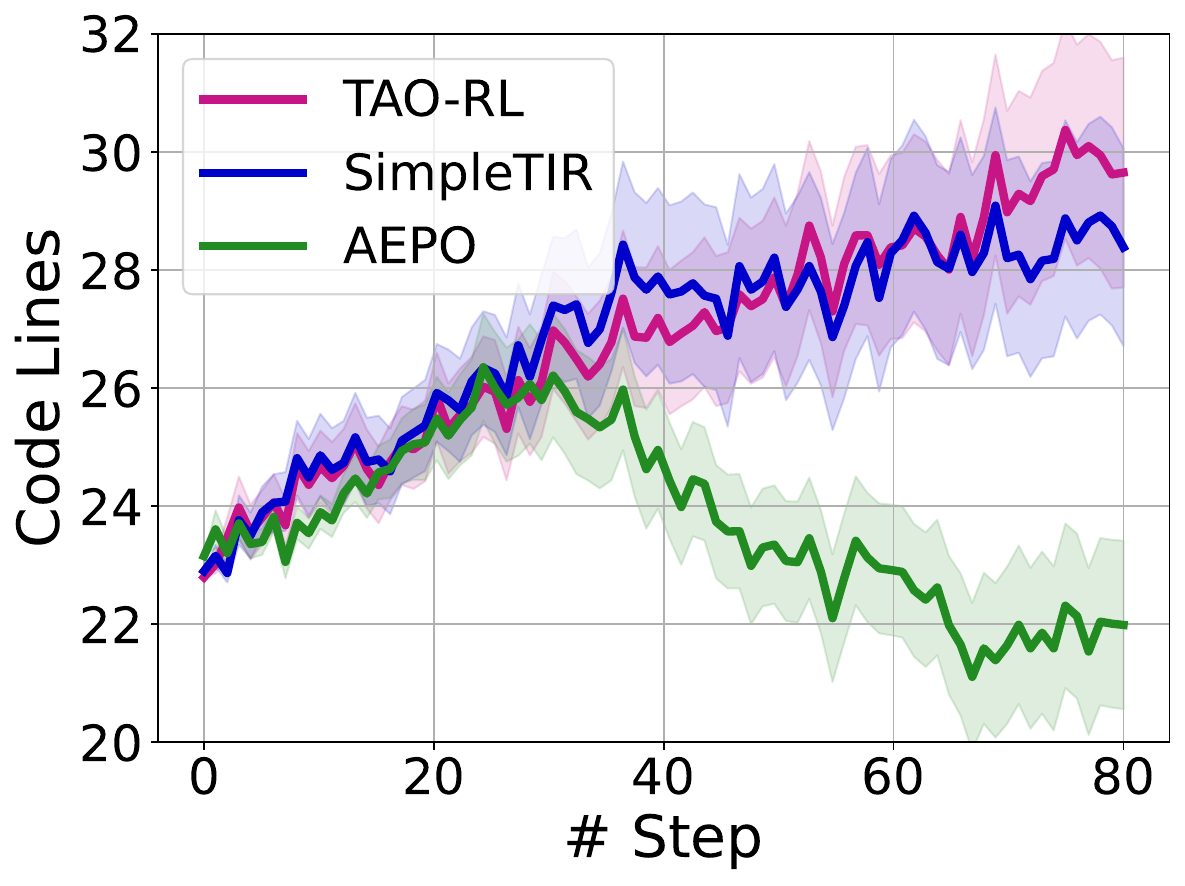} 
  \includegraphics[width=0.49\linewidth]{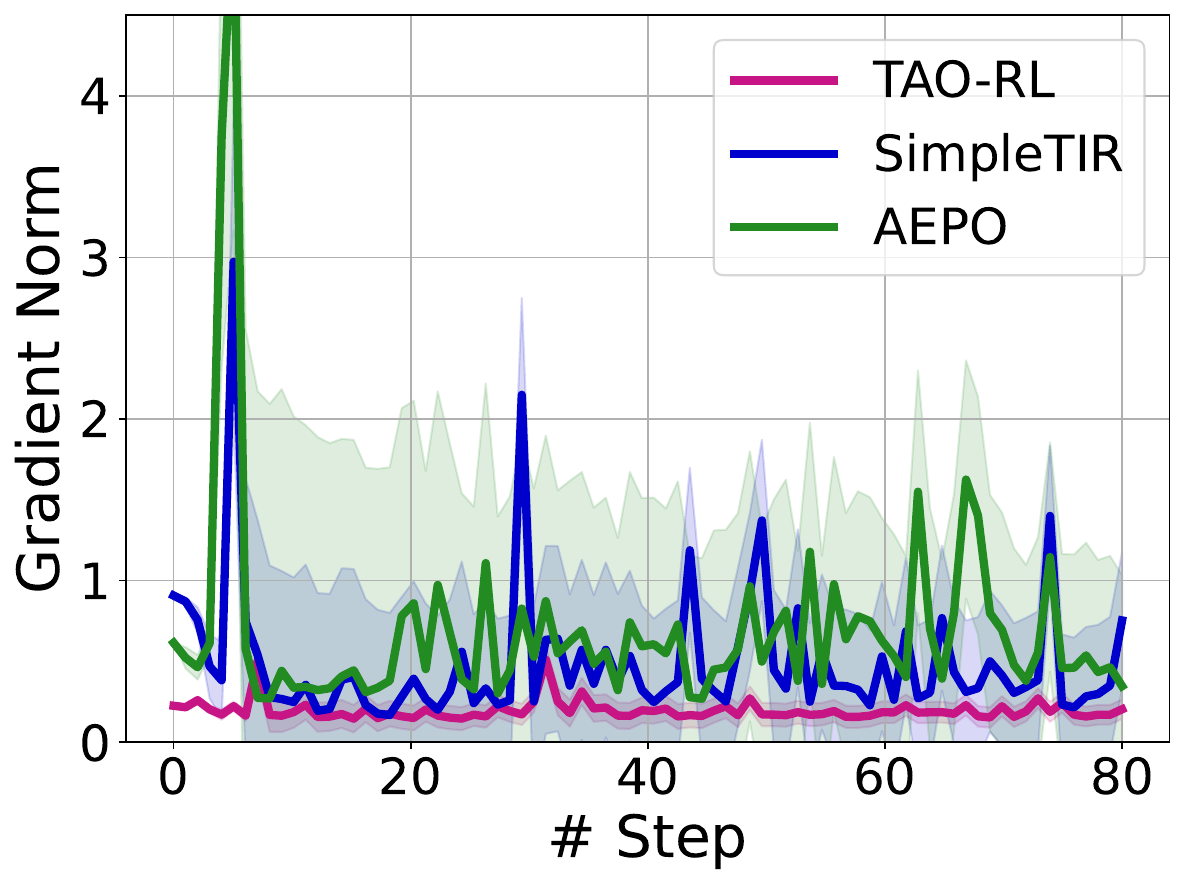} 
  \caption{Comparison of AEPO, SimpleTIR and \method across \textit{Avg@16} in MATH500, AIME 25, code lines and gradient norm under Qwen2.5-7B base model during training.}
  \label{fig:aepo}
\end{figure}

\begin{table}[t]
 \caption{Evaluation of \textit{Tool\_Call} for AEPO and \method under Qwen2.5-7B base model in five benchmarks.} 
\renewcommand{\arraystretch}{1.2}
\setlength{\tabcolsep}{5pt} 
\centering
\begin{tabular}{lccccc}
\toprule
                          & \multicolumn{5}{c}{\textit{{Tool\_Call}}}                                                     \\ \cmidrule(lr){2-6} 
\textbf{Method}            & \textbf{AIME24} & \textbf{AIME25} & \textbf{AMC23}  & \textbf{MATH500} & \textbf{Minerva}   \\ \bottomrule
\textbf{AEPO}                  & 3.59   &   3.25  & 3.22 &  2.97      &       3.16       \\
\rowcolor{blue!10} \textbf{\method} &      2.26   &  2.35      &    1.79   &    1.45    &    1.75                   \\ 
\quad $\Delta$ &\color{darkred}{-1.33 } & \color{darkred}{-0.90 } & \color{darkred}{-1.43 } & \color{darkred}{-1.52 } &  \color{darkred}{-1.41}
\\
\bottomrule
\end{tabular}
\label{tab:aepo_tool_call}
\end{table}

\subsubsection{Comparison of AEPO}
\label{sec:aepo}

We further compare \method against AEPO, an exploration-based agentic RL method, as shown in Fig.~\ref{fig:aepo}. Despite its exploration mechanism, AEPO exhibits progressive performance degradation and pronounced gradient norm fluctuations in later training stages, revealing two fundamental limitations: exploration incentives are amplified over unfiltered noisy rollouts, and entropy bonuses are applied uniformly across all token positions without distinguishing the structurally critical post-tool-call tokens from inconsequential ones. In contrast, \method addresses both gaps simultaneously, achieving stable and targeted exploration throughout training.

Notably, as shown in Table~\ref{tab:aepo_tool_call}, AEPO invokes tools substantially more frequently than \method across five benchmarks. Rather than reflecting effective tool use, this excessive tool invocation is symptomatic of uncontrolled exploration: without filtering out degenerate trajectories, the policy is encouraged to invoke tools indiscriminately, resulting in redundant and uninformative tool interactions that ultimately destabilize training and degrade performance. This finding reveals a critical distinction between \textit{quantity} and \textit{quality} of tool use that \textbf{more frequent tool invocations do not translate into better reasoning if the underlying trajectories lack valid environmental grounding.} In contrast, \method achieves higher accuracy with fewer but more purposeful tool calls, demonstrating that tool-aware trajectory filtering is indispensable for converting exploration incentives into genuine reasoning gains.

\begin{figure}[t]
  \centering
    \includegraphics[width=1\linewidth]{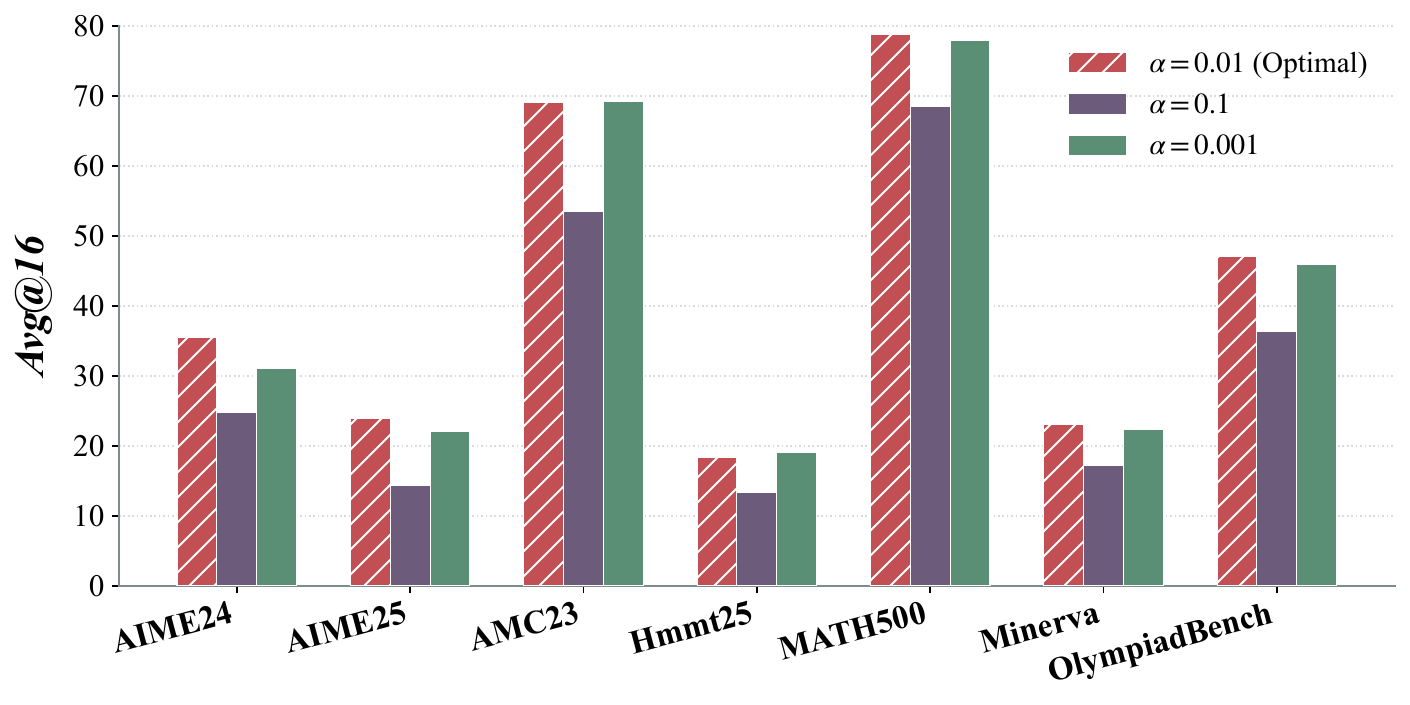} 
  \caption{Hyperparameter analysis of $\alpha$ for \method of \textit{Avg@16} in seven benchmarks under Qwen2.5-7B base model.}
  \label{fig:hps}
\end{figure}

\begin{table}[t]
 \caption{Generalization study on LiveCodeBench benchmark under $7$B base model. We compare \method with other baselines.} 
\renewcommand{\arraystretch}{1.1}
\setlength{\tabcolsep}{12pt} 
\centering
\begin{tabular}{lccc}
\toprule
                          & \multicolumn{3}{c}{\textbf{{LiveCodeBench v5}}}                                                     \\ \cmidrule(lr){2-4} 
\textbf{Method}                    & \textit{Pass@16} &  \textit{Len@16} & \textit{Code\_Line}  \\ \bottomrule
\textbf{Base Model}                      &  14.20   &  1653.78    &   2.76  \\
\textbf{TIR}    &  25.91   &   302.45     &   4.11    \\
\textbf{AEPO}   & 43.41    &   1697.81     &  19.95      \\
\textbf{SimpleTIR}                      &   \underline{43.86}  &   \underline{1733.98}     &    \underline{28.08}                                    \\
\rowcolor{blue!10} \textbf{\method} &     \textbf{45.34}    &  \textbf{1971.88}      &   \textbf{28.27}                                      \\ 
\quad $\Delta$ & \color{darkgreen}{+1.48 } &\color{darkgreen}{+237.90 } & \color{darkgreen}{+0.19 } 
\\
\bottomrule
\end{tabular}
\label{tab:generalizaion}
\end{table}

\subsubsection{Generalization Analysis}

To evaluate the out-of-distribution generalization capability of \method, we conduct experiments on LiveCodeBench v5, a challenging code generation benchmark that differs substantially from the mathematical reasoning tasks used during training. As shown in Table~\ref{tab:generalizaion}, \method achieves the best performance across all three metrics. In terms of Pass@16, \method obtains $45.34\%$, surpassing the strongest baseline SimpleTIR by $1.48$ percentage points, and outperforming AEPO and TIR by considerable margins. In terms of Len@16, \method generates substantially longer responses than all baselines ($237.90$ over SimpleTIR), indicating that the deeper tool-use behaviors cultivated during training transfer effectively to the code generation domain. The Code\_Line metric further confirms this: \method achieves $28.27$ lines per rollout, comparable to SimpleTIR, demonstrating that the quality and depth of code generation generalizes well beyond the training distribution.

\begin{figure}[t]
  \centering
    \includegraphics[width=1\linewidth]{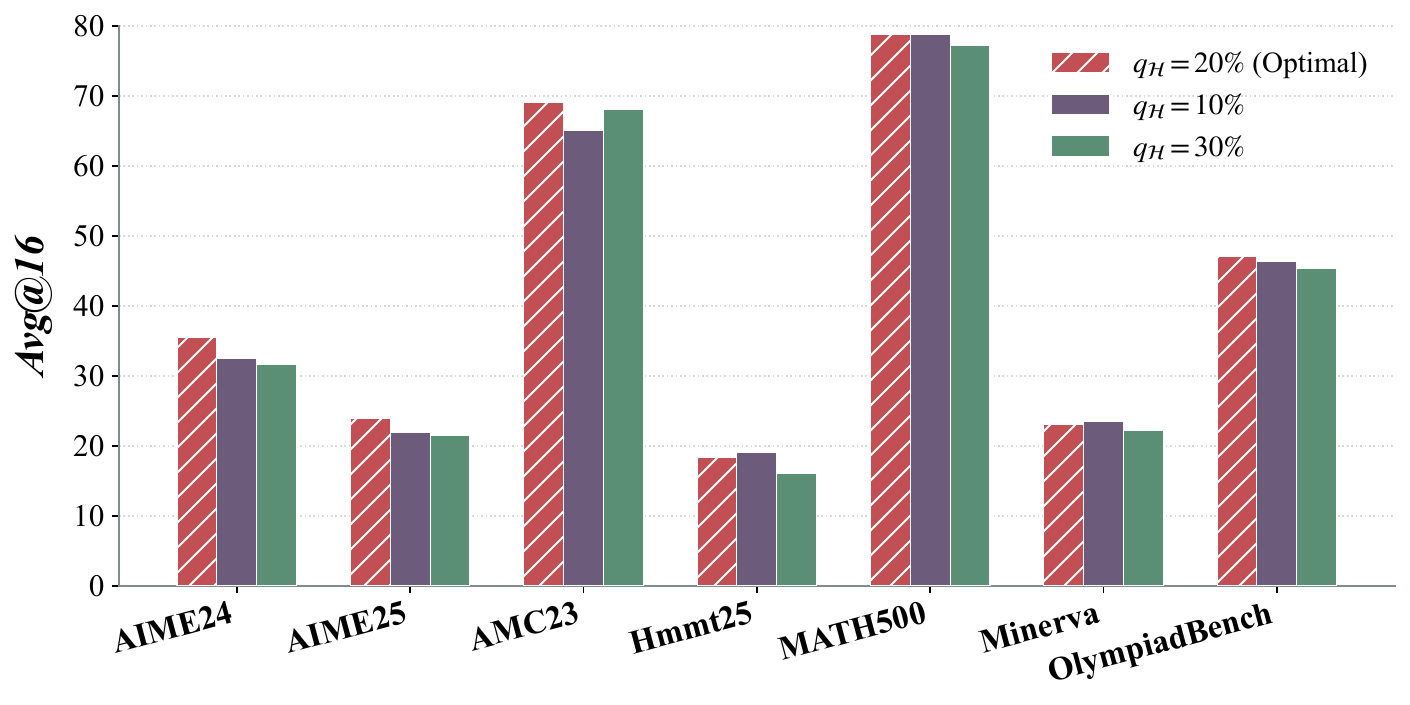} 
  \caption{Hyperparameter analysis of $q_{\mathcal{H}}$ for \method of \textit{Avg@16} in seven benchmarks under Qwen2.5-7B base model.}
  \label{fig:hps_percent}
\end{figure}

\subsubsection{Hyperparameter Analysis}
\label{sec:hps}
Fig.~\ref{fig:hps} presents the sensitivity analysis of the entropy bonus coefficient $a$ across seven benchmarks under Qwen2.5-7B. When $a$ is set to $0.1$, performance degrades substantially across all benchmarks, as the excessively large entropy bonus over-amplifies gradients at post-tool-call positions, inducing unstable policy updates. When $a$ is reduced to $0.001$, the model becomes overly conservative: the entropy bonus provides insufficient exploration incentive, and performance falls slightly below the $a=0.01$ setting across most benchmarks. The setting of $a=0.01$ achieves the best balance between exploration and stability, consistently delivering the strongest Avg@16 performance across all seven benchmarks. We therefore adopt $a=0.01$ as the default value in all experimental settings.
Furthermore, as shown in Fig.~\ref{fig:hps_percent}, the setting of $q_{\mathcal{H}}=20\%$ strikes the optimal balance, targeting a focused subset of genuinely uncertain post-tool-call tokens while avoiding over- or under-coverage, and achieves the best Avg@16 performance across four benchmarks. We therefore adopt $q_{\mathcal{H}}=20\%$ as the default gating proportion.

\begin{table*}[h]
 \caption{Ablation study on six benchmarks of \textit{Pass@16} and \textit{Avg@16} under Qwen2.5-7B base model. \textbf{w/ Fil} denotes adding tool-aware trajectory filtering alone to TIR; \textbf{w/ En} denotes adding entropy-guided exploration with degenerate trajectory removal to TIR; \method combines both components. We bold the best results and underline the sub-optimal results.}
\renewcommand{\arraystretch}{1.2}
\setlength{\tabcolsep}{4pt} 
    \centering 
\begin{tabular}{lcccccccccccc}
\toprule
\textbf{}                          & \multicolumn{2}{c}{\textbf{AIME24}} & \multicolumn{2}{c}{\textbf{AIME25}} & \multicolumn{2}{c}{\textbf{AMC23}} & \multicolumn{2}{c}{\textbf{MATH500}} & \multicolumn{2}{c}{\textbf{OlympiadBench}} & \multicolumn{2}{c}{\textbf{Hmmt25}} 
\\ \cmidrule(lr){2-3} \cmidrule(lr){4-5} \cmidrule(lr){6-7} \cmidrule(lr){8-9} \cmidrule(lr){10-11} \cmidrule(lr){12-13}
\textbf{Method}      & \textit{Avg@16}               & \textit{Pass@16}  & \textit{Avg@16} & \textit{Pass@16}  & \textit{Avg@16} & \textit{Pass@16} & \textit{Avg@16} & \textit{Pass@16}  & \textit{Avg@16}  & \textit{Pass@16}     & \textit{Avg@16}     & \textit{Pass@16}  \\ \bottomrule
 \textbf{Base model}
 &   5.00    &     36.67
 &   3.96    &     26.67    
 &   21.56   &     80.00      
 &   31.99   &     88.00    
 &   13.80   &     55.00         
 &   6.67    &     26.67        \\
\textbf{TIR}
&    10.63   &     40.00 
&    10.42   &     20.00       
&    28.91   &     70.00      
&    42.11   &     79.80           
&    20.88   &     49.80 
&    11.04   &     23.33        \\
\quad \textbf{w/ \emph{Fil}} 
&    29.17    &    63.33 
&    21.25    &    \underline{40.00}   
&    62.81    &    90.00 
&    76.89    &    \underline{94.80}
&    42.98    &    68.60   
&     16.25   &    \underline{43.33}        \\
\quad \textbf{w/ \emph{En}}
&    \textbf{35.83}   &     \underline{66.67}
&    \underline{22.29}   &     36.67      
&    \underline{66.25}   &     \underline{95.00} 
&    \underline{78.55}   &     \textbf{95.60}     
&    \underline{45.43}   &     \textbf{70.60}     
&    \textbf{18.75}   &     \textbf{46.67}     \\
\rowcolor{blue!10}
\textbf{\method} 
&    \underline{35.42}    &   \textbf{73.33} 
&    \textbf{23.96}    &   \textbf{46.67} 
&    \textbf{69.06}    &   \textbf{97.50} 
&    \textbf{78.80}    &   \textbf{95.60}
&    \textbf{46.99}    &   \underline{70.00}  
&    \underline{18.33}    &   \textbf{46.67} \\ \bottomrule 
\end{tabular}
\label{tab:res_ablation}
\end{table*}

\subsubsection{Ablation Study}
\label{sec:ablation_study}

Table~\ref{tab:res_ablation} presents the ablation results under Qwen2.5-7B, isolating the contribution of each component of \method.
\emph{Effect of Trajectory Filtering (w/ Fil)}: Adding tool-aware trajectory filtering alone to TIR yields dramatic improvements across all benchmarks. On AIME24, Avg@16 improves from $10.63$ to $29.17$, and Pass@16 from $40.00$ to $63.33$. These results confirm that removing tool-execution-failed and discriminability-degenerate trajectories substantially stabilizes training.
\emph{Effect of Entropy-Guided Exploration (w/ En)}: Incorporating entropy-guided advantage reshaping, while also removing trajectories with code execution format errors, further improves performance over w/ Fil on most benchmarks. On AIME24, Avg@16 reaches $35.83$, and notable gains are also observed on OlympiadBench ($+2.45$) and Hmmt25 ($+2.50$). This demonstrates that concentrating exploration incentives at high-entropy post-tool-call tokens effectively raises the reasoning ceiling.
Combining both components consistently achieves the best or near-best results across all benchmarks.

\begin{table*}[t]
\centering
\caption{Case Study on AIME 2024 Problem 18. \method successfully establishes a mathematical abstraction (log-space linear system) and leverages \textit{post-tool-call exploration} to recover from an intermediate \texttt{NameError}, whereas SimpleTIR fails.}
\label{tab:case_study}
\begin{tabular}{p{0.48\textwidth} p{0.48\textwidth}}
\toprule
\multicolumn{2}{p{0.96\textwidth}}{\textbf{Problem:} Given positive real numbers $x, y, z$ satisfying $\log_2(\frac{x}{yz}) = \frac{1}{2}, \log_2(\frac{y}{xz}) = \frac{1}{3}, \log_2(\frac{z}{xy}) = \frac{1}{4}$, find $|\log_2(x^4 y^3 z^2)|$. Express your answer as $m/n$, and output $m+n$. \hfill \textbf{Ground Truth:} $33$} \\
\midrule
\rowcolor{gray!15}
\textbf{SimpleTIR (Baseline)} & \textbf{\method (Ours)} \\
\midrule
\begin{minipage}[t]{\linewidth}
Given: $\log_2(\frac{x}{yz}) = \frac{1}{2} \rightarrow \frac{x}{yz} = 2^{1/2} \dots$ \\
Which can be rewritten directly as $x = 2^{1/2}yz$. \\
Similarly for $y$ and $z$. We can now solve this system of \\
equations with sympy to find numerical roots \dots \\
\vspace{0.2em} \\
We discard negative solutions. We will now use these positive \\
values to compute the expression $|\log_2(x^4 y^3 z^2)|$ \dots
\begin{lstlisting}[language=Python, escapeinside={(*@}{@*)}, commentstyle=\color{gray}]
# Use the explicitly identified positive solutions
x_val, y_val, z_val = 0.8169, 0.7711, 0.7491
expression = x_val**4 * y_val**3 * z_val**2
log_value = sp.log(expression, 2)
(*@\textcolor{red}{\ttfamily abs\_log\_value = sp.Abs(log\_value)}@*)
\end{lstlisting}
\vspace{0.1em}
\textcolor{execresult}{\textbf{NameError: name 'sp' is not defined}} \\
\vspace{0.01em} \\
It seems there was an issue because the log function was
not simplified to a numerical value \dots Let's 
recompute using sympy's rational exact numbers.
\begin{lstlisting}[language=Python, escapeinside={(*@}{@*)}, commentstyle=\color{gray}]
import sympy as sp
# Define irrational numbers for precision
sqrt2 = sp.Rational(2).sqrt()
cbrt2 = 2**(sp.Rational(1, 3))
# Attempt to simplify the values symbolically
(*@\textcolor{red}{\ttfamily x\_val = sqrt2 * sp.Rational(1) / cbrt2}@*)
\end{lstlisting}
\vspace{0.1em}
\textcolor{execresult}{\textbf{AttributeError: 'Integer' object has no attribute 'sqrt'}} \\
\vspace{0.2em} \\
Failed to handle rational/irrational parts carefully \dots \\
The correct exact fraction cannot be retrieved. \\
\vspace{0.3em} \newline \fcolorbox{red}{white}{\textbf{Final Answer: WRONG}} \vspace{0.5em}
\end{minipage} &
\begin{minipage}[t]{\linewidth}
Using properties of logarithms, we rewrite the system: \\
\textbf{$\log_2(x) - (\log_2(y) + \log_2(z)) = \frac{1}{2} \dots$} \\
Let $\log_2(x)=a, \log_2(y)=b, \log_2(z)=c \implies a - (b+c) = \frac{1}{2}$. \\
Similarly for $b$ and $c$. Let's solve this linear system \dots \\
\vspace{0.2em} \\
\textbf{We know $\log_2(x^4 y^3 z^2) = 4a + 3b + 2c \dots$} \\ Let's 
calculate this expression to find its absolute value.
\begin{lstlisting}[language=Python, escapeinside={(*@}{@*)}, commentstyle=\color{gray}]
# Extract the exactly solved symbolic variables
a_val, b_val = solution[a], solution[b]
c_val = solution[c]
expr = 4*a_val + 3*b_val + 2*c_val
(*@\textbf{\ttfamily frac = sp.Rational(abs(expr)).limit\_denominator()}@*)
\end{lstlisting}
\vspace{0.1em}
\textcolor{execresult}{\textbf{NameError: name 'solution' is not defined}} \\
\vspace{0.01em} \\
\textbf{It seems there was an issue with accessing the solution directly.} \textbf{Let's recompute the value of $4a + 3b + 2c$ } \textbf{using exact Rational types \dots}
\begin{lstlisting}[language=Python, escapeinside={(*@}{@*)}, commentstyle=\color{gray}]
import sympy as sp
# Re-evaluate the linear system with exact Rational types
eq1 = sp.Eq(a - (b+c), sp.Rational(1, 2))
# ... [eq2, eq3 exact definitions omitted] ...
solution = sp.solve((eq1, eq2, eq3), (a, b, c))
(*@\textbf{\ttfamily expr\_val = 4*solution[a] + 3*solution[b] + 2*solution[c]}@*)
\end{lstlisting}
\vspace{0.1em}
\textcolor{execresult}{\textbf{Result: (25/8, -25/8)}} \\
\vspace{0.2em} \\
The absolute value of the expression is $25/8$. \\
Thus $m=25$ and $n=8$, yielding $m+n=33$. \\
\vspace{0.3em} \newline \fcolorbox{green!50!black}{white}{\textbf{Final Answer: 33}} \vspace{0.5em}
\end{minipage} \\
\bottomrule
\end{tabular}
\end{table*}

\vspace{-2mm}
\subsection{Case Study}
\label{sec:case_study}

To qualitatively demonstrate the effectiveness of proposed \method framework, we conduct a comparative case study against the SimpleTIR on a representative logarithmic system problem from the AIME~2024 test set, as shown in Table~\ref{tab:case_study}. When faced with this complex system, SimpleTIR naively relies on a direct numerical approximation and treats code execution as a terminal oracle, ultimately failing due to compounding execution errors and a lack of robust error recovery, a preventable \texttt{NameError} derails its pipeline, and subsequent attempts to recover only deepen the confusion, culminating in an unrecoverable symbolic \texttt{AttributeError}. In contrast, \method first establishes a proper mathematical abstraction by formulating a log-space linear system before generating any code. Crucially, this problem highlights the power of \method's robust post-tool-call exploration: when the initial Python execution unexpectedly triggers a \texttt{NameError}, \method does not halt. Instead, it explicitly diagnoses the execution state and dynamically restructures its subsequent code to leverage exact rational types via \texttt{SymPy}, seamlessly bridging abstract logic with precise symbolic execution to successfully recover from the error. We provide more experimental analysis in Appendix D.

\section{Conclusion and Limitation}
\label{sec:conclusion}

In this work, we present \method, a unified framework for efficient agentic reinforcement learning that addresses the fundamental tension between training stability and effective exploration introduced by tool integration. \method operates jointly at two levels: at the data level, tool-aware trajectory filtering removes degenerate rollouts by jointly considering tool execution validity and answer discriminability; at the algorithmic level, entropy-guided advantage reshaping concentrates exploration incentives at high entropy post-tool-call tokens. Extensive experiments across challenging reasoning benchmarks and three model scales consistently demonstrate that \method outperforms existing agentic RL methods in both reasoning performance and training stability, while also cultivating deeper and more structured tool-use behaviors as evidenced by substantial gains in response length, code lines, and tool invocation quality.

\textbf{Limitation and Future Work:}
\method has several limitations worth noting. The hyperparameters governing the entropy bonus and the gating proportion require tuning. Also, \method considers only code interpreter as the external tool, which represents a relatively constrained form of tool interaction. Real-world agentic scenarios involve more diverse and compositional tool ecosystems, such as search engines, databases, and multi-step calls. The scalability of \method to broader agentic settings involving diverse tool types, multi-turn interactions, or open-ended tasks remains to be explored in future work.


\bibliographystyle{IEEEtran}
\bibliography{tpami.bib}

\clearpage
\onecolumn

\appendix
\subsection{Broader Impact}

\method addresses a fundamental challenge in agentic reinforcement learning: achieving stable and effective policy optimization in the presence of tool interaction. By coupling tool-aware trajectory filtering with entropy-guided exploration, \method provides a principled framework for improving reasoning capabilities while maintaining training stability. This has promising implications for developing more capable and reliable agentic systems across diverse domains, from mathematical problem-solving to code generation and complex multi-step decision-making tasks.

The unified framework that jointly addresses data quality and algorithmic exploration offers a generalizable paradigm extending beyond the specific setting of tool-integrated reasoning. The principle of filtering degenerate training signals before applying targeted exploration incentives is broadly applicable to any RL setting where noisy or uninformative rollouts risk destabilizing policy optimization. Similarly, the idea of concentrating exploration at structurally critical positions, rather than applying uniform incentives across all tokens, provides a general design principle for more efficient policy gradient methods in long-horizon sequential decision-making.

From a societal perspective, more stable and efficient agentic RL training reduces the computational resources required to develop capable reasoning models, lowering the barrier to research and deployment of LLM-based agents. More reliable tool-use behavior also improves the trustworthiness of agentic systems in real-world applications, where erratic tool invocations can have unintended downstream consequences. We hope the insights from \method inspire further research into structured, tool-aware training paradigms for the broader agentic AI community.



\subsection{Theoretical Analysis}
\label{sec:app theory}


We first provide a formal proof that restricting the policy gradient expectation to $\mathcal{O}_{\text{valid}}$ strictly improves the gradient signal-to-noise ratio (SNR) over standard GRPO.

\textbf{Setup.} Let $\mathcal{O} = \{o^i\}_{i=1}^G$ denote the full set of rollout trajectories, partitioned as $\mathcal{O} = \mathcal{O}_{\text{valid}} \cup \mathcal{O}_{\text{deg}}$, where $|\mathcal{O}_{\text{valid}}| = n$ and $|\mathcal{O}_{\text{deg}}| = G - n$. Define the per-trajectory gradient as:
\begin{equation*}
    \mathbf{g}_i = \sum_{t=1}^{|o^i|} \nabla_\theta \log\pi_\theta(o^i_t | s_t) \cdot \hat{A}^i_t.
\end{equation*}

The GRPO and \method gradient estimators are then:
\begin{equation*}
    \nabla\mathcal{J}_{\text{GRPO}} = \frac{1}{G}\sum_{i=1}^{G}\mathbf{g}_i, \qquad \nabla\mathcal{J}_{\method} = \frac{1}{n}\sum_{i\in\mathcal{O}_{\text{valid}}}\mathbf{g}_i.
\end{equation*}

We establish two key properties of $\mathbf{g}_i$ for $o^i \in \mathcal{O}_{\text{deg}}$:

\textit{Property 1 (Zero mean signal).} For queries with $\rho(q)\in\{0,1\}$, all rollouts receive identical rewards, so the group-normalized advantage satisfies $\hat{A}^i_t = 0$ for all $t$, giving $\mathbf{g}_i = \mathbf{0}$. For trajectories with all tool calls failed, the policy receives no valid environmental grounding, so the expected gradient contribution is approximately zero: $\mathbb{E}[\mathbf{g}_i] \approx \mathbf{0}$.

\textit{Property 2 (Non-zero variance).} Despite carrying zero expected signal, degenerate trajectories exhibit non-zero gradient variance $\sigma^2_{\text{deg}} = \mathrm{Var}[\mathbf{g}_i] > 0$ due to stochastic sampling and pseudo-random advantage fluctuations induced by failed tool execution.

\emph{Expected Gradient.} Using Property 1:
\begin{equation*}
\mathbb{E}[\nabla\mathcal{J}_{\text{GRPO}}] = \frac{1}{G}\sum_{i\in\mathcal{O}_{\text{valid}}}\mathbb{E}[\mathbf{g}_i] = \frac{n}{G}\cdot\mathbb{E}[\nabla\mathcal{J}_{\method}].
\end{equation*}

Hence, $\|\mathbb{E}[\nabla\mathcal{J}_{\text{GRPO}}]\|^2 = \frac{n^2}{G^2}\|\mathbb{E}[\nabla\mathcal{J}_{\method}]\|^2$.

\emph{Gradient Variance.} Let $\sigma^2_{\text{valid}} = \frac{1}{n}\sum_{i\in\mathcal{O}_{\text{valid}}}\mathrm{Var}[\mathbf{g}_i]$. Assuming independence across trajectories:
\begin{equation*}
    \mathrm{Var}[\nabla\mathcal{J}_{\text{GRPO}}] = \frac{n\sigma^2_{\text{valid}} + (G-n)\sigma^2_{\text{deg}}}{G^2}, \qquad
    \mathrm{Var}[\nabla\mathcal{J}_{\method}] = \frac{\sigma^2_{\text{valid}}}{n}.
\end{equation*}

\emph{SNR Comparison.} Denoting $\mu^2 = \|\mathbb{E}[\nabla\mathcal{J}_{\method}]\|^2$:
\begin{equation*}
    \text{SNR}_{\method} = \frac{\mu^2}{\sigma^2_{\text{valid}}/n} = \frac{n\mu^2}{\sigma^2_{\text{valid}}},
\end{equation*}
\begin{equation*}
    \text{SNR}_{\text{GRPO}} = \frac{(n^2/G^2)\mu^2}{(n\sigma^2_{\text{valid}} + (G-n)\sigma^2_{\text{deg}})/G^2} = \frac{n^2\mu^2}{n\sigma^2_{\text{valid}} + (G-n)\sigma^2_{\text{deg}}}.
\end{equation*}

Taking the ratio:
\begin{equation*}
    \frac{\text{SNR}_{\method}}{\text{SNR}_{\text{GRPO}}} = \frac{n\mu^2/\sigma^2_{\text{valid}}}{n^2\mu^2/(n\sigma^2_{\text{valid}} + (G-n)\sigma^2_{\text{deg}})} = 1 + \frac{(G-n)\sigma^2_{\text{deg}}}{n\sigma^2_{\text{valid}}},
\end{equation*}
since $G > n$ (degenerate trajectories exist) and $\sigma^2_{\text{deg}} > 0$ (Property 2), we conclude:
\begin{equation*}
    \frac{\text{SNR}_{\method}}{\text{SNR}_{\text{GRPO}}} > 1 \quad \Longrightarrow \quad \text{SNR}_{\method} > \text{SNR}_{\text{GRPO}},
\end{equation*}
which proves that trajectory filtering strictly improves the gradient SNR by simultaneously preserving the expected gradient signal and eliminating the variance contribution of degenerate rollouts.

\textbf{Proposition 1} (Logit Change in Policy Gradient~\cite{NIPS2001_4b86abe4,cui2025entropy}). \textit{When updating the policy via Policy Gradient with learning rate $\eta$, the logit difference satisfies:}
\begin{equation*}
\theta^{k+1}_{s_t,o_t} - \theta^k_{s_t,o_t} = \eta\cdot\pi^k_\theta(o_t|s_t)\cdot A_t.
\end{equation*}
\begin{proof}[Proof adapted from~\cite{cui2025entropy}]
\begin{equation*}
\begin{aligned}
& \theta^{k+1}_{s_t,o_t} - \theta^k_{s_t,o_t} \\
&= \eta \cdot \nabla_{\theta_{s_t,o_t}} \mathcal{J}(\theta) \\
&= \eta \cdot \mathbb{E}_{o'_t \sim \pi^k_{\theta}(\cdot | s_t)} \left[ \nabla_{\theta_{s_t,o_t}} \log \pi^k_{\theta}(o'_t | s_t) \cdot A_t \right] \\
&= \eta \cdot \sum_{o'_t \in \mathcal{O}} \left[ \pi^k_{\theta}(o'_t | s_t) \cdot \left( \mathbbm{1}\{o_t = o'_t\} - \pi^k_{\theta}(o_t | s_t) \right) \cdot A_t \right] \\
&= \eta \cdot \pi^k_{\theta}(o_t | s_t) \cdot \left[ A_t - \mathbb{E}_{o'_t \sim \pi^k_{\theta}(\cdot | s_t)}[A_t] \right] \\
&= \eta \cdot \pi^k_{\theta}(o_t | s_t) \cdot A_t.
\end{aligned}
\end{equation*}
The last step holds because the GRPO advantage $\hat{A}_t$ is group-normalized and thus has zero mean by construction, i.e., $\mathbb{E}_{o'_t \sim \pi^k_\theta}[A_t] = 0$.
\end{proof}

\textbf{Proposition 2} (Logit Change under \method). \textit{When updating the policy via $\mathcal{J}_{\method}$ with the reshaped advantage $\tilde{A}_t = \hat{A}_t \cdot (1 + a \cdot g_t \cdot \tilde{A}_t^{\Delta\mathcal{H}})$ and learning rate $\eta$, the logit difference satisfies:}
\begin{equation*}
\theta^{k+1}_{s_t,o_t} - \theta^k_{s_t,o_t} =
\underbrace{\eta\cdot\pi^k_\theta(o_t|s_t)\cdot\hat{A}_t}_{\text{basic update}}
+\underbrace{\eta\cdot a\cdot g_t\cdot\pi^k_\theta(o_t|s_t)\cdot\hat{A}_t\cdot\tilde{A}_t^{\Delta\mathcal{H}}}_{\text{exploration bonus}}.
\end{equation*}
\begin{proof}
Following the same derivation as Proposition 1, with the reshaped advantage $\tilde{A}_t$ substituted for $A_t$:
\begin{equation*}
\begin{aligned}
& \theta^{k+1}_{s_t,o_t} - \theta^k_{s_t,o_t} \\
&= \eta \cdot \nabla_{\theta_{s_t,o_t}} \mathcal{J}_{\method}(\theta) \\
&= \eta \cdot \mathbb{E}_{o'_t \sim \pi^k_{\theta}(\cdot | s_t)} \left[ \nabla_{\theta_{s_t,o_t}} \log \pi^k_{\theta}(o'_t | s_t) \cdot \tilde{A}_t \right] \\
&= \eta \cdot \pi^k_{\theta}(o_t | s_t) \cdot \tilde{A}_t \quad \text{(by Proposition 1)}\\
&= \eta \cdot \pi^k_{\theta}(o_t | s_t) \cdot \hat{A}_t \cdot \left(1 + a \cdot g_t \cdot \tilde{A}_t^{\Delta\mathcal{H}}\right) \\
&= \eta\cdot\pi^k_\theta(o_t|s_t)\cdot\hat{A}_t
+ \eta\cdot a\cdot g_t\cdot\pi^k_\theta(o_t|s_t)\cdot\hat{A}_t\cdot\tilde{A}_t^{\Delta\mathcal{H}}.
\end{aligned}
\end{equation*}
For tokens outside $\mathcal{M}_{\text{post}}$, $g_t = 0$ and the update reduces to the standard Policy Gradient step $\eta\cdot\pi^k_\theta(o_t|s_t)\cdot\hat{A}_t$. For gated tokens ($g_t=1$), the additional exploration bonus term amplifies the logit update proportionally to the token's normalized entropy deviation $\tilde{A}_t^{\Delta\mathcal{H}}$, concentrating stronger optimization pressure at the most uncertain post-tool-call positions.
\end{proof}

\subsection{Experimental Setup}
\label{sec:app setup}
\subsubsection{Experimental Setup}

To ensure reproducibility, we provide a detailed description of the training and evaluation hyperparameters in Table~\ref{tab:train_hyper} and Table~\ref{tab:eval_hyper}. Parameters not explicitly listed follow the default configurations of the VeRL~\cite{sheng2025hybridflow} framework. Hyperparameters specific to individual baselines are configured strictly according to their original papers, as described in Section~\ref{sec:experiments}. Codes are available at: \codesite.

As shown in Table~\ref{tab:train_hyper}, we conduct RL training on three models of different sizes: Qwen2.5-1.5B, Qwen3-4B-Base, and Qwen2.5-7B. For training data, we use the \textbf{Math3-5} dataset from SimpleRL~\cite{zeng2025simplerl} dataset (8K samples) for the 1.5B model, and the \textbf{DAPO-MATH-17K} dataset~\cite{yu2025dapo} for both the 4B and 7B models. All baseline methods are reproduced under identical hyperparameters within the VeRL platform. We set the number of training epochs to $10$ for the 1.5B model, as its relatively weak initial capability requires more training steps to develop stable reasoning and tool-use behaviors, and $5$ for both the 4B and 7B models. The rollout group size $G$ is set to $16$ for all models, with a minimum valid rollout threshold of $4$ per prompt to ensure sufficient contrastive signal. Rejection sampling is enabled with an oversampling multiplier of $2.0$ to compensate for trajectories removed by filtering. The clipping ratio bounds are $\epsilon_{\text{high}} = 0.28$ and $\epsilon_{\text{low}} = 0.20$, with a dual-clip ratio $c = 3.0$. The KL divergence loss and standard entropy regularization coefficient are both set to zero across all settings, as the entropy bonus in \method is applied exclusively via the advantage reshaping mechanism described in Section~\ref{sec:approach}. Tool-output tokens and void turns are masked during policy gradient computation. For \method, the entropy bonus coefficient $\alpha$ is set to $0.01$ and the entropy gating percentile $q_{\mathcal{H}}$ is set to 20\%, consistent with the hyperparameter analysis in Section~\ref{sec:hps}. All experiments are conducted on eight NVIDIA A100 GPUs.

\subsubsection{Benchmarks}
In our experiments, we conduct extensive validation on the following seven challenging mathematical reasoning benchmarks, along with one additional benchmark for generalization evaluation, to comprehensively assess model performance.

\begin{itemize}
    \item \textbf{AIME 2024 \& 2025}~\cite{codeforcesamerican}: A collection of 30 problems from the American Invitational Mathematics Examination 2024/2025, a prestigious high school mathematics competition featuring challenging multi-step problems across various mathematical domains.
    
    \item \textbf{AMC 2023}~\cite{amc23}: A set of 40 problems from the 2023 American Mathematics Competitions.
    
    \item \textbf{MATH500}~\cite{hendrycks2measuring}: A 500-problem subset from the MATH dataset covering seven subjects including Algebra, Geometry, Number Theory, and Precalculus at competition mathematics level.
    
    \item \textbf{OlympiadBench}~\cite{he2024olympiadbench}: A challenging benchmark that provides Olympiad-level, bilingual, multimodal scientific problems designed to evaluate advanced mathematical and scientific reasoning in large language models. In our experiments, We only use the OE\_TO\_maths\_en\_COMP subset consisting of 674 open-ended, text-only English mathematics competition problems.
    
    \item \textbf{Minerva}~\cite{lewkowycz2022solving}: Minerva is a benchmark designed to evaluate mathematical and quantitative reasoning capabilities of LLMs. It consists of 272 problems sourced primarily from MIT OpenCourseWare courses, covering advanced STEM subjects such as solid-state chemistry, astronomy, differential equations, and special relativity at the university and graduate level.

    \item \textbf{HMMT 25}~\cite{balunovic_srimatharena_2025}: A challenging dataset sourced from the 2025 Harvard-MIT Mathematics Tournament, featuring highly complex, Olympiad-level mathematics problems. As a recently released competition, it serves as a rigorous, contamination-free benchmark to evaluate the advanced mathematical reasoning and problem-solving capabilities of large language models.

    \item \textbf{LiveCodeBench v5}~\cite{jain2025livecodebench}: A holistic and contamination-free evaluation benchmark for code generation. Version 5 consists of 880 algorithmic problems continuously collected from competitive programming platforms (such as LeetCode, AtCoder, and Codeforces) between May 2023 and January 2025, designed specifically to assess models on unseen, real-world coding challenges.
\end{itemize}

\begin{table}[t]
\centering
\caption{Training hyperparameters used in our experiments.}
\label{tab:train_hyper}
\setlength{\tabcolsep}{5pt}
\renewcommand{\arraystretch}{1.15}
\begin{tabular}{lccc}
\toprule
\textbf{Hyperparameter}
    & \textbf{Qwen2.5-1.5B-Base}
    & \textbf{Qwen3-4B-Base}
    & \textbf{Qwen2.5-7B-Base} \\
\midrule
Training dataset
    & SimpleRL-Math3-5 (8K)
    & DAPO-Math-17K
    & DAPO-Math-17K \\
Training epochs          & 10    & 5     & 5     \\
GRPO rollout size $G$    & 16    & 16    & 16    \\
Min.\ valid rollouts     & 4     & 4     & 4     \\
Oversampling multiplier  & 2.0   & 2.0   & 2.0   \\
Temperature              & 1.0   & 1.0   & 1.0   \\
Learning rate            & 1e-6  & 1e-6  & 1e-6  \\
Max prompt length        & 16000  & 8000  & 8000  \\
Max response length      & 8000  & 4000  & 4000  \\
Max observation length   & 256   & 256   & 256   \\
Max agent turns          & 5     & 5     & 5     \\
PPO mini-batch size      & 128   & 128   & 128   \\
Clip ratio high $\epsilon_{\text{high}}$  & 0.28  & 0.28  & 0.28  \\
Clip ratio low $\epsilon_{\text{low}}$    & 0.20  & 0.20  & 0.20  \\
Dual-clip ratio $c$      & 3.0   & 3.0   & 3.0   \\
KL loss coefficient      & 0     & 0     & 0     \\
Entropy bonus coeff $\alpha$ (\method)
    & 0.01  & 0.01  & 0.01  \\
Entropy gate $q_{\mathcal{H}}$ (\method)
    & 20\%  & 20\%  & 20\%  \\
\bottomrule
\end{tabular}
\end{table}

\begin{table}[t]
\centering
\caption{Evaluation hyperparameters used in our experiments.}
\label{tab:eval_hyper}
\begin{tabular}{lc}
\toprule
\textbf{Hyperparameter} & \textbf{Value} \\
\midrule
Temperature              & 1.0  \\
Top-$p$                  & 0.7  \\
Max prompt length        & 36000 \\
Max response length      & 12000 \\
Max agent turns          & 10   \\
Validation sample size   & 500   \\
\bottomrule
\end{tabular}
\end{table}

\begin{figure*}[t]
\centering
\includegraphics[width=0.24\linewidth]{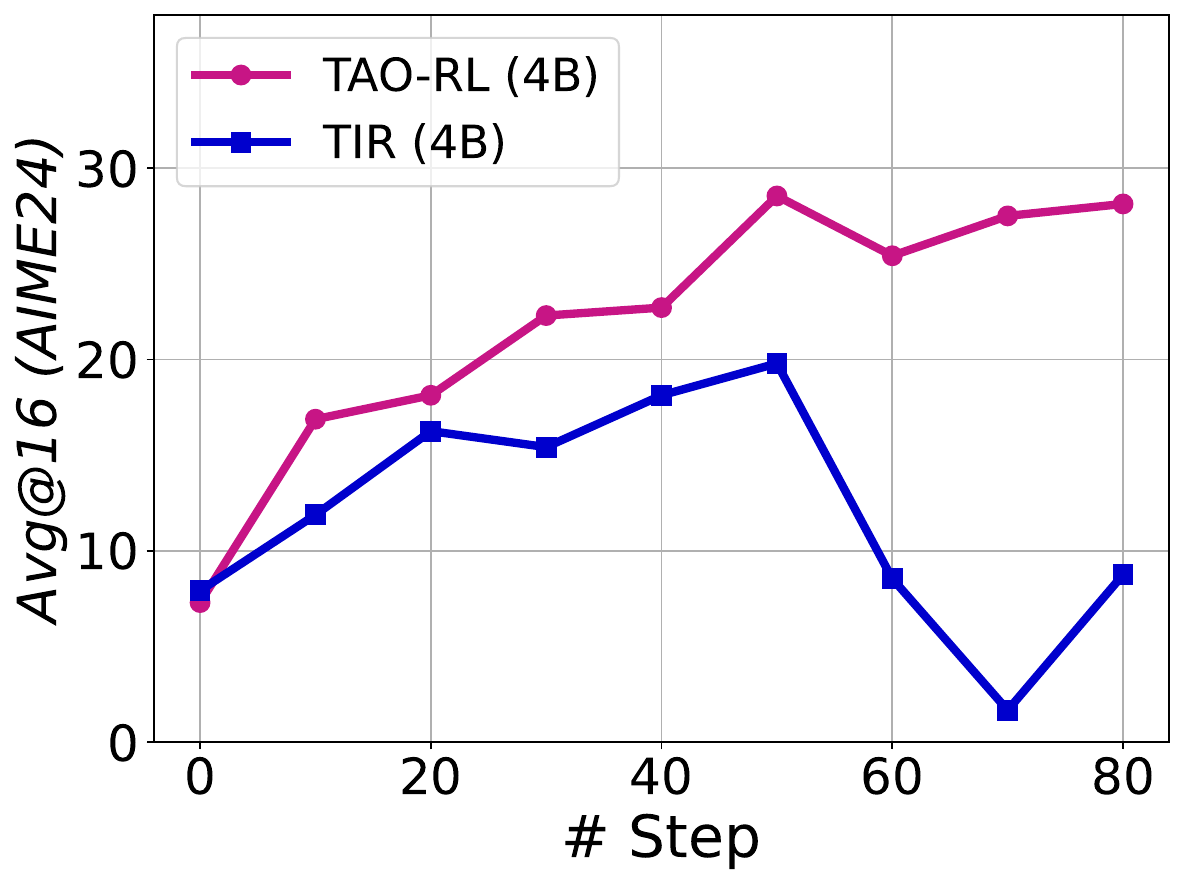}
\includegraphics[width=0.24\linewidth]{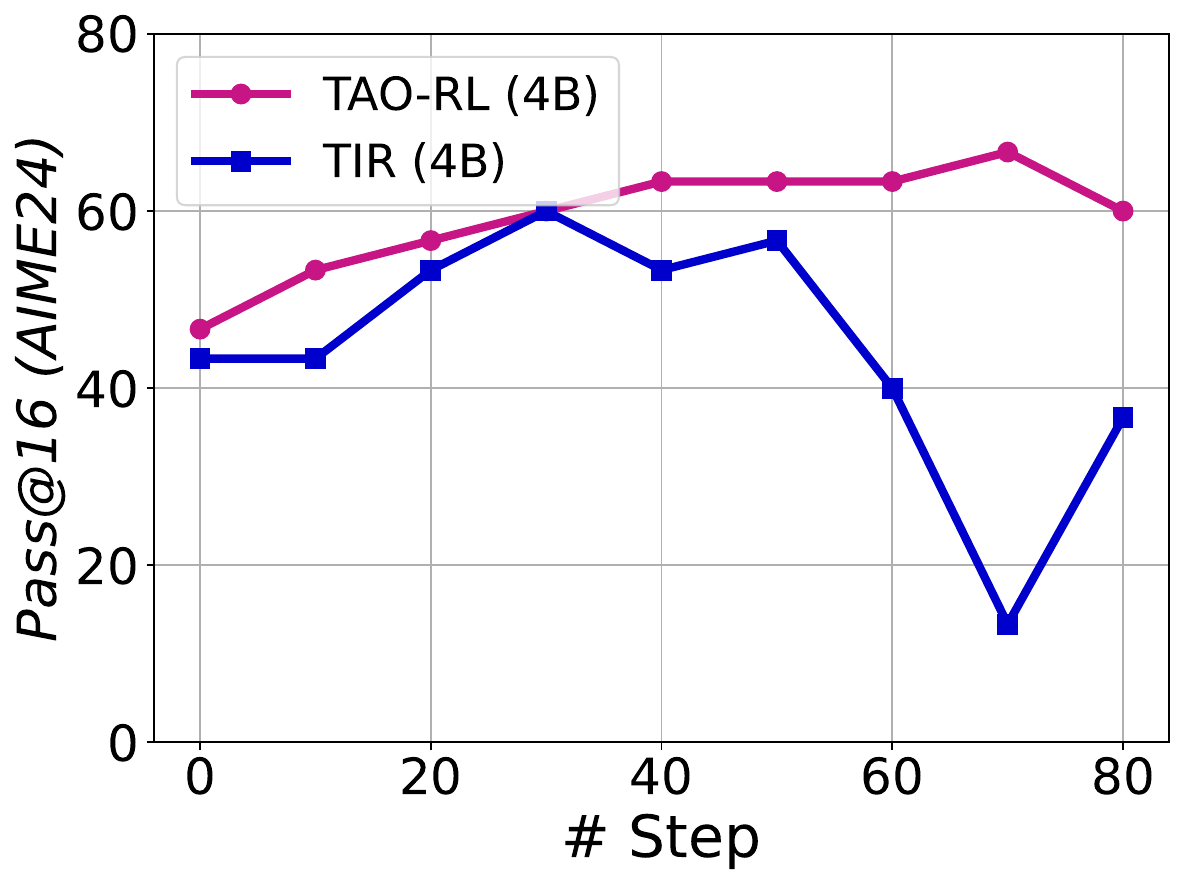}
\includegraphics[width=0.24\linewidth]{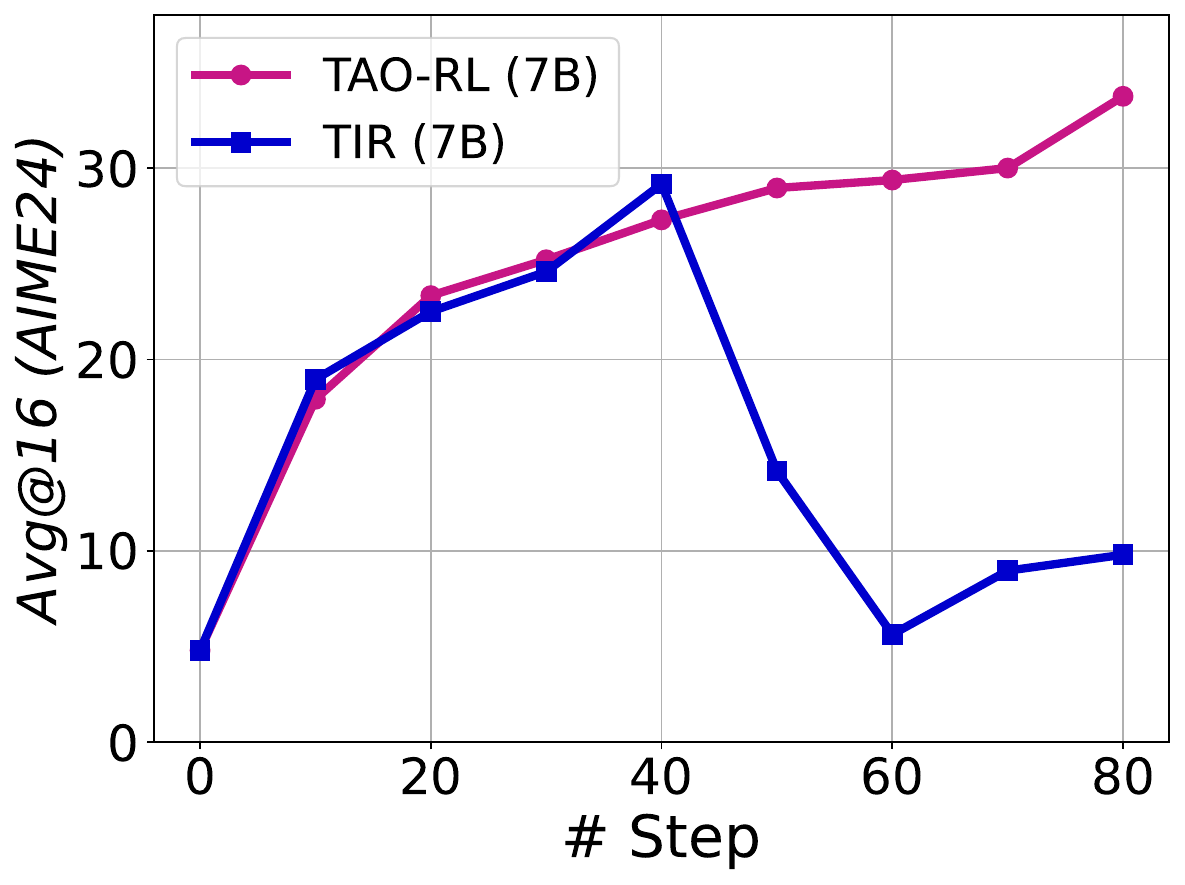}
\includegraphics[width=0.24\linewidth]{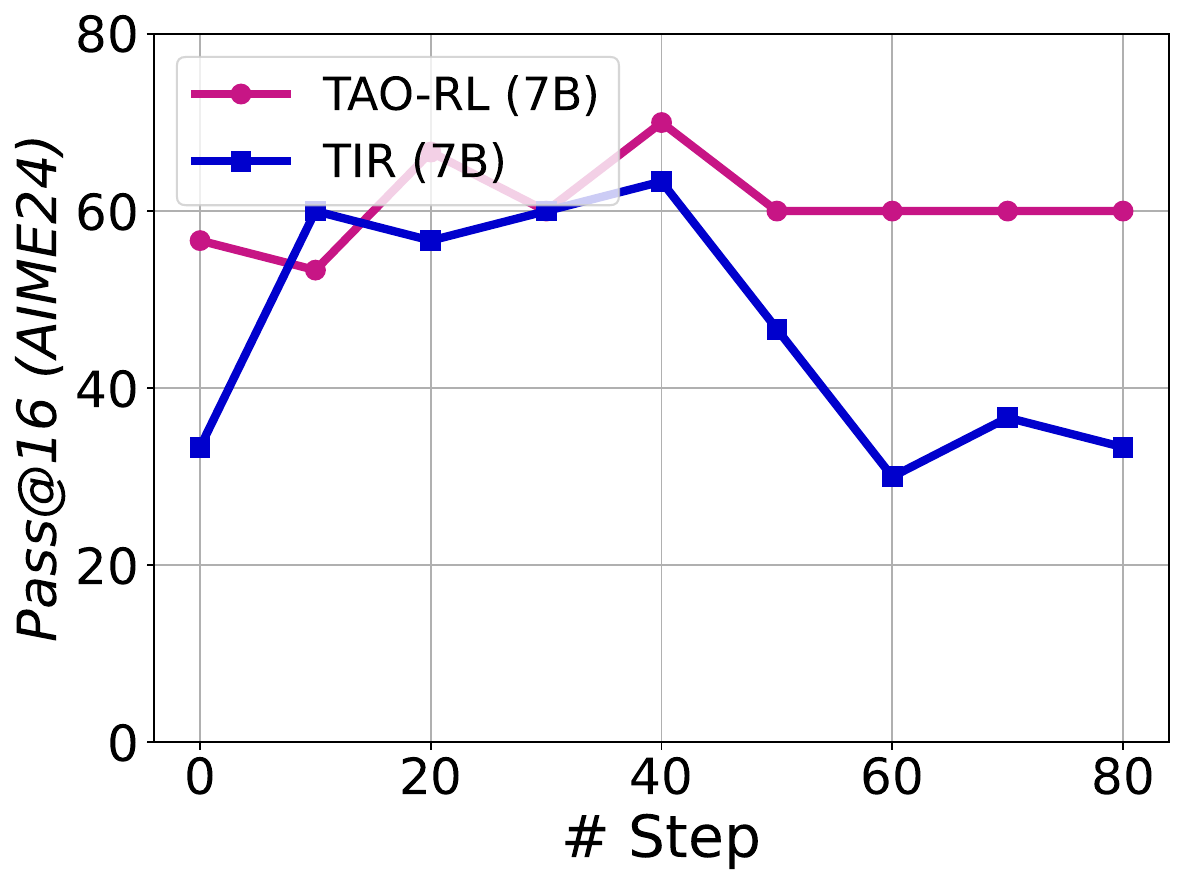}
\includegraphics[width=0.24\linewidth]{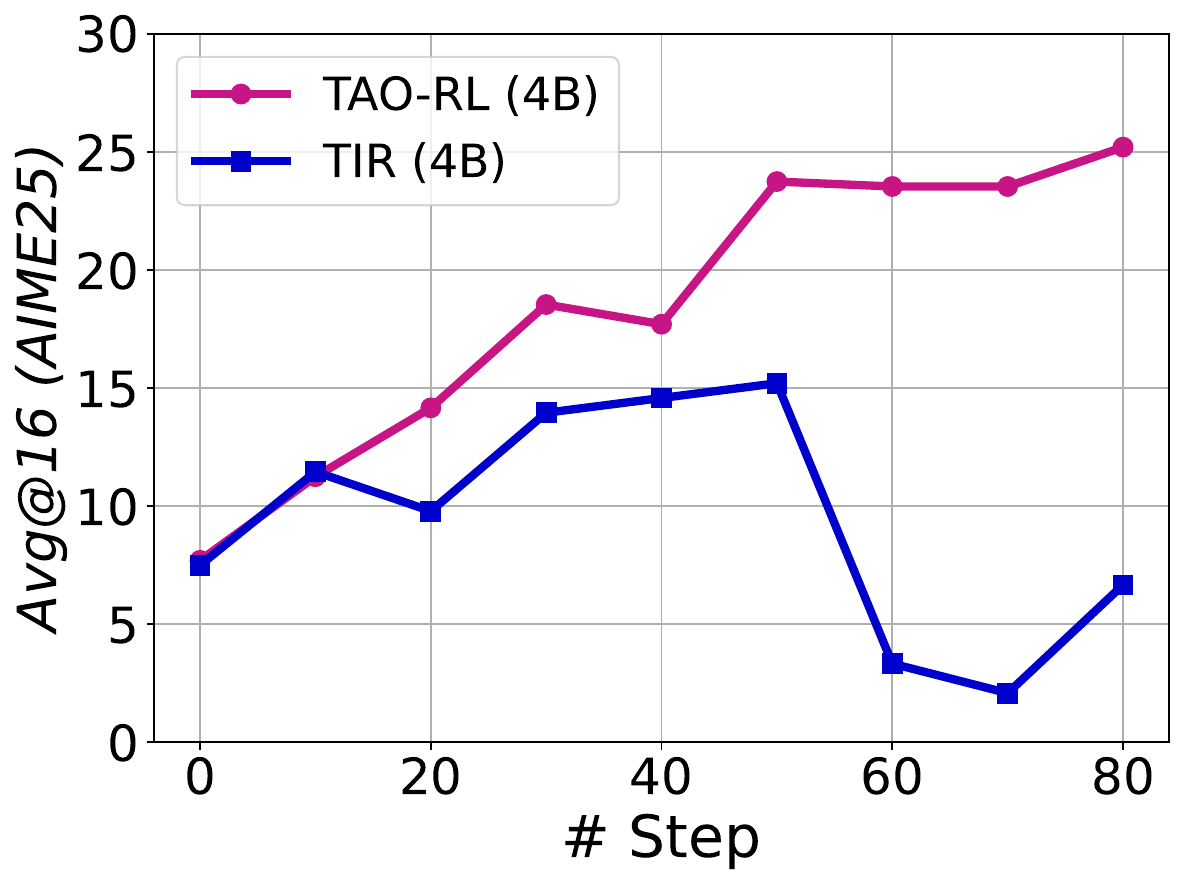}
\includegraphics[width=0.24\linewidth]{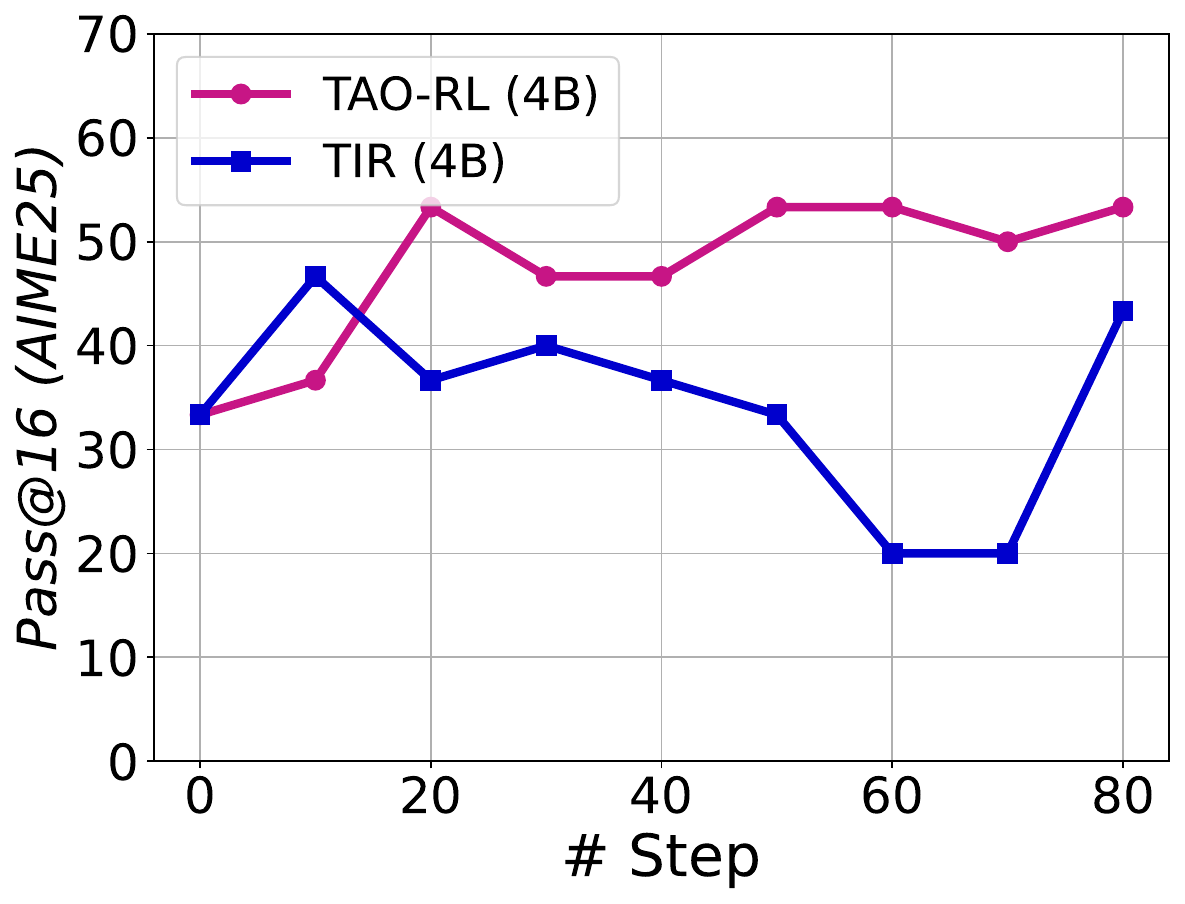}
\includegraphics[width=0.24\linewidth]{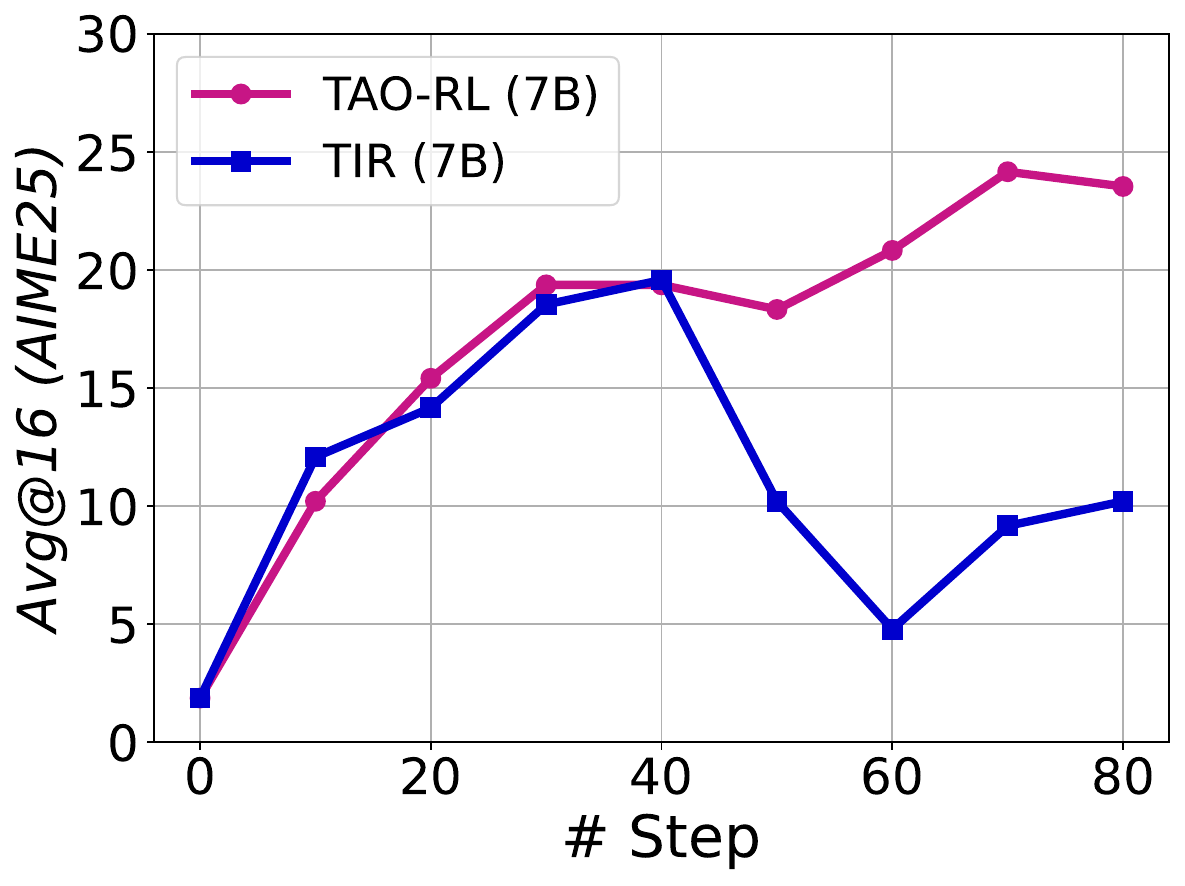}
\includegraphics[width=0.24\linewidth]{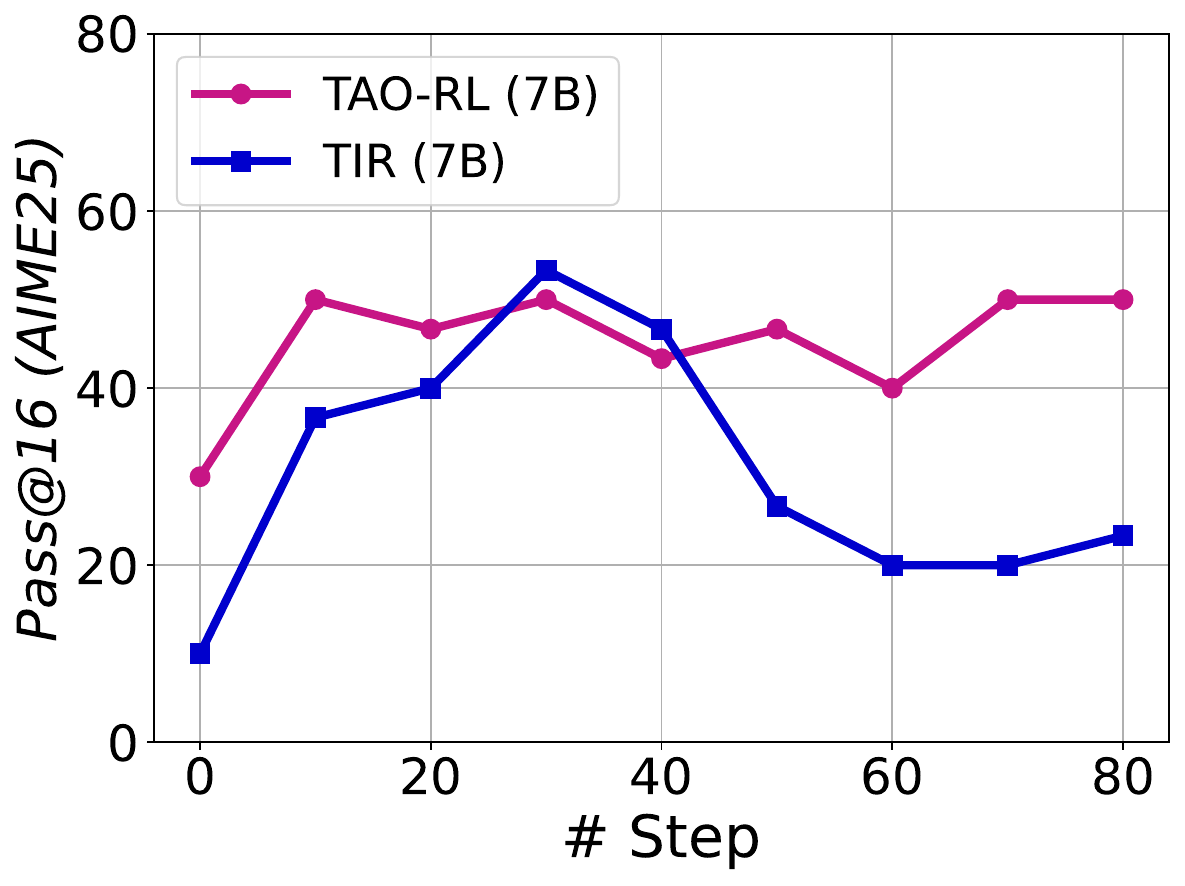}

\includegraphics[width=0.24\linewidth]{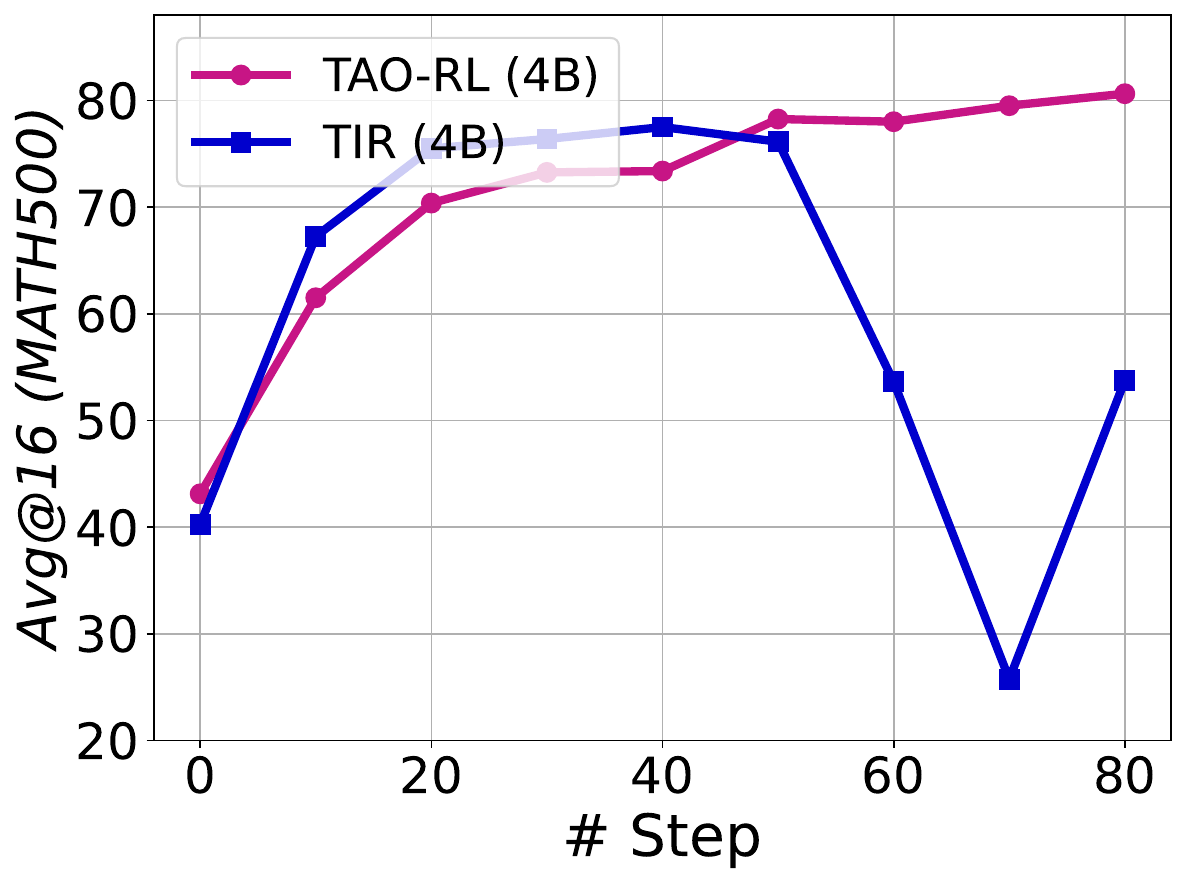}
\includegraphics[width=0.24\linewidth]{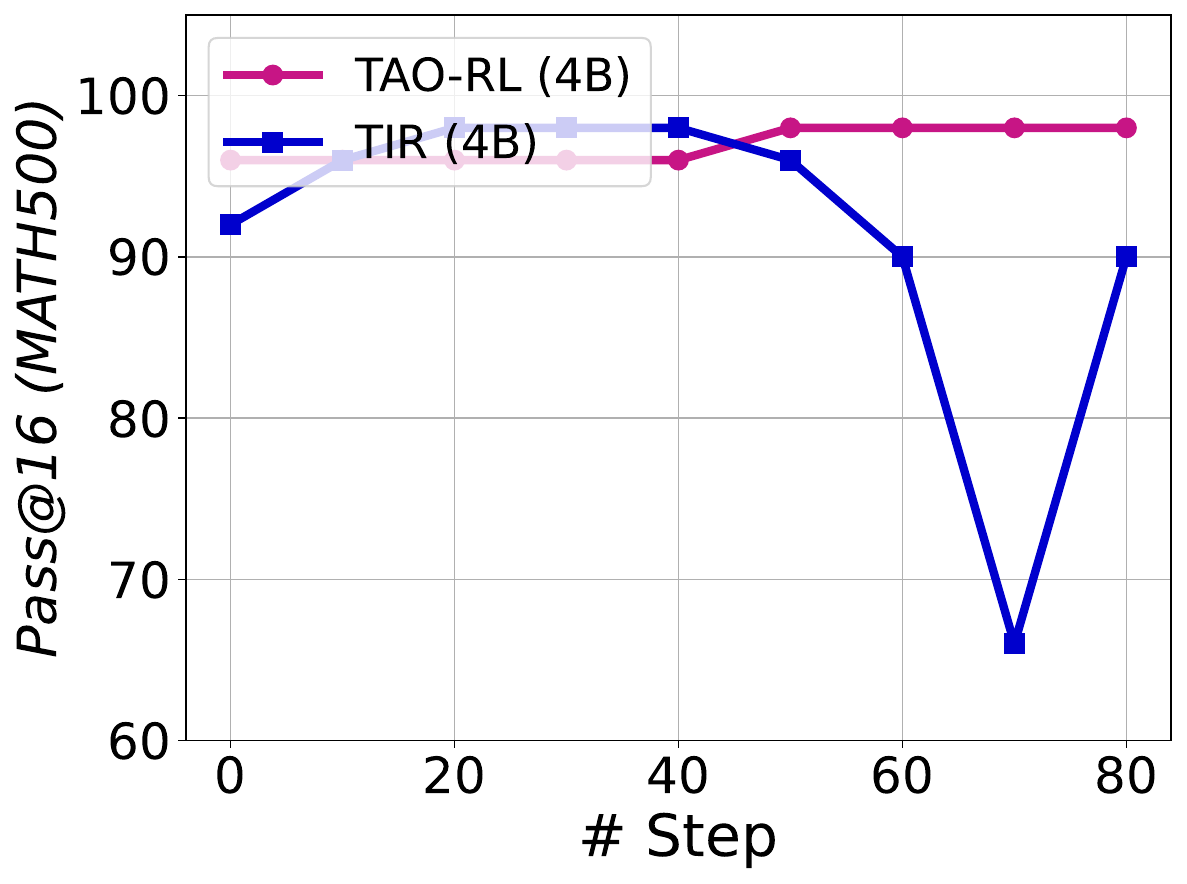}
\includegraphics[width=0.24\linewidth]{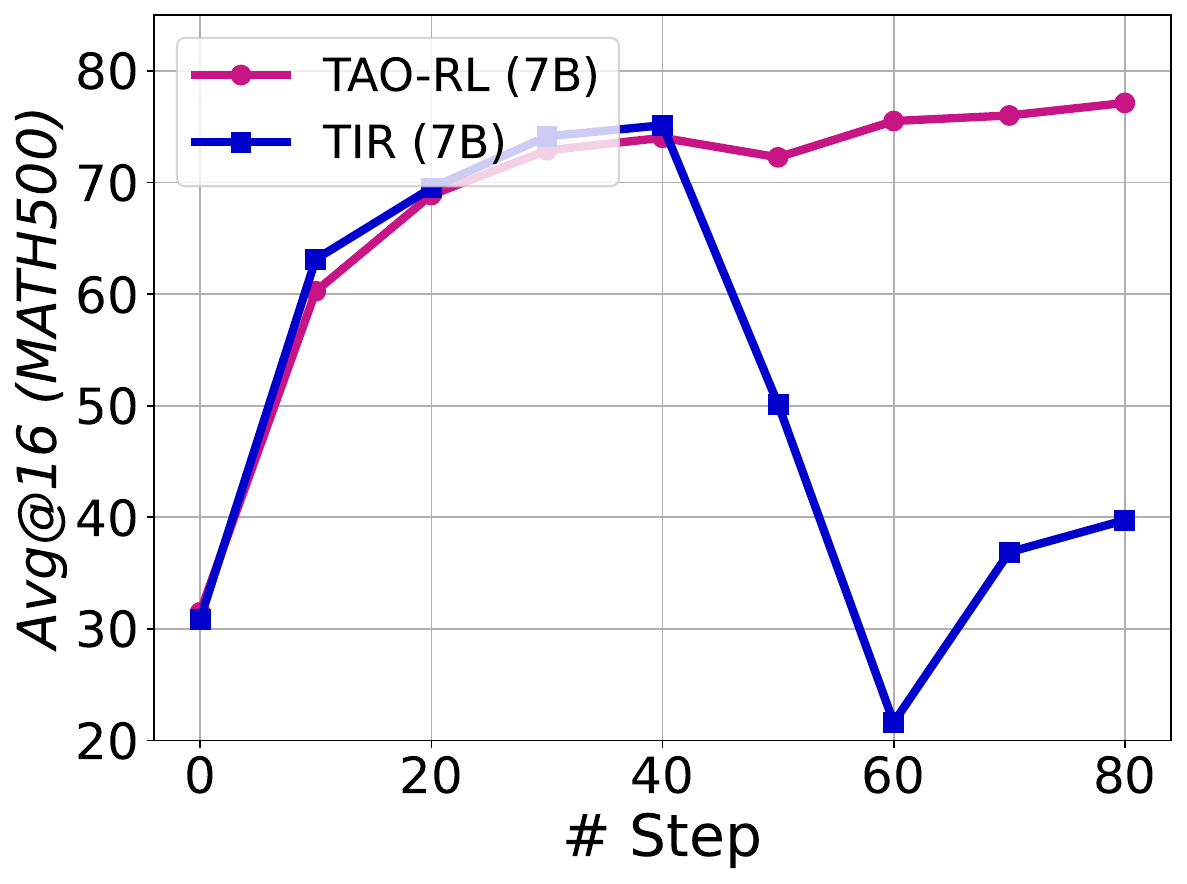}
\includegraphics[width=0.24\linewidth]{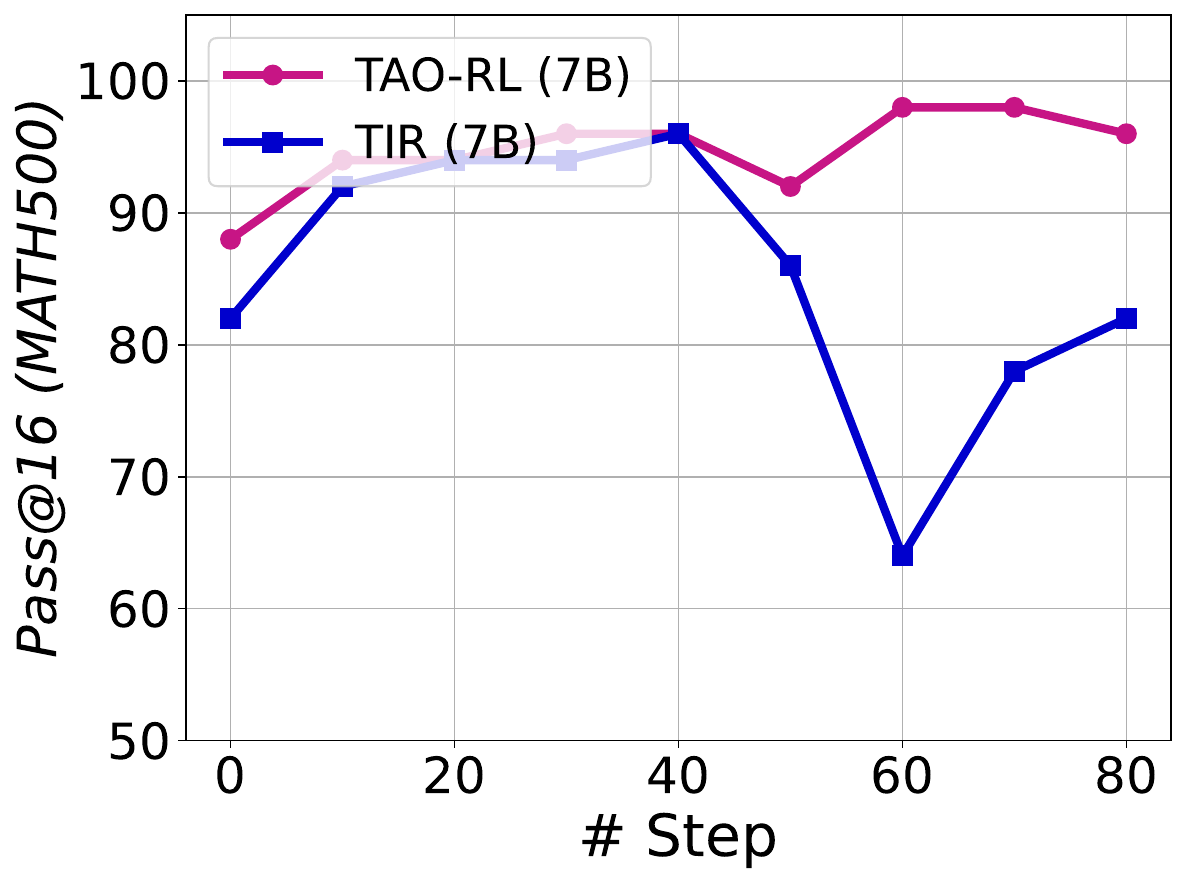}

\includegraphics[width=0.24\linewidth]{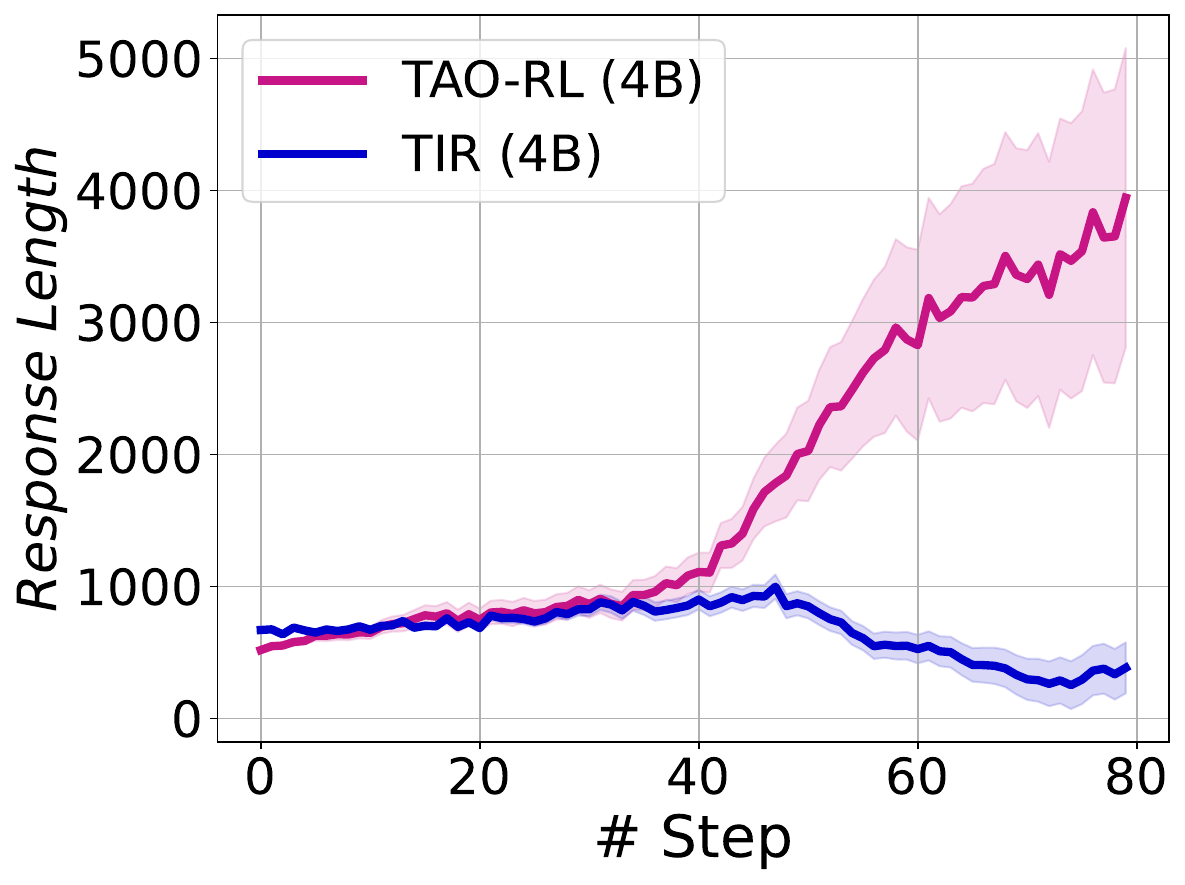}
\includegraphics[width=0.24\linewidth]{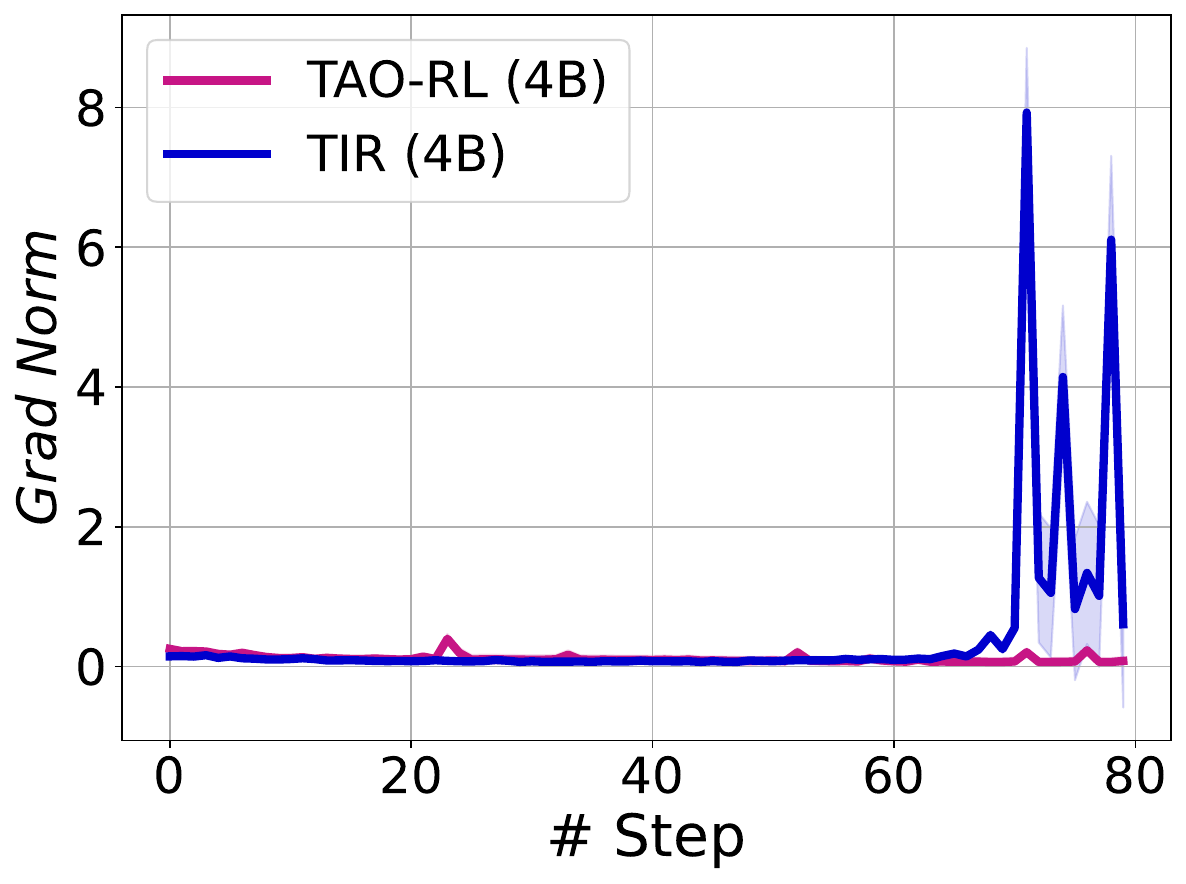}
\includegraphics[width=0.24\linewidth]{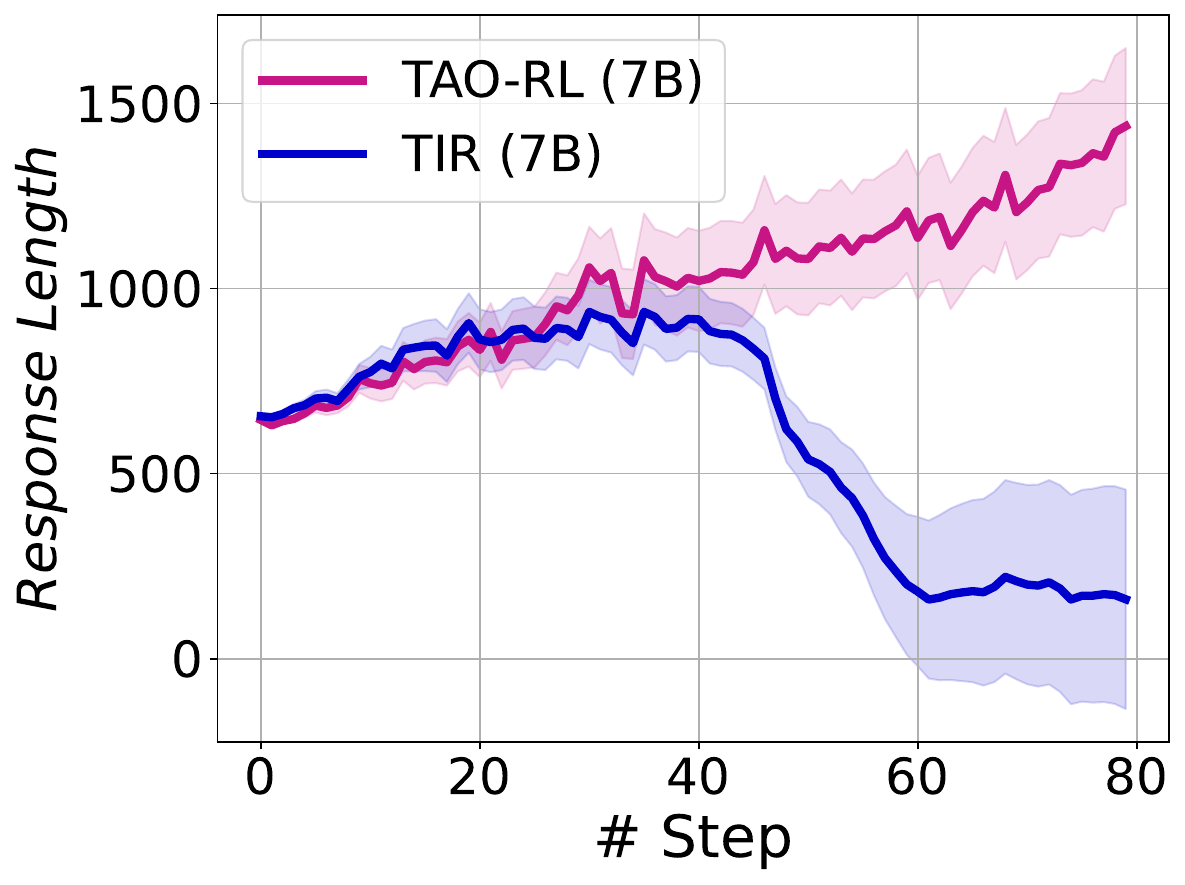}
\includegraphics[width=0.24\linewidth]{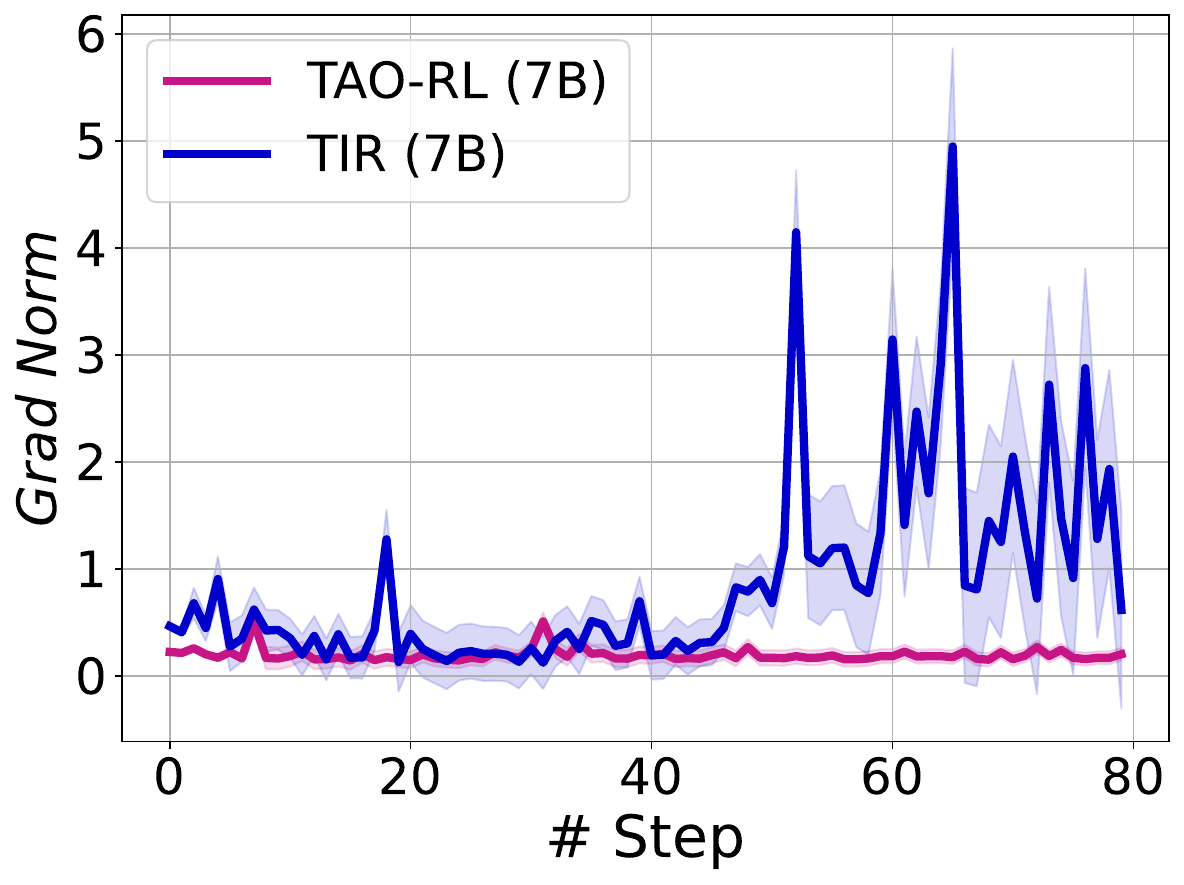}

\includegraphics[width=0.24\linewidth]{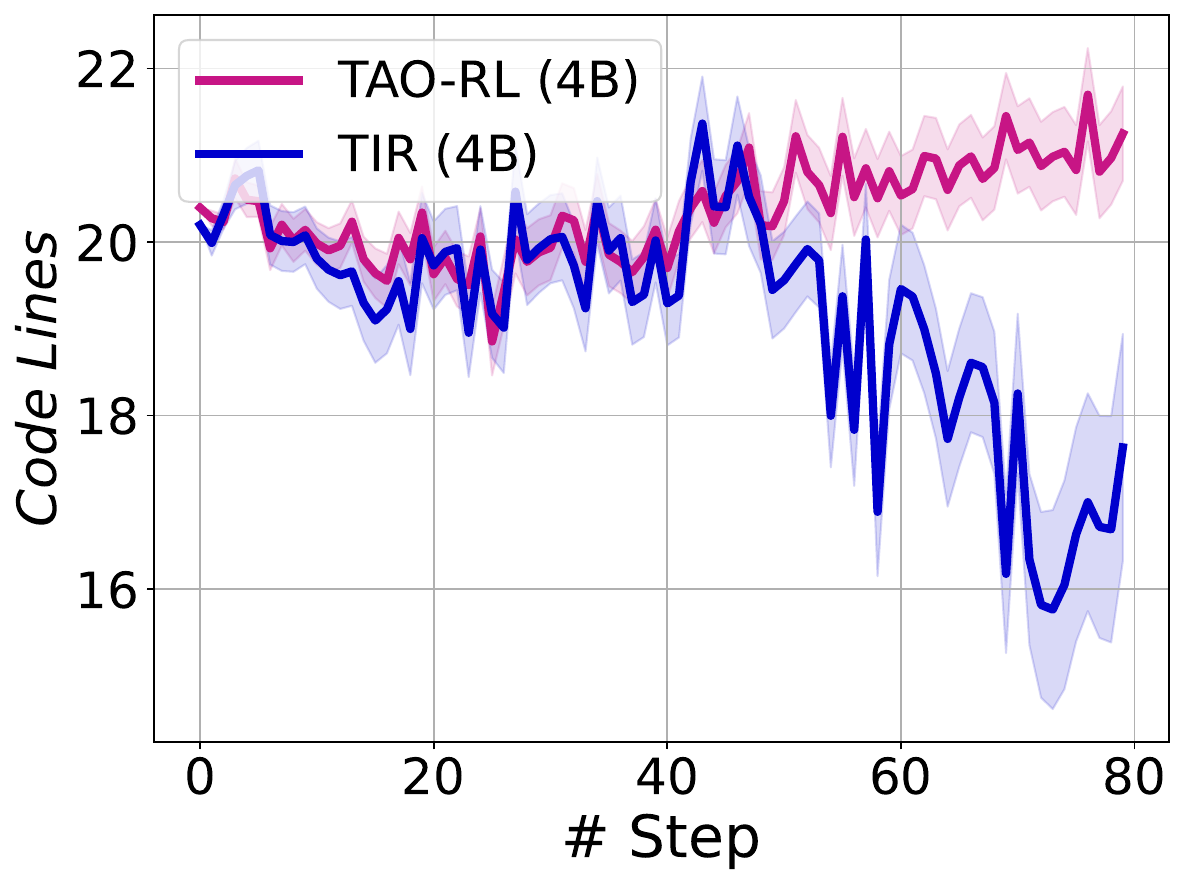}
\includegraphics[width=0.24\linewidth]{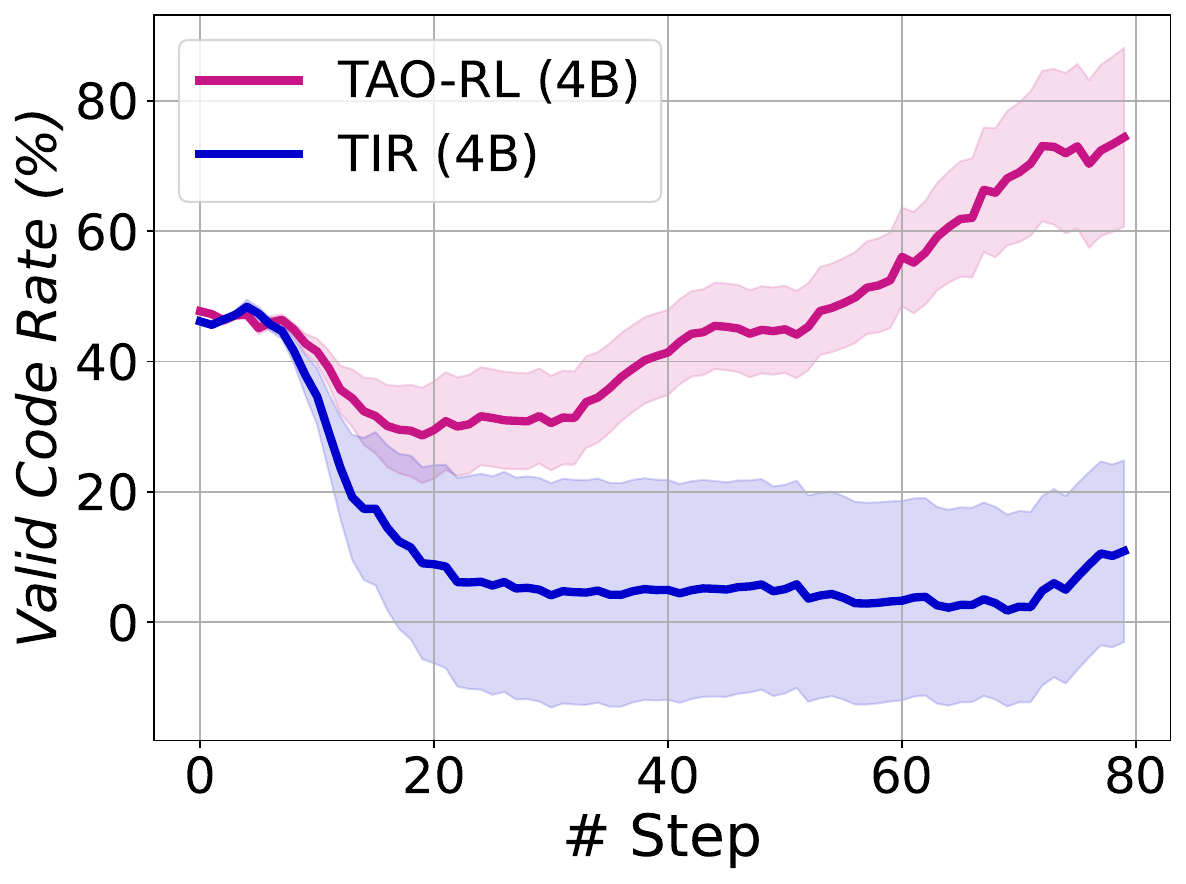}
\includegraphics[width=0.24\linewidth]{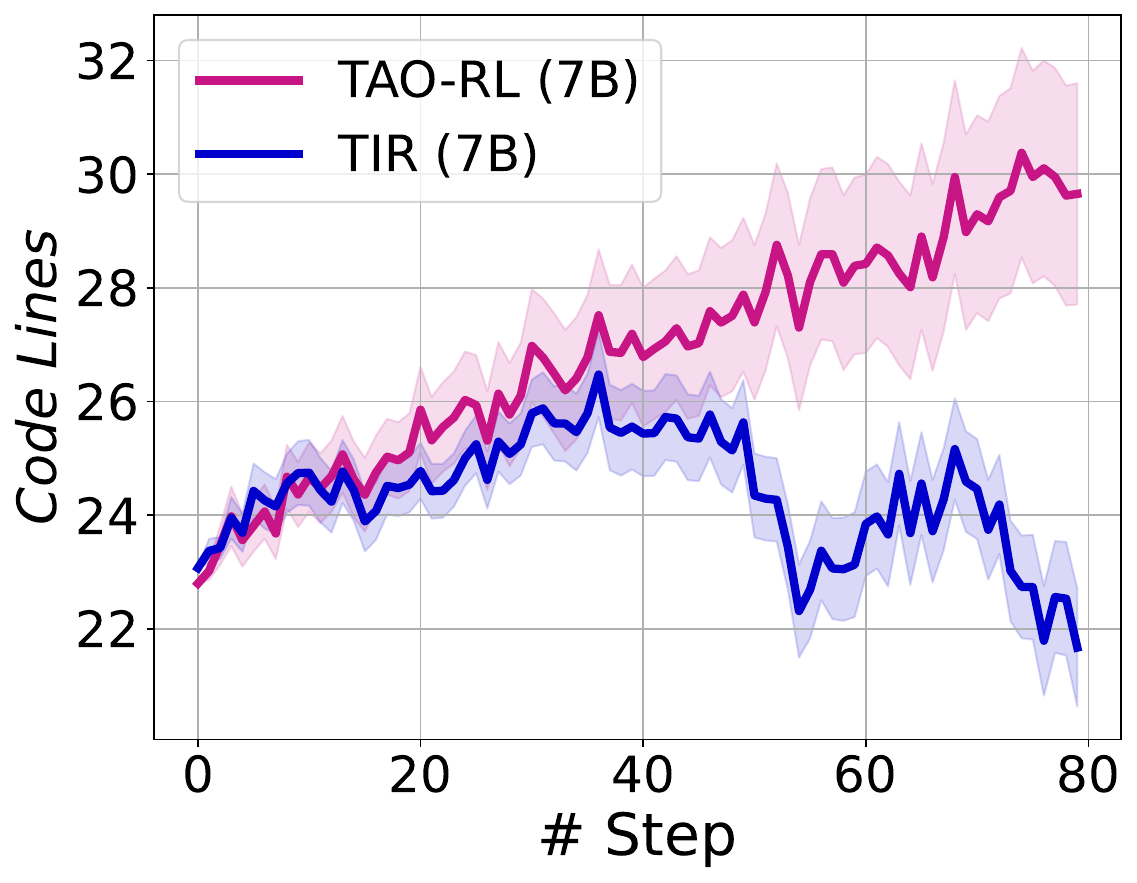}
\includegraphics[width=0.24\linewidth]{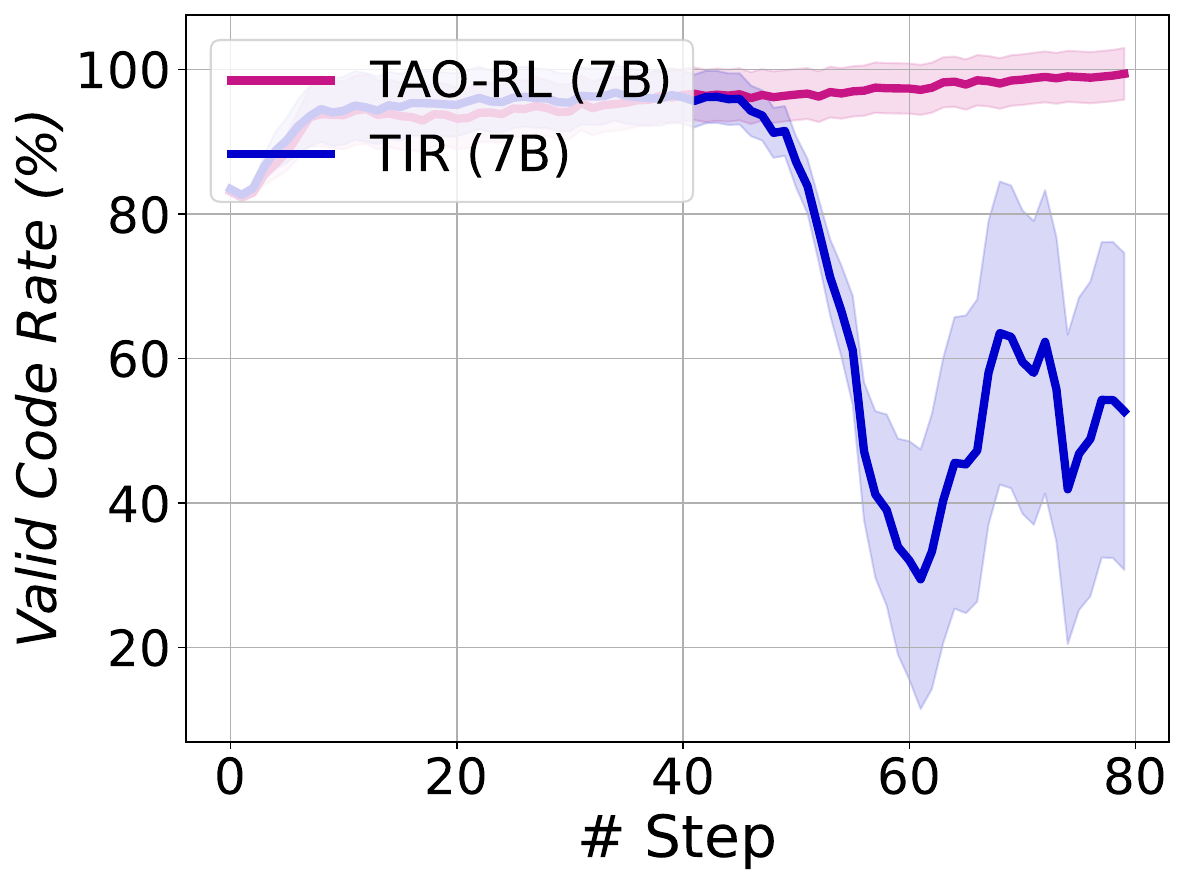}
\caption{Comprehensive training dynamics comparison between \method and TIR under Qwen3-4B-Base and Qwen2.5-7B base models across reasoning performance (\textit{Avg@16} and \textit{Pass@16} on AIME24, AIME25, and MATH500) and agentic behavior metrics (response length, gradient norm, code lines, and valid code rate) throughout training.}
\label{fig:tir}
\end{figure*}

\subsection{More Results}
\label{sec:app more results}

\begin{figure*}[t]
\centering
\includegraphics[width=0.24\linewidth]{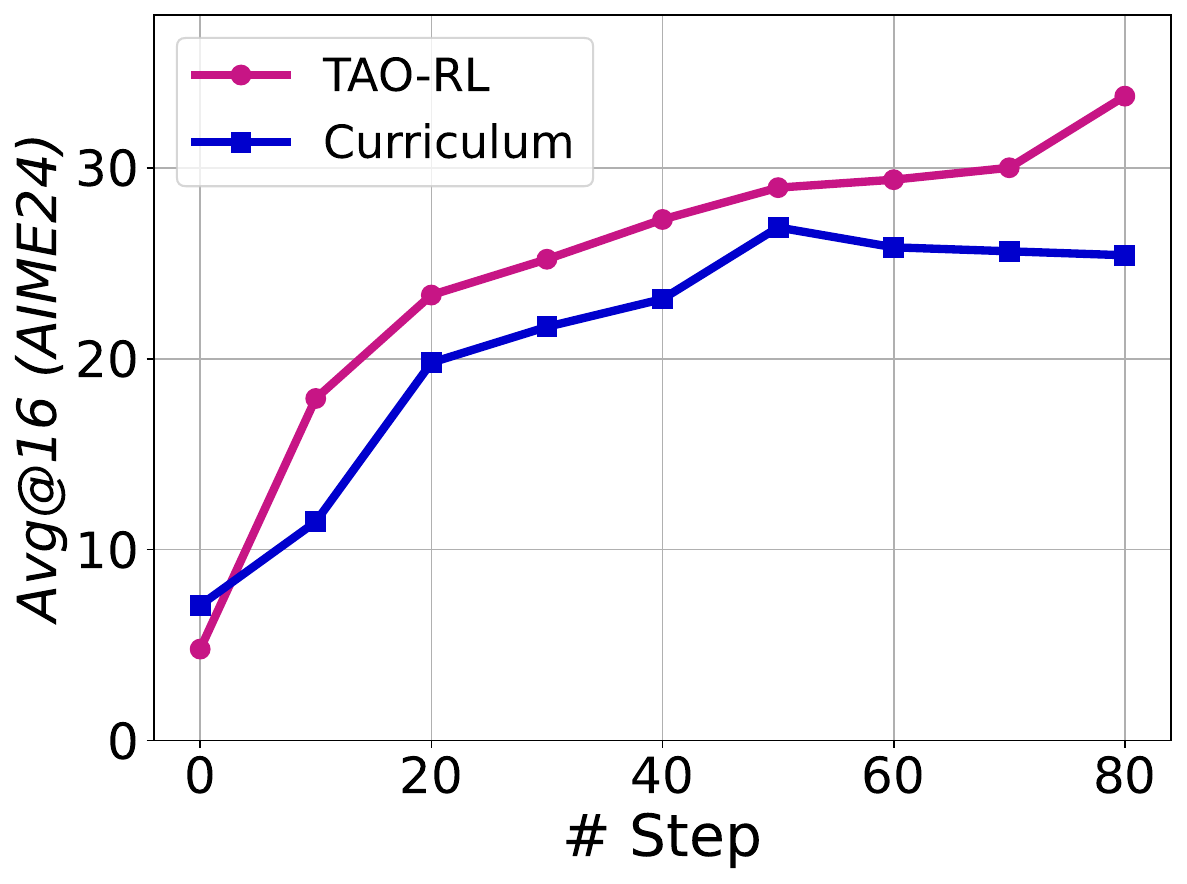}
\includegraphics[width=0.24\linewidth]{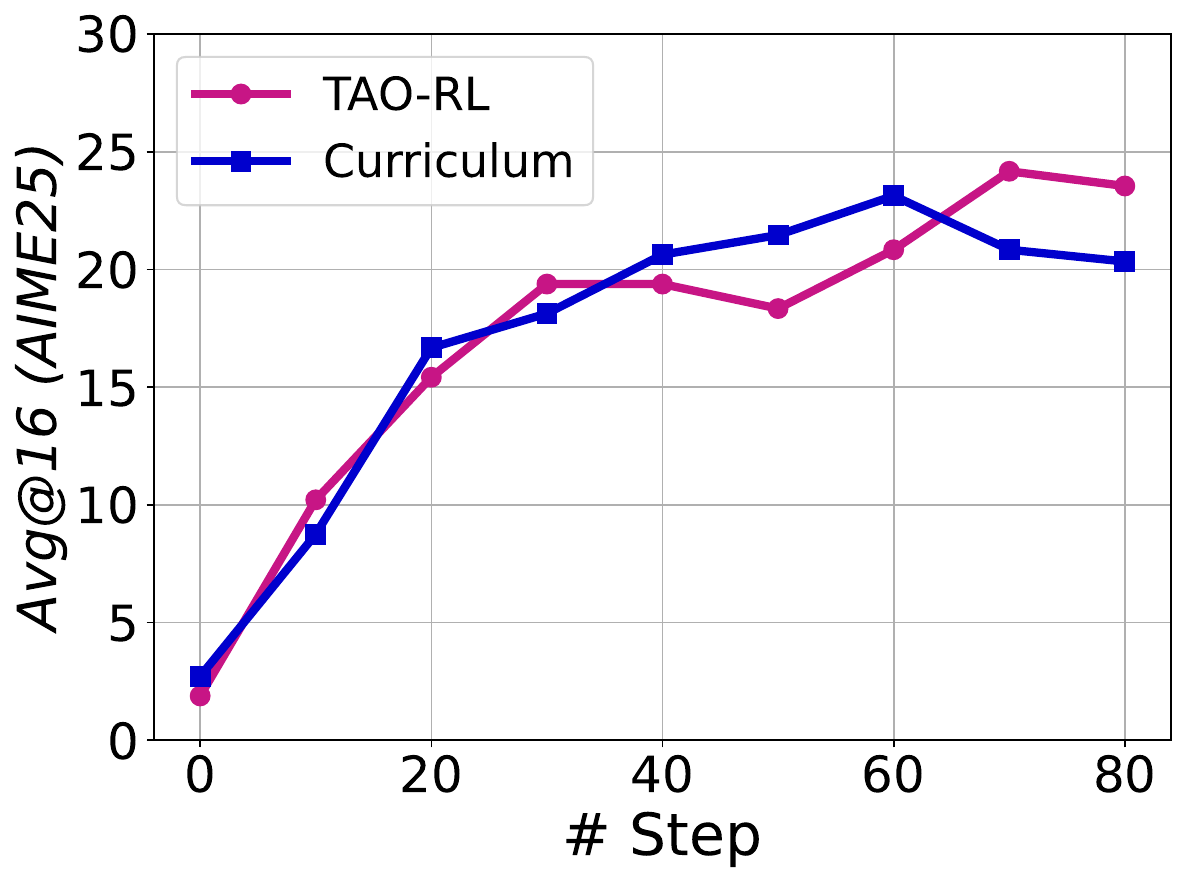}
\includegraphics[width=0.24\linewidth]{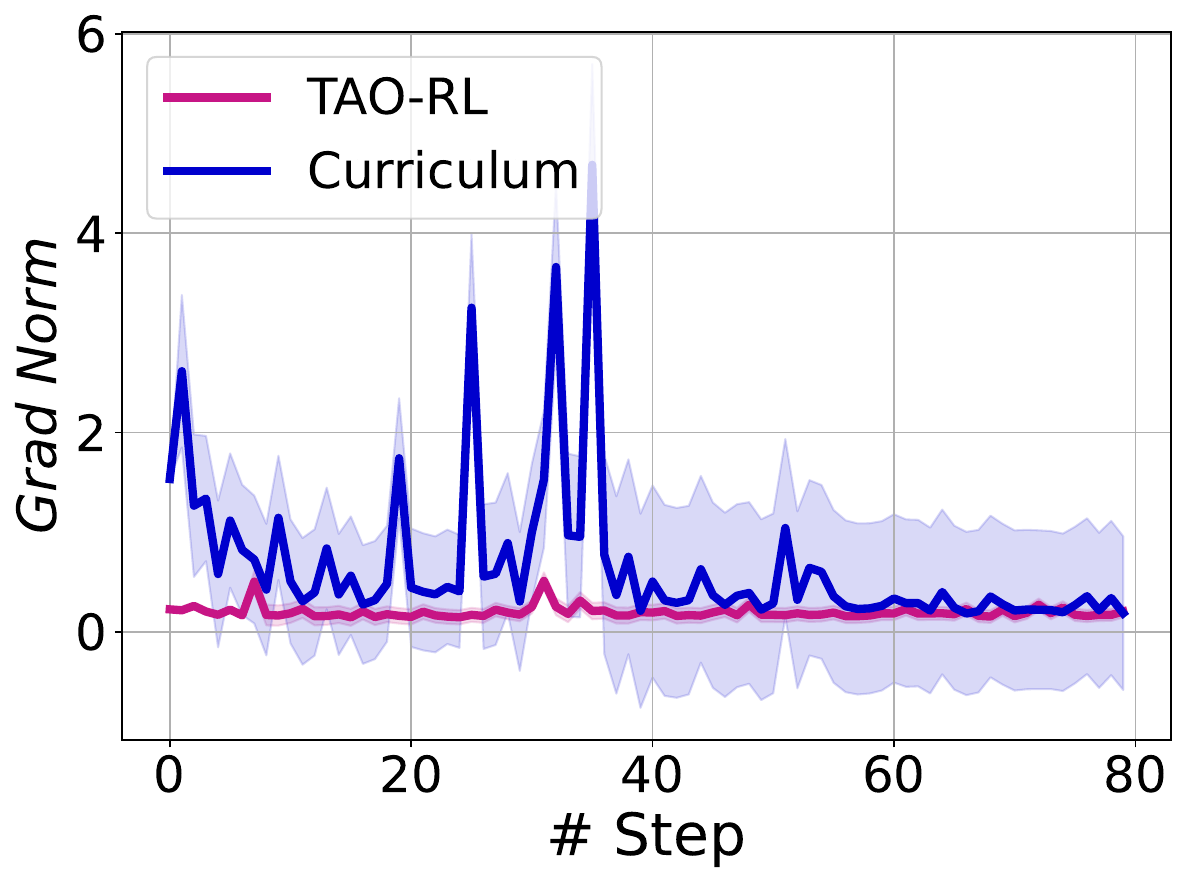}
\includegraphics[width=0.24\linewidth]{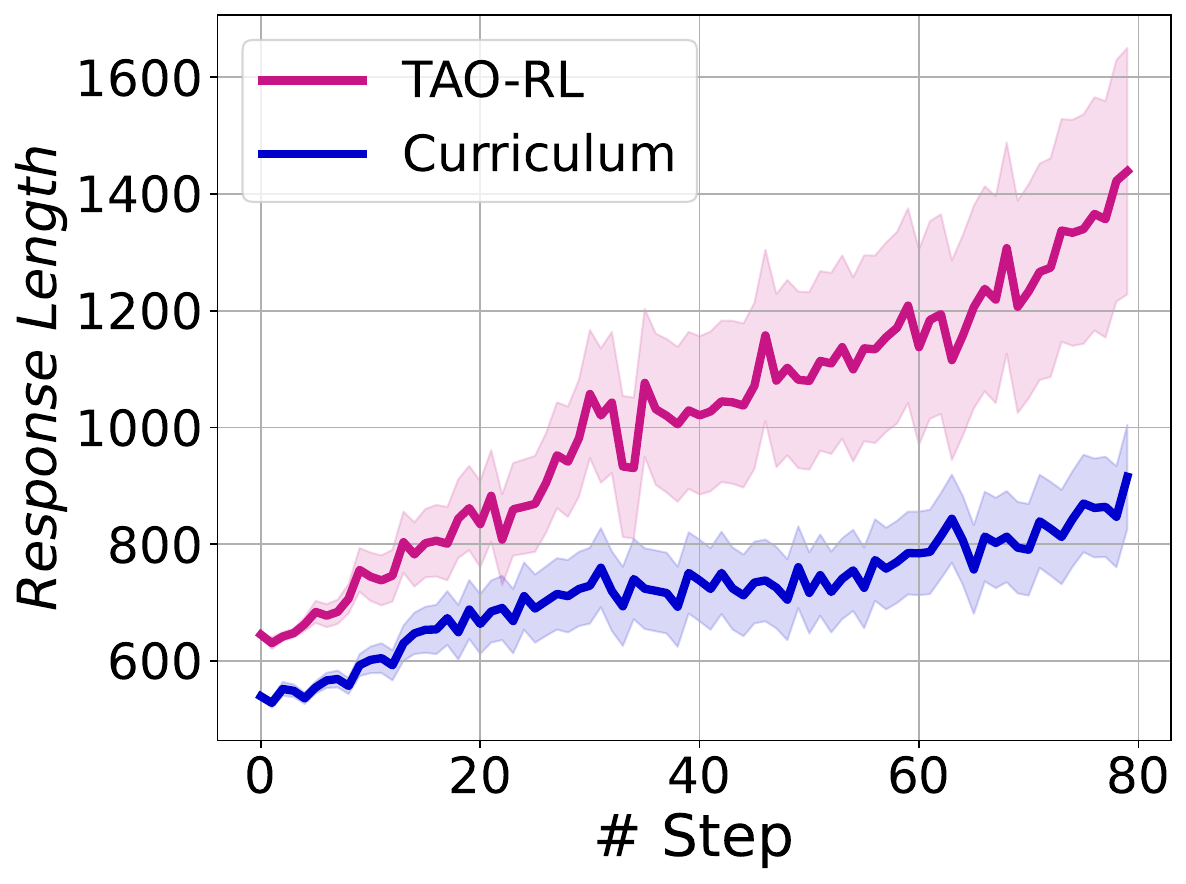}
\caption{Asymptotic performance of \textit{Avg@16}, grad norm and response length of \method compared with Curriculum learning setting under Qwen2.5-$7$B base model during training.}
\label{fig:curr}
\end{figure*}

\subsubsection{Comparison with TIR}

To provide a comprehensive view of how \method improves upon naive tool integration, we present a detailed comparison against TIR across multiple dimensions throughout training, as shown in Fig.~\ref{fig:tir}.

\textbf{Reasoning Performance.} On all three benchmarks of AIME24, AIME25, and MATH500, \method consistently outperforms TIR in both Avg@16 and Pass@16 throughout training. While TIR exhibits erratic fluctuations and fails to maintain a stable improvement trajectory, \method achieves steady and monotonic performance gains, ultimately reaching substantially higher accuracy on both average and peak metrics. The gap is particularly pronounced on the harder AIME benchmarks, where TIR's instability leads to severe performance degradation in later training stages.

\textbf{Training Dynamics.} The training curves reveal a stark contrast in optimization behavior. TIR exhibits large and irregular gradient norm fluctuations throughout training, reflecting the destabilizing effect of unfiltered tool interactions on policy updates. In contrast, \method maintains consistently lower and more stable gradient norms, confirming that tool-aware trajectory filtering effectively suppresses noisy gradient signals from degenerate rollouts.

\textbf{Tool-Use Quality.} The response length and code lines curves further highlight the depth of tool engagement cultivated by \method. While TIR produces shorter responses with fewer lines of code, \method generates substantially longer and more elaborate reasoning trajectories throughout training, reflecting deeper and more structured tool-use behavior. Most notably, the valid code rate of TIR remains markedly lower and more volatile compared to \method, which maintains a near-ceiling valid code rate throughout training. This confirms that without filtering tool-execution-failed trajectories, TIR consistently produces malformed tool interactions that contaminate the training distribution.

Taken together, these results demonstrate that simply incorporating tools into the training loop without principled data filtering and targeted exploration incentives leads to unstable optimization, shallow tool engagement, and degraded reasoning performance, all of which \method systematically addresses through its two complementary components.

\subsubsection{Comparison of Curriculum Learning}
\label{sec:app_curr}

To further investigate whether preserving data quantity through structured data organization can serve as a viable alternative to our quality-driven filtering strategy, we compare \method against a curriculum learning variant that organizes the full training data in a progressive easy-to-hard order based on task difficulty. Unlike our filtering approach, this variant retains all trajectories, including degenerate ones, aiming to stabilize training through difficulty-aware data ordering rather than explicit data removal.

As shown in Fig.~\ref{fig:curr}, the curriculum-based approach does yield a moderate stabilization effect compared to vanilla TIR: both gradient norm fluctuations and response length variability are somewhat reduced, confirming that progressive data organization provides a degree of training stability. However, the improvement is limited precisely because degenerate trajectories, those with failed tool executions or uninformative advantage estimates, remain in the training distribution and continue to inject noisy gradient signals throughout training.

In contrast, \method achieves consistently higher Avg@16 on both AIME24 and AIME25, along with substantially lower and more stable gradient norms and greater response lengths throughout training. These results demonstrate that explicitly removing degenerate rollouts is fundamentally more effective than reorganizing them, as the root cause of instability lies in the quality of individual trajectories rather than their ordering. Structured data organization can partially mitigate the symptoms of tool-induced instability, but only principled trajectory filtering can address the underlying noise at its source.

\begin{table*}[t]
 \caption{Ablation study on retaining all-wrong rollout groups under the Qwen2.5-7B base model. Both TAO-RL configurations use $\alpha=0.01$ and $q_{\mathcal{H}}=20\%$. $\Delta$ denotes the performance difference between the default strategy and the Keep-Wrong variant.}
 \vspace{-2mm}
\renewcommand{\arraystretch}{1.3}
\setlength{\tabcolsep}{3pt}
    \centering 
\begin{tabular}{lcccccccccccc}
\toprule
\textbf{} 
& \multicolumn{2}{c}{\textbf{AIME24}} 
& \multicolumn{2}{c}{\textbf{AIME25}} 
& \multicolumn{2}{c}{\textbf{AMC23}} 
& \multicolumn{2}{c}{\textbf{MATH500}} 
& \multicolumn{2}{c}{\textbf{OlympiadBench}} 
& \multicolumn{2}{c}{\textbf{Hmmt25}}
\\ 
\cmidrule(lr){2-3} 
\cmidrule(lr){4-5} 
\cmidrule(lr){6-7} 
\cmidrule(lr){8-9} 
\cmidrule(lr){10-11} 
\cmidrule(lr){12-13}
\textbf{Method} 
& \textit{Avg@16} & \textit{Pass@16} 
& \textit{Avg@16} & \textit{Pass@16} 
& \textit{Avg@16} & \textit{Pass@16} 
& \textit{Avg@16} & \textit{Pass@16} 
& \textit{Avg@16} & \textit{Pass@16} 
& \textit{Avg@16} & \textit{Pass@16}  \\ 
\midrule

\textbf{\method} (\textit{Keep-Wrong}) 
& 30.00 & 60.00 
& 22.92 & 46.67 
& 65.63 & 90.00 
& 76.73 & 95.00 
& 44.25 & 68.00 
& 15.83 & 43.33 \\

\rowcolor{blue!10} 
\textbf{\method} (\textit{Default})
& \textbf{35.42} & \textbf{73.33} 
& \textbf{23.96} & \textbf{46.67} 
& \textbf{69.06} & \textbf{97.50} 
& \textbf{78.80} & \textbf{95.60} 
& \textbf{46.99} & \textbf{70.00} 
& \textbf{18.33} & \textbf{46.67} \\

\quad $\Delta$ 
& \color{darkgreen}{+5.42} & \color{darkgreen}{+13.33} 
& \color{darkgreen}{+1.04} & 0.00 
& \color{darkgreen}{+3.43} & \color{darkgreen}{+7.50} 
& \color{darkgreen}{+2.07} & \color{darkgreen}{+0.60} 
& \color{darkgreen}{+2.74} & \color{darkgreen}{+2.00} 
& \color{darkgreen}{+2.50} & \color{darkgreen}{+3.34} \\
\bottomrule 
\end{tabular}
\label{tab:app_keep_wrong}
\end{table*}

\begin{table*}[t]
 \caption{Hyperparameter analysis of $\alpha$ on six benchmarks of \textit{Pass@16} and \textit{Avg@16} under Qwen2.5-7B base model. We bold the best results, underline the sub-optimal results, and highlight our selection.}
\renewcommand{\arraystretch}{1.2}
\setlength{\tabcolsep}{5pt} 
    \centering 
\begin{tabular}{lcccccccccccc}
\toprule
\textbf{}                          & \multicolumn{2}{c}{\textbf{AIME24}} & \multicolumn{2}{c}{\textbf{AIME25}} & \multicolumn{2}{c}{\textbf{AMC23}} & \multicolumn{2}{c}{\textbf{MATH500}} & \multicolumn{2}{c}{\textbf{OlympiadBench}} & \multicolumn{2}{c}{\textbf{Hmmt25}} 
\\ \cmidrule(lr){2-3} \cmidrule(lr){4-5} \cmidrule(lr){6-7} \cmidrule(lr){8-9} \cmidrule(lr){10-11} \cmidrule(lr){12-13}
\textbf{Method}      & \textit{Avg@16}               & \textit{Pass@16}  & \textit{Avg@16} & \textit{Pass@16}  & \textit{Avg@16} & \textit{Pass@16} & \textit{Avg@16} & \textit{Pass@16}  & \textit{Avg@16}  & \textit{Pass@16}     & \textit{Avg@16}     & \textit{Pass@16}  \\ \bottomrule
 \textbf{Base model}
 &   5.00    &     36.67
 &   3.96    &     26.67  
 &   21.56   &     80.00    
 &   31.99   &     88.00  
 &   13.80   &     55.00        
 &   6.67    &     26.67        \\
\textbf{TIR}
&    10.63   &     40.00 
&    10.42   &     20.00       
&    28.91   &     70.00   
&    42.11   &     79.80           
&    20.88   &     49.80 
&    11.04   &     23.33        \\
\textbf{$\alpha=0.1$} 
&    24.79    &    60.00  
&    14.38    &    \underline{40.00}   
&    53.44    &    85.00
&    68.46    &    93.00 
&    36.38    &    68.80     
&    13.33    &    33.33          \\
\rowcolor{blue!10} \textbf{$\alpha=0.01$} 
&    \textbf{35.42}    &    \textbf{73.33}   
&    \textbf{23.96}    &    \textbf{46.67}   
&    \underline{69.06}    &    \textbf{97.50}
&    \textbf{78.80}    &    \textbf{95.60}
&    \textbf{46.99}    &    \textbf{70.00}     
&    \underline{18.33}    &    \textbf{46.67}          \\
\textbf{$\alpha=0.001$} 
&    \underline{31.88}    &    \underline{63.33}   
&    \underline{22.08}    &    \textbf{46.67}   
&    \textbf{69.22}    &    \underline{95.00} 
&    \underline{77.79}    &    \underline{95.40} 
&    \underline{45.93}    &    \underline{69.60}   
&    \textbf{19.38}    &    \underline{40.00}          \\
\bottomrule 
\end{tabular}
\label{tab:app_alpha_avg_pass}
\end{table*}

\begin{table*}[t]
 \caption{Hyperparameter analysis of $\alpha$ on six benchmarks of \textit{Len@16} and \textit{Code\_Line} under Qwen2.5-7B base model. We bold the best results, underline the sub-optimal results, and highlight our selection.}
\renewcommand{\arraystretch}{1.2}
\setlength{\tabcolsep}{4pt} 
    \centering 
\begin{tabular}{lcccccccccccc}
\toprule
\textbf{}                          & \multicolumn{2}{c}{\textbf{AIME24}} & \multicolumn{2}{c}{\textbf{AIME25}} & \multicolumn{2}{c}{\textbf{AMC23}} & \multicolumn{2}{c}{\textbf{MATH500}} & \multicolumn{2}{c}{\textbf{OlympiadBench}} & \multicolumn{2}{c}{\textbf{Hmmt25}} 
\\ \cmidrule(lr){2-3} \cmidrule(lr){4-5} \cmidrule(lr){6-7} \cmidrule(lr){8-9} \cmidrule(lr){10-11} \cmidrule(lr){12-13}
\textbf{Method}    & \textit{Len@16}   & \textit{Code\_Line}  & \textit{Len@16}   & \textit{Code\_Line}  & \textit{Len@16}   & \textit{Code\_Line}  & \textit{Len@16}   & \textit{Code\_Line}  & \textit{Len@16}   & \textit{Code\_Line}  & \textit{Len@16}   & \textit{Code\_Line}                 \\ \bottomrule
 \textbf{Base model}
 &   2629.63   &    30.65 
 &   2788.65   &    33.02     
 &   2503.33   &    27.79        
 &   1927.51   &    21.49     
 &   1070.29   &    26.94           
 &   1126.61   &    33.86          \\
\textbf{TIR}
&    620.38    &    10.34   
&    576.61    &    9.51         
&    482.45    &    9.93        
&    447.86    &    9.07             
&    502.86    &    10.50  
&    514.22    &    9.02          \\
\textbf{$\alpha=0.1$} 
&    \textbf{8352.68}   &    \textbf{101.54}  
&    \textbf{8319.16}   &    \textbf{104.73}    
&    \textbf{5856.61}   &    \textbf{70.18} 
&    \textbf{4534.80}   &    \textbf{49.45} 
&    \textbf{6096.70}   &    \textbf{70.58}    
&    \textbf{7805.77}   &    \textbf{92.69}         \\
\rowcolor{blue!10} \textbf{$\alpha=0.01$} 
&    \underline{6655.03}   &    \underline{75.35}  
&    \underline{7217.36}   &    \underline{87.26}    
&    \underline{4145.81}   &    \underline{46.62} 
&    \underline{2828.20}   &    \underline{30.22} 
&    \underline{4746.74}   &    \underline{52.18}    
&    \underline{7161.11}   &    \underline{86.76}         \\
\textbf{$\alpha=0.001$} 
&    5370.72   &    56.84  
&    5422.14   &    63.93    
&    3309.05   &    39.97 
&    2290.91   &    25.51 
&    3723.36   &    41.76    
&    5243.76   &    62.77         \\
\bottomrule 
\end{tabular}
\label{tab:app_alpha_valid_tool_call}
\end{table*}

\begin{table*}[t]
 \caption{Hyperparameter analysis of $\alpha$ on six benchmarks of \textit{Valid\_Code} and \textit{Tool\_Call} under Qwen2.5-7B base model. We bold the best results, underline the sub-optimal results, and highlight our selection.}
\renewcommand{\arraystretch}{1.2}
\setlength{\tabcolsep}{2.5pt} 
    \centering 
\begin{tabular}{lcccccccccccc}
\toprule
\textbf{}                          & \multicolumn{2}{c}{\textbf{AIME24}} & \multicolumn{2}{c}{\textbf{AIME25}} & \multicolumn{2}{c}{\textbf{AMC23}} & \multicolumn{2}{c}{\textbf{MATH500}} & \multicolumn{2}{c}{\textbf{OlympiadBench}} & \multicolumn{2}{c}{\textbf{Hmmt25}} 
\\ \cmidrule(lr){2-3} \cmidrule(lr){4-5} \cmidrule(lr){6-7} \cmidrule(lr){8-9} \cmidrule(lr){10-11} \cmidrule(lr){12-13}
\textbf{Method}    & \textit{Valid\_Code}   & \textit{Tool\_Call}  & \textit{Valid\_Code}   & \textit{Tool\_Call} & \textit{Valid\_Code}   & \textit{Tool\_Call} & \textit{Valid\_Code}   & \textit{Tool\_Call} & \textit{Valid\_Code}   & \textit{Tool\_Call} & \textit{Valid\_Code}   & \textit{Tool\_Call}            \\ \bottomrule
 \textbf{Base model}
 &    81.46    &    1.72  
 &    78.96    &    1.47      
 &    82.19    &    1.51        
 &    84.39    &    \underline{1.45}      
 &    82.75    &    1.53            
 &    81.67    &    1.62          \\
\textbf{TIR}
&     48.96    &    0.49   
&     45.63    &    0.46         
&     56.41    &    0.56         
&     57.40    &    0.57             
&     63.60    &    0.64  
&     44.58    &    0.45          \\
\textbf{$\alpha=0.1$} 
&     96.88    &    \textbf{3.36}  
&     97.08    &    \textbf{3.09}    
&     \underline{96.88}    &    \textbf{2.85} 
&     \underline{97.75}    &    \textbf{2.49} 
&     97.43    &    \textbf{2.82}    
&     95.00    &    \textbf{2.91}         \\
\rowcolor{blue!10} \textbf{$\alpha=0.01$} 
&     \textbf{97.92}    &    \underline{2.26}  
&     \textbf{99.17}    &    \underline{2.35}    
&     \textbf{99.69}    &    \underline{1.79} 
&     \textbf{99.64}    &    \underline{1.45} 
&     \textbf{99.38}    &    \underline{1.87}    
&     \textbf{99.79}    &    \underline{2.33}         \\
\textbf{$\alpha=0.001$} 
&     \underline{97.29}    &    1.92  
&     \underline{97.50}    &    1.93    
&     \textbf{99.69}    &    1.64 
&     \textbf{99.64}    &    1.35 
&     \underline{99.35}    &    1.66    
&     \underline{98.54}    &    1.88         \\
\bottomrule 
\end{tabular}
\label{tab:app_alpha_len_code_line}
\end{table*}

\begin{table*}[t]
 \caption{Hyperparameter analysis of $q_{\mathcal{H}}$ on six benchmarks of \textit{Pass@16} and \textit{Avg@16} under Qwen2.5-7B base model. We bold the best results, underline the sub-optimal results, and highlight our selection.}
\renewcommand{\arraystretch}{1.2}
\setlength{\tabcolsep}{5pt} 
    \centering 
\begin{tabular}{lcccccccccccc}
\toprule
\textbf{}                          & \multicolumn{2}{c}{\textbf{AIME24}} & \multicolumn{2}{c}{\textbf{AIME25}} & \multicolumn{2}{c}{\textbf{AMC23}} & \multicolumn{2}{c}{\textbf{MATH500}} & \multicolumn{2}{c}{\textbf{OlympiadBench}} & \multicolumn{2}{c}{\textbf{Hmmt25}} 
\\ \cmidrule(lr){2-3} \cmidrule(lr){4-5} \cmidrule(lr){6-7} \cmidrule(lr){8-9} \cmidrule(lr){10-11} \cmidrule(lr){12-13}
\textbf{Method}      & \textit{Avg@16}               & \textit{Pass@16}  & \textit{Avg@16} & \textit{Pass@16}  & \textit{Avg@16} & \textit{Pass@16} & \textit{Avg@16} & \textit{Pass@16}  & \textit{Avg@16}  & \textit{Pass@16}     & \textit{Avg@16}     & \textit{Pass@16}  \\ \bottomrule
 \textbf{Base model}
 &   5.00    &     36.67
 &   3.96    &     26.67  
 &   21.56   &     80.00   
 &   31.99   &     88.00  
 &   13.80   &     55.00        
 &   6.67    &     26.67        \\
\textbf{TIR}
&    10.63   &     40.00 
&    10.42   &     20.00       
&    28.91   &     70.00   
&    42.11   &     79.80           
&    20.88   &     49.80 
&    11.04   &     23.33        \\
\textbf{$q_{\mathcal{H}}=10\%$} 
&    \underline{32.50}   &     \underline{70.00} 
&    \underline{21.88}   &     \underline{43.33}   
&    65.47   &     \underline{92.50} 
&    \textbf{78.81}   &     \underline{95.20} 
&    \underline{46.55}   &     \textbf{71.60}   
&    \textbf{19.17}   &     \underline{40.00}        \\
\rowcolor{blue!10} \textbf{$q_{\mathcal{H}}=20\%$} 
&    \textbf{35.42}   &     \textbf{73.33} 
&    \textbf{23.96}   &     \textbf{46.67}   
&    \textbf{69.06}   &     \textbf{97.50} 
&    \underline{78.80}   &     \textbf{95.60} 
&    \textbf{46.99}   &     \underline{70.00}   
&    \underline{18.33}   &     \textbf{46.67}          \\
\textbf{$q_{\mathcal{H}}=30\%$} 
&    31.67   &     \underline{70.00} 
&    21.46   &     \underline{43.33}   
&    \underline{68.13}   &     \underline{92.50}  
&    77.28   &     95.00 
&    45.44   &     69.60   
&    16.04   &     \underline{40.00}        \\
\bottomrule 
\end{tabular}
\label{tab:app_q_avg_pass}
\end{table*}

\begin{table*}[t]
 \caption{Hyperparameter analysis of $q_{\mathcal{H}}$ on six benchmarks of \textit{Len@16} and \textit{Code\_Line} under Qwen2.5-7B base model. We bold the best results, underline the sub-optimal results, and highlight our selection.}
\renewcommand{\arraystretch}{1.2}
\setlength{\tabcolsep}{4pt} 
    \centering 
\begin{tabular}{lcccccccccccc}
\toprule
\textbf{}                          & \multicolumn{2}{c}{\textbf{AIME24}} & \multicolumn{2}{c}{\textbf{AIME25}} & \multicolumn{2}{c}{\textbf{AMC23}} & \multicolumn{2}{c}{\textbf{MATH500}} & \multicolumn{2}{c}{\textbf{OlympiadBench}} & \multicolumn{2}{c}{\textbf{Hmmt25}} 
\\ \cmidrule(lr){2-3} \cmidrule(lr){4-5} \cmidrule(lr){6-7} \cmidrule(lr){8-9} \cmidrule(lr){10-11} \cmidrule(lr){12-13}
\textbf{Method}    & \textit{Len@16}   & \textit{Code\_Line}  & \textit{Len@16}   & \textit{Code\_Line}  & \textit{Len@16}   & \textit{Code\_Line}  & \textit{Len@16}   & \textit{Code\_Line}  & \textit{Len@16}   & \textit{Code\_Line}  & \textit{Len@16}   & \textit{Code\_Line}                 \\ \bottomrule
 \textbf{Base model}
 &   2629.63   &    30.65 
 &   2788.65   &    33.02     
 &   2503.33   &    27.79        
 &   1927.51   &    21.49     
 &   1070.29   &    26.94           
 &   1126.61   &    33.86          \\
\textbf{TIR}
&    620.38    &    10.34   
&    576.61    &    9.51         
&    482.45    &    9.93        
&    447.86    &    9.07             
&    502.86    &    10.50  
&    514.22    &    9.02          \\
\textbf{$q_{\mathcal{H}}=10\%$} 
&    \textbf{6725.54}   &    63.71  
&    6609.16   &    69.72     
&    \underline{3959.16}   &    43.00 
&    \underline{2648.22}   &    26.49 
&    \underline{4496.27}   &    44.83     
&    \underline{6655.13}   &    69.91         \\
\rowcolor{blue!10} \textbf{$q_{\mathcal{H}}=20\%$} 
&    \underline{6655.03}   &    \textbf{75.35}  
&    \textbf{7217.36}   &    \textbf{87.26}    
&    \textbf{4145.81}   &    \textbf{46.62} 
&    \textbf{2828.20}   &    \textbf{30.22} 
&    \textbf{4746.74}   &    \textbf{52.18}    
&    \textbf{7161.11}   &    \textbf{86.76}         \\
\textbf{$q_{\mathcal{H}}=30\%$} 
&    6239.55   &    \underline{69.14}  
&    \underline{6818.38}   &    \underline{82.04}    
&    3659.44   &    \underline{45.15} 
&    2454.78   &    \underline{26.99} 
&    3953.71   &    \underline{44.20}    
&    6438.75   &    \underline{76.70}         \\
\bottomrule 
\end{tabular}
\label{tab:app_q_valid_tool_call}
\end{table*}

\begin{table*}[t]
 \caption{Hyperparameter analysis of $q_{\mathcal{H}}$ on six benchmarks of \textit{Valid\_Code} and \textit{Tool\_Call} under Qwen2.5-7B base model. We bold the best results, underline the sub-optimal results, and highlight our selection.}
\renewcommand{\arraystretch}{1.2}
\setlength{\tabcolsep}{2.5pt} 
    \centering 
\begin{tabular}{lcccccccccccc}
\toprule
\textbf{}                          & \multicolumn{2}{c}{\textbf{AIME24}} & \multicolumn{2}{c}{\textbf{AIME25}} & \multicolumn{2}{c}{\textbf{AMC23}} & \multicolumn{2}{c}{\textbf{MATH500}} & \multicolumn{2}{c}{\textbf{OlympiadBench}} & \multicolumn{2}{c}{\textbf{Hmmt25}} 
\\ \cmidrule(lr){2-3} \cmidrule(lr){4-5} \cmidrule(lr){6-7} \cmidrule(lr){8-9} \cmidrule(lr){10-11} \cmidrule(lr){12-13}
\textbf{Method}    & \textit{Valid\_Code}   & \textit{Tool\_Call}  & \textit{Valid\_Code}   & \textit{Tool\_Call} & \textit{Valid\_Code}   & \textit{Tool\_Call} & \textit{Valid\_Code}   & \textit{Tool\_Call} & \textit{Valid\_Code}   & \textit{Tool\_Call} & \textit{Valid\_Code}   & \textit{Tool\_Call}            \\ \bottomrule
 \textbf{Base model}
 &    81.46    &    1.72  
 &    78.96    &    1.47      
 &    82.19    &    1.51        
 &    84.39    &    \textbf{1.45}      
 &    82.75    &    1.53            
 &    81.67    &    1.62          \\
\textbf{TIR}
&     48.96    &    0.49   
&     45.63    &    0.46         
&     56.41    &    0.56         
&     57.40    &    0.57             
&     63.60    &    0.64  
&     44.58    &    0.45          \\
\textbf{$q_{\mathcal{H}}=10\%$} 
&     \underline{96.88}    &    2.03  
&     \underline{97.92}    &    2.01    
&     \underline{99.06}    &    1.68 
&     \underline{99.35}    &    1.35 
&     \underline{98.59}    &    1.68    
&     \underline{98.54}    &    1.98         \\
\rowcolor{blue!10} \textbf{$q_{\mathcal{H}}=20\%$} 
&     \textbf{97.92}    &    \underline{2.26}  
&     \textbf{99.17}    &    \textbf{2.35}    
&     \textbf{99.69}    &    \textbf{1.79} 
&     \textbf{99.64}    &    \textbf{1.45} 
&     \textbf{99.38}    &    \textbf{1.87}    
&     \textbf{99.79}    &    \textbf{2.33}         \\
\textbf{$q_{\mathcal{H}}=30\%$} 
&     93.96    &    \textbf{2.28}   
&     93.96    &    \underline{2.30}    
&     96.41    &    \underline{1.78} 
&     98.09    &    \underline{1.38} 
&     96.66    &    \underline{1.70}    
&     95.83    &    \underline{2.18}         \\
\bottomrule 
\end{tabular}
\label{tab:app_q_len_code_line}
\end{table*}

\subsubsection{Analysis on Trajectory Filtering Criteria: Retaining All-Wrong Trajectories}
\label{sec:app_keep_wrong}

A natural concern about our trajectory filtering strategy is that removing uniformly incorrect rollout groups may discard useful negative signals, since such groups can still contain valid tool calls or partial reasoning traces. To test this possibility, we introduce a \textbf{Keep-Wrong} variant under the Qwen2.5-7B base model, which discards only all-correct queries with $\rho(q)=1$ while retaining all-incorrect queries with $\rho(q)=0$, provided they satisfy the tool-validity criterion.

Table~\ref{tab:app_keep_wrong} shows that retaining all-wrong rollout groups consistently hurts performance. Across six benchmarks, \textit{Keep-Wrong} reduces the average \textit{Avg@16} from $45.43$ to $42.56$ and the average \textit{Pass@16} from $71.63$ to $67.17$. On AIME24, the degradation is more pronounced, with \textit{Avg@16} dropping by $5.42$ and \textit{Pass@16} by $13.33$. This suggests that all-wrong groups are not difficult-but-useful samples, but low-information instances that dilute effective optimization.

This degradation is consistent with the contrastive nature of GRPO-style optimization. When all sampled rollouts for a query are incorrect, their final-answer rewards are uniformly low, providing little reward contrast under group-normalized advantage estimation and thus yielding zero or near-zero informative policy-gradient signals. Unlike incorrect rollouts in mixed groups satisfying $0<\rho(q)<1$, all-wrong groups cannot support meaningful within-query comparisons between successful and failed reasoning or tool-use behaviors. They therefore mainly consume rollout and optimization budget without contributing discriminative learning signals, confirming the necessity of our answer-discriminability criterion.

\subsubsection{Hyperparameter Analysis}
\label{sec:app_hyperparameter_analysis}

We conduct a systematic sensitivity analysis of two core hyperparameters in the entropy-guided exploration module: the entropy bonus coefficient $\alpha$ and the entropy gating percentile $q_{\mathcal{H}}$. The former controls the magnitude of advantage reshaping at selected post-tool-call tokens, while the latter determines the proportion of high-entropy critical decision points that receive targeted exploration incentives.

\paragraph{Entropy Bonus Coefficient ($\alpha$)}
Tables~\ref{tab:app_alpha_avg_pass}--\ref{tab:app_alpha_len_code_line} report results for $\alpha \in \{0.1, 0.01, 0.001\}$ under Qwen2.5-7B.
An excessively large bonus ($\alpha{=}0.1$) over-incentivizes exploration: \textit{Len@16} surges beyond $8300$ on AIME24/25 and average \textit{Tool\_Call} reaches $2.92$, injecting noise that destabilizes policy updates and degrades accuracy across most benchmarks.
At the other extreme, $\alpha{=}0.001$ provides too weak an incentive, yielding shorter reasoning trajectories and lower code-generation volumes that cap performance on challenging problems.
Setting $\alpha{=}0.01$ strikes the optimal exploration–exploitation balance, achieving the highest \textit{Avg@16} and \textit{Pass@16} on the majority of benchmarks while maintaining stable tool-call frequencies.

\paragraph{Entropy Gating Percentile ($q_{\mathcal{H}}$)}
Tables~\ref{tab:app_q_avg_pass}--\ref{tab:app_q_len_code_line} evaluate $q_{\mathcal{H}} \in \{10\%, 20\%, 30\%\}$.
A stricter gate ($q_{\mathcal{H}}{=}10\%$) restricts advantage reshaping to too few tokens, limiting the diversity of reasoning paths at critical tool-interaction junctures and resulting in fewer tool calls and shorter code lengths.
A looser gate ($q_{\mathcal{H}}{=}30\%$) extends the bonus to tokens where the model is already relatively confident, introducing variance that dilutes the targeted nature of our exploration strategy and noticeably reduces accuracy.
Targeting the top $q_{\mathcal{H}}{=}20\%$ of post-tool-call tokens isolates genuinely uncertain decision points, maximizing reasoning capability without compromising tool-use stability, and consistently yields the strongest \textit{Avg@16} and \textit{Valid\_Code} performance.

\subsubsection{Case Study}

To further demonstrate the robust generalization of \method, we provide two additional case studies from the AIME 2024 test set. These cases highlight our method's capacity for dynamic strategy switching (Table~\ref{tab:case_study_p27}) and conceptual mathematical abstraction (Table~\ref{tab:case_study_p20}), contrasting with the baseline's tendency to succumb to unrecoverable symbolic artifacts or brute-force timeouts.

\subsection{Experimental Platforms and Licenses}
\subsubsection{Platforms}
All experiments of this approach are implemented on two Intel Xeon Platinum 8480+ CPUs and eight NVIDIA A100 GPUs.

\subsubsection{Licenses}
In our implementation, we have utilized the following open-source libraries, each governed by its respective license agreements:
\begin{itemize}
    \item \textbf{VeRL}: Apache License 2.0
    \item \textbf{Ray}: Apache License 2.0
    \item \textbf{vLLM}: Apache License 2.0
    \item \textbf{Weights \& Biase}: MIT License
    \item \textbf{FlashAttention}: BSD 3-Clause License
    \item \textbf{TensorDict}: MIT License
\end{itemize}

\begin{table*}[htbp]
\centering
\caption{Case Study on AIME 2024 Problem 27. \method encounters a \texttt{Timeout} during symbolic computation but successfully analyzes the bottleneck and pivots to a numerical approach, demonstrating robust strategy switching. SimpleTIR, conversely, falls into a hallucination spiral after its symbolic check yields a spurious complex remainder.}
\label{tab:case_study_p27}
\renewcommand{\arraystretch}{1.3}
\begin{tabular}{p{0.48\textwidth} p{0.48\textwidth}}
\toprule
\multicolumn{2}{p{0.96\textwidth}}{\textbf{Problem:} Let $\omega \neq 1$ be a 13th root of unity (i.e., $\omega^{13}=1$). Compute the remainder when $\prod_{k=0}^{12} (2 - 2\omega^k + \omega^{2k})$ is divided by 1000. \vspace{0.2em} \newline  \textbf{Ground Truth:} $321$} \\
\midrule
\rowcolor{gray!15}
\textbf{SimpleTIR (Baseline)} & \textbf{\method (Ours)} \\
\midrule
\begin{minipage}[t]{\linewidth}
To solve the given problem, follow these steps: \newline
1. First, we need to identify the 13th roots of unity. Since $\omega \neq 1$ is a 13th root of unity, $\omega^{13} = 1$ and $\omega \neq 1$, which means $\omega$ is a complex root of unity. \newline
2. Next, we need to compute the expression $\prod_{k=0}^{12}(2-2\omega^k+\omega^{2k})$. \newline
3.  After calculating the product, we need to find the remainder of this product when divided by 1000. \newline
Let's execute the steps in Python to find the final answer.
\begin{lstlisting}[language=Python, escapeinside={(*@}{@*)}, commentstyle=\color{gray}]
import sympy as sp
# Define the 13th root of unity omega
omega = sp.exp(2 * sp.pi * sp.I / 13)
# Calculate the product
product = 1
for k in range(13):
    term = 2 - 2 * omega**k + omega**(2 * k)
    product *= term
# Evaluate the product
product_value = product.evalf()
# Find the remainder when product_value is divided by 1000
(*@\textbf{\ttfamily remainder = abs(product\_value) \% 1000}@*)
print(remainder)
\end{lstlisting}
\vspace{0.1em}
\textcolor{execresult}{\textbf{Code Execution Result: 321.000000000000}} \\
\vspace{0.3em} \\
\textbf{Let's confirm} the accuracy of the result by breaking down the final step in Python. \newline
We have identified that the remainder we are looking for is 321. Before finalizing this, let's ensure that the entire process was carried out accurately via a symbolic calculation. \newline
To do so, let's confirm the final mathematical representation and the remainder via a symbolic calculation using sympy \dots \newline
Let's reimplement this step-by-step in Python to confirm the accuracy.
\begin{lstlisting}[language=Python, escapeinside={(*@}{@*)}, commentstyle=\color{gray}]
# Simplify the product to get a clear representation
(*@\textcolor{red}{\ttfamily product\_simplified = sp.simplify(product)}@*)
product_value = product_simplified.evalf()
remainder = abs(product_value) % 1000
\end{lstlisting}
\vspace{0.1em}
\textcolor{execresult}{\textbf{Result: \dots 654.123528195508 + 18.4839399297822}} \\
\vspace{0.3em} \\
Given the presence of a non-zero imaginary component in the numerical value and remainder, it suggests that our approach might need a refined understanding or computation approach. \newline
We need to re-examine and ensure that the product is indeed real. Let's simplify the problem and re-compute using real arithmetic directly \dots \\
\vspace{0.95em} \newline \fcolorbox{red}{white}{\textbf{Final Answer: WRONG (Failed to resolve)}} \vspace{0.5em}
\end{minipage} &
\begin{minipage}[t]{\linewidth}
To solve the given problem, we'll use the properties of roots of unity and symbolic computation. An $n$-th root of unity is a complex number $\omega$ that satisfies $\omega^n = 1$. \dots \newline
Let's break the problem into smaller steps to find the remainder: \newline
1. Generate the 13th roots of unity, excluding 1. \newline
2. Compute the expression $2 - 2\omega^k + \omega^{2k}$ for each $k$. \newline
3. Find the remainder when the product is divided by 1000. \newline
We can use Python and sympy to perform symbolic computation.
\begin{lstlisting}[language=Python, escapeinside={(*@}{@*)}, commentstyle=\color{gray}]
import sympy as sp
# Define the 13th root of unity
n = 13
omega = sp.exp(2 * sp.pi * sp.I / n)
# Initialize the product
product = 1
# Compute the product
for k in range(n):
    term = 2 - 2 * omega**k + omega**(2*k)
    product *= term
# Simplify the product
(*@\textbf{\ttfamily product\_simplified = sp.simplify(product)}@*)
# Find the remainder when the product is divided by 1000
\end{lstlisting}
\vspace{0.1em}
\textcolor{execresult}{\textbf{Code Execution Result: Timeout}} \\
\vspace{0.3em} \\
\textbf{It seems that} direct symbolic computation might be too complex or inefficient for a straightforward calculation within this environment. We can try a different approach. \newline
\textbf{Alternative Approach}: Instead of directly computing symbolically, we can use numerical methods to evaluate the product due to the complex nature of the roots of unity. \newline
Let's rewrite the code using a numerical approach to evaluate the product. We'll use the numpy library to handle the complex numbers and compute the product.
\begin{lstlisting}[language=Python, escapeinside={(*@}{@*)}, commentstyle=\color{gray}]
import numpy as np
# Evaluate product using numerical approximation
(*@\textbf{\ttfamily omega = np.exp(2j * np.pi / 13)}@*)
product = 1
for k in range(13):
    product *= (2 - 2 * omega**k + omega**(2*k))
remainder = np.round(product.real) % 1000
\end{lstlisting}
\vspace{0.1em}
\textcolor{execresult}{\textbf{Result: The remainder when the product is divided by 1000 is: 321.0}} \\
\vspace{0.3em} \\
Based on the computation, the remainder when $\prod_{k=0}^{12} (2 - 2\omega^k + \omega^{2k})$ is divided by 1000 is $\boxed{321}$. \\
\newline \fcolorbox{green!50!black}{white}{\textbf{Final Answer: 321}} \vspace{0.5em}
\end{minipage} \\
\bottomrule
\end{tabular}
\end{table*}


\clearpage

\begin{table*}[htbp]
\centering
\caption{Case Study on AIME 2024 Problem 20. \method establishes the correct mathematical formula based on binary subset representations, while SimpleTIR formulates an incorrect equation and ultimately defaults to an intractable brute-force search, causing a \texttt{Timeout}.}
\label{tab:case_study_p20}
\renewcommand{\arraystretch}{1.3}
\begin{tabular}{p{0.48\textwidth} p{0.48\textwidth}}
\toprule
\multicolumn{2}{p{0.96\textwidth}}{\textbf{Problem:} Alice chooses a set $A$ of positive integers. Bob then lists all finite nonempty sets $B$ of positive integers such that the maximum element of $B$ belongs to $A$. If Bob's list contains exactly 2024 sets, find the sum of the elements of $A$. \hfill \textbf{Ground Truth:} $55$} \\
\midrule
\rowcolor{gray!15}
\textbf{SimpleTIR (Baseline)} & \textbf{\method (Ours)} \\
\midrule
\begin{minipage}[t]{\linewidth}
To solve this problem, let's first understand what it means for Bob's list to have 2024 sets, and how that relates to the set ( A ). \newline
Say $A = \{a_1, a_2, \dots, a_n\}$. For each $a_i \in A$, we can \\
construct sets $B$ with maximum element $a_i$ \dots \\
This is equivalent to $2^{i-1}$ ways to choose subsets for $a_i$. \\
To get the total number of sets, we sum from $a_1$ to $a_k$: \\
$2^0 + 2^1 + \dots + 2^{k-1} = 2^k - 1$ \\
This is given to be 2024. Let's solve $2^k - 1 = 2024$.
\begin{lstlisting}[language=Python, escapeinside={(*@}{@*)}, commentstyle=\color{gray}]
import sympy as sp
k = sp.symbols('k')
solution = sp.nsolve(sp.Eq(2**k - 1, 2024), k, 11)
\end{lstlisting}
\vspace{0.1em}
\textcolor{execresult}{\textbf{Code Execution Result: 10.9837061926593}} \\
\vspace{0.2em} \\
The solution $k \approx 10.98$ suggests $k = 11$. If we assume \\
$A = \{1, 2, 3, \dots, 11\}$ \dots 
Let's attempt a brute-force method to find the correct sum.
\begin{lstlisting}[language=Python, escapeinside={(*@}{@*)}, commentstyle=\color{gray}]
from itertools import combinations

def count_sets(A):
    count = 0
    n = len(A)
    for r in range(1, n + 1):
        for combo in combinations(A, r):
            count += 1
    return count

# Brute-force search to find set A returning 2024
def find_set_with_2024():
    for s in range(1, 50):  # searching in range
        (*@\textcolor{red}{\ttfamily for comb in combinations(range(1, s+1), s):}@*)
            A = set(comb)
            if count_sets(A) == 2024:
                return sum(A)
    return None

sum_of_elements_in_A = find_set_with_2024()
print(sum_of_elements_in_A)
\end{lstlisting}
\vspace{0.1em}
\textcolor{execresult}{\textbf{Code Execution Result: Timeout}} \\
\vspace{0.2em} \\
The timeout issue suggests that iterating over possible \\
combinations might not be fast enough \dots \\
\vspace{0.6em} \newline \fcolorbox{red}{white}{\textbf{Final Answer: WRONG (Timeout)}} \vspace{0.5em}
\end{minipage} &
\begin{minipage}[t]{\linewidth}
To solve this problem, we need to understand the relationship between the set ( A ) chosen by Alice and the list of sets ( B ) that Bob creates. Specifically, Bob lists all finite nonempty sets ( B ) such that the maximum element of ( B ) belongs to ( A ). \\
1. For a set $B$ to be included, its maximum element must \\
be in $A$. \\
2. For each element $m \in A$, we can form sets $B$ by \\
choosing elements from $\{1, 2, \dots, m-1\}$. The number \\
of such sets $B$ is $2^{m-1}$. \\
\textbf{3. The total number of sets $B$ is the sum of $2^{m-1}$} \\
\textbf{for all $m \in A$. This is given as 2024.} \\
\textbf{So, we have the equation: $\sum_{m \in A} 2^{m-1} = 2024$} \\
4. Find the set $A$ of positive integers such that the sum \\
of $2^{m-1}$ equals 2024. Let's write a Python program.
\begin{lstlisting}[language=Python, escapeinside={(*@}{@*)}, commentstyle=\color{gray}]
# We need to find a set A such that the sum of 2^(m-1) 
# for all m in A equals 2024.
def find_set_A(target_sum):
    from itertools import combinations
    from math import inf
    
    # Start with an empty set and gradually add elements
    best_set = []
    best_sum = inf
    
    # We will check sets up to a reasonable limit, say 20
    for r in range(1, 21):
        for combo in combinations(range(1, 21), r):
            (*@\textbf{\ttfamily current\_sum = sum(2**(m-1) for m in combo)}@*)
            if current_sum == target_sum:
                return combo
            elif current_sum < target_sum and best_sum > sum(combo):
                best_set = combo
                best_sum = sum(combo)
    return best_set

# Find the set A
set_A = find_set_A(2024)
sum_of_elements_A = sum(set_A)
print((set_A, sum_of_elements_A))
\end{lstlisting}
\vspace{0.1em}
\textcolor{execresult}{\textbf{Result: ((4, 6, 7, 8, 9, 10, 11), 55)}} \\
\vspace{0.2em} \\
The set $A$ that satisfies the condition is 
$\{4, 6, 7, 8, 9, 10, 11\}$. Thus, the sum of the elements of ( A ) is (\boxed{55}). \\
\vspace{0.3em} \newline \fcolorbox{green!50!black}{white}{\textbf{Final Answer: 55}} \vspace{0.5em}
\end{minipage} \\
\bottomrule
\end{tabular}
\end{table*}

\end{document}